%% file: main.tex
\documentclass[acmsmall,screen,nonacm,10pt]{acmart}
\usepackage{colortbl}
\usepackage{caption}
\DeclareCaptionFont{surveytablecaption}{\fontsize{8}{10}\selectfont}
\usepackage{longtable}
\usepackage{array}     
\usepackage{multirow}     
\usepackage{ragged2e}    
\usepackage{makecell}    
\usepackage{wrapfig}
\usepackage{needspace}
\newlength{\surveycolumnwidth}
\AtBeginDocument{\setlength{\surveycolumnwidth}{\columnwidth}\setlength{\intextsep}{4pt}\setlength{\textfloatsep}{8pt plus 1pt minus 1pt}\setlength{\dbltextfloatsep}{8pt plus 1pt minus 1pt}\setlength{\floatsep}{8pt plus 1pt minus 1pt}\setlength{\dblfloatsep}{8pt plus 1pt minus 1pt}}
\usepackage{threeparttable}
\usepackage{url}
\usepackage{tablefootnote}
\usepackage{tabularx}
\usepackage{graphicx}             
\usepackage{placeins}
\usepackage{afterpage}
\usepackage{stfloats}      
\usepackage{pdflscape}    
\usepackage[caption=false,font=normalsize,labelfont=sf,textfont=sf]{subfig} 
\usepackage{fontawesome5}
\usepackage{tikz}
\usetikzlibrary{trees}
\usepackage{forest}
\usepackage{algorithm}
\usepackage[noend]{algpseudocode} 
\usepackage[utf8]{inputenc}
\definecolor{lightblue}{rgb}{0.87, 0.92, 0.97}
\definecolor{lightgray}{rgb}{0.84, 0.86, 0.89}

\AtBeginDocument{%
  }
\setcopyright{none}
\acmDOI{}
\acmISBN{}
\acmArticle{}

\begin{document}

\title{LLM-based Agentic Reasoning Frameworks: A Survey from Methods to Scenarios}

\author{Bingxi Zhao}
\authornote{Both authors contributed equally to this research.}
\email{bingxizhao@bjtu.edu.cn}
\affiliation{%
  \institution{Beijing Jiaotong University}
  \city{Beijing}
  \country{China}
}
\affiliation{%
  \institution{Lancaster University}
  \city{Lancaster}
  \country{United Kingdom}
}

\author{Lin Geng Foo}
\authornotemark[1]
\email{lfoo@mpi-inf.mpg.de}
\affiliation{%
  \institution{Max Planck Institute for Informatics, Saarland Informatics Campus}
  \city{Saarbrücken}
  \country{Germany}
}

\author{Ping Hu}
\affiliation{%
  \institution{University of Electronic Science and Technology of China}
  \city{Chengdu}
  \country{China}
}

\author{Christian Theobalt}
\affiliation{%
  \institution{Max Planck Institute for Informatics, Saarland Informatics Campus}
  \city{Saarbrücken}
  \country{Germany}
}

\author{Hossein Rahmani}
\affiliation{%
  \institution{Lancaster University}
  \city{Lancaster}
  \country{United Kingdom}
}

\author{Jun Liu}
\authornote{Corresponding author.}
\email{j.liu81@lancaster.ac.uk}
\affiliation{%
  \institution{Lancaster University}
  \city{Lancaster}
  \country{United Kingdom}
}

\renewcommand{\shortauthors}{Zhao et al.}

\begin{abstract}
Recent advances in LLM-based agents highlight the importance of their reasoning frameworks, which guide the problem-solving process in diverse ways. 
This survey introduces a unified formal language to systematically categorize these frameworks at three compositional levels: single-agent, tool-based, and multi-agent methods. 
Following our taxonomy, we review key application scenarios across scientific discovery, healthcare, software engineering, society, economics, and general-purpose tasks. 
It also compares the distinct features and evaluation strategies of each category. 
Through our taxonomy and comparisons, our survey explores the designs and strengths of LLM-based agentic frameworks in different scenarios, reviewing the fast-paced development of complex agentic systems in the real world.

\end{abstract}


\maketitle
\input{chapters/01_introduction}
\input{chapters/02_related_surveys}
\input{chapters/03_methods}
\input{chapters/04_scenarios}
\input{chapters/05_future_prospects}
\input{chapters/06_limitations}
\input{chapters/07_conclusion}

\bibliographystyle{ACM-Reference-Format-no-links}
\bibliography{main}
\enlargethispage{1.5\baselineskip}
\end{document}

%% file: chapters/01_introduction.tex
\section{Introduction}
\label{sec:introduction}
Large Language Models (LLMs), with their powerful generalization and promising reasoning capabilities, have been rapidly reshaping numerous domains from our daily lives (e.g., idea creation, email writing, or learning of new concepts) to domain-specific research \cite{naveed2023comprehensive}. 
Researchers have been increasingly leveraging LLMs as core components to empower research and innovation \cite{liu2025advances}, from domain-specific knowledge Q\&A \cite{yue2025survey} and code generation \cite{jiang2024survey}, to assisting in research endeavors \cite{liao2024llms}. Through these aspects, LLMs are rapidly becoming a key part of modern life and research.

Yet, despite their immense potential across various fields, LLMs have intrinsic limitations, which may limit their usefulness. For instance, LLMs often suffer from issues such as hallucinations, outdated knowledge, and high training and inference costs \cite{huang2025survey}. These issues often lead to problems in the reliability and consistency of LLMs, and consequently restrict their application in critical fields like healthcare and software engineering, which demand highly dependable outcomes. 

\begin{figure}[t]
  \centering
  \includegraphics[width=1\linewidth]{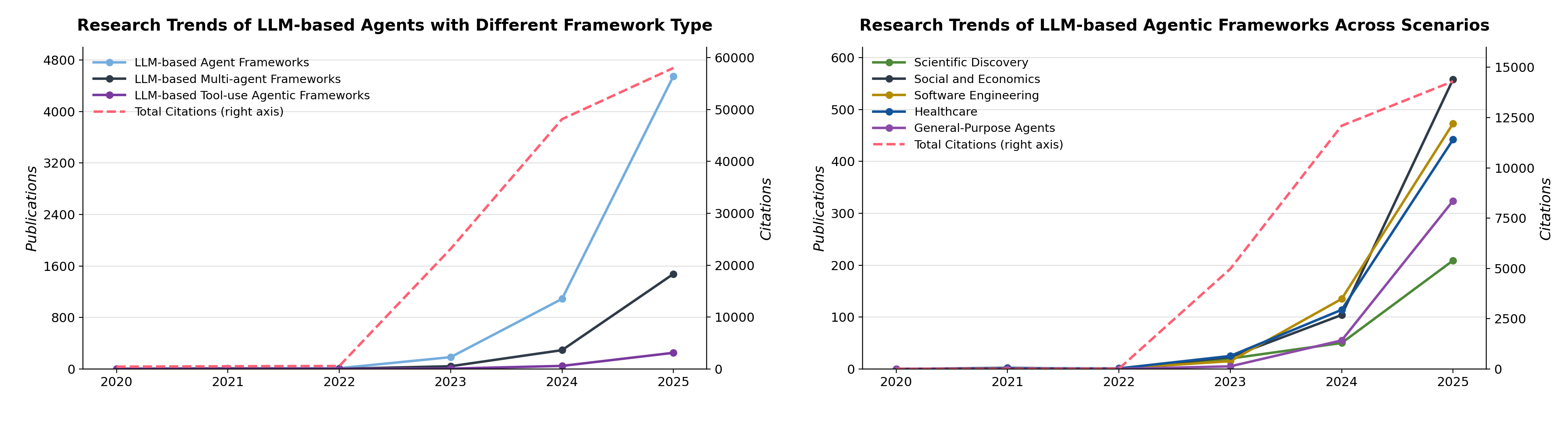}
  \caption[Growth of indexed LLM-agent literature]{Growth of indexed LLM-agent literature from Semantic Scholar Graph API\protect\footnotemark{} keyword searches. \textbf{Left:} three framework queries. \textbf{Right:} four domain queries and a general-purpose agent query. Solid lines count matching journal and conference papers by publication year (2020--2025). Dashed red lines sum citations accrued by papers published in each year in the corresponding citation search, measured at retrieval. All queries were retrieved on 29 September 2026; recent counts may reflect indexing lag.}
  \Description{Two line charts span 2020 to 2025. The left chart shows indexed publication counts for broad LLM-agent, multi-agent and tool-use framework keyword searches, with a separate citation-cohort curve on the right axis. The right chart shows indexed publication counts for scientific discovery, social and economics, software engineering, healthcare, and general-purpose agent keyword searches, with a separate citation-cohort curve on the right axis. In 2025 the five right-panel series respectively reach 209, 558, 473, 442 and 324 records.}
  \label{fig:trend}
\end{figure}
\afterpage{\footnotetext[1]{\url{https://www.semanticscholar.org/product/api}}}

To overcome this barrier, the academic community has been actively exploring the use of LLMs as a core engine to build LLM-based agentic reasoning frameworks capable of executing complex, multi-step reasoning tasks \cite{wang2024survey,qu23lmc}. In the indexed sample shown in Fig.~\ref{fig:trend}, publication counts rose by more than an order of magnitude between 2023 and 2025. Initially, ``Agents'' are defined in \cite{russell2016artificial} as systems that ``perceive their environment through sensors and act upon that environment through actuators'', can dynamically adapt to their environments and take corresponding actions \cite{liu2025advances}. This emerging paradigm organically integrates key modules like planning, memory, and tool-use, reshaping the LLM into a task executor that can perceive its environment, adapt dynamically, and take sustained action \cite{wang2024survey,huang2024understanding,li2025review}. These frameworks extend a model invocation into a workflow that can retrieve evidence, use tools and revise intermediate decisions in response to feedback.

This trend has also been widely mirrored in industry, where tech giants are actively integrating agent workflows into their core businesses. For instance, frameworks like Microsoft's AutoGen\footnote{\url{https://github.com/microsoft/autogen}} are designed to empower enterprises to build customized multi-agent applications. Moreover, from ``vibe coding'' editors like Cursor\footnote{\url{https://cursor.com/en/dashboard}} that deeply integrate agentic capabilities to autonomous AI software engineers like Devin\footnote{\url{https://devin.ai/}}, a clear evolution based on agentic reasoning frameworks is gaining widespread recognition, gradually replacing traditional development approaches.

However, at the same time, the explosive growth in this field has also blurred the boundaries of LLM-based agents \cite{yehudai2025survey}. For instance, the overlap with concepts from areas like traditional multi-agent systems \cite{gronauer2022multi,zhang2024multi,chai2025survey} and autonomous systems \cite{tang2022perception} makes it difficult to define the scope of research. Meanwhile, it is often hard to clearly separate whether enhanced capabilities of an agent come from careful framework design, model-level improvements, or technological advancements. This dual ambiguity poses a serious challenge for the horizontal comparison of different projects and risks overlooking the foundational role of framework design in an agent system's reasoning ability.

Therefore, we believe that it is timely for a survey to systematically summarize the \textit{recent progress and application scenarios of agentic reasoning frameworks}. We first clearly define the boundaries of these frameworks and, based on that, propose a unified methodological classification system. We then further analyze the application and evaluation strategies of these methods across diverse scenarios, aiming to provide a clear roadmap for the standardized and safe development of agentic developments. We also discuss how our taxonomy relates to long-term memory, knowledge management, continual learning and context engineering in \S\ref{sec:emerging}, and examine these capabilities in general-purpose systems in \S\ref{sec:general_agents}.

Overall, the contributions of our survey are as follows:
\begin{itemize}
\item We propose a unified methodological taxonomy that connects the core reasoning mechanisms of agentic frameworks across single-agent, tool-based and multi-agent levels;
\item We employ a formal language to describe the reasoning process, clearly illustrating the impact of different methods on key steps;
\item We extensively investigate the application of agent reasoning frameworks in several key domain-specific and general-purpose scenarios. In these application scenarios, we conduct in-depth analyses of representative works according to our proposed taxonomy, present a collection of evaluation setups with datasets, and distill cross-scenario design considerations that connect task constraints to framework choices (Table~\ref{tab:cross_scenario}, \S\ref{sec:scenario_synthesis}).
\end{itemize}

The structure of the survey is as follows:
Chapter \S\ref{sec:related} introduces related surveys and compares them with our survey. Chapter \S\ref{sec:method} will present the taxonomy of techniques, which systematically analyses the existing techniques for agentic reasoning. Chapter \S\ref{sec:scenarios} will further provide application scenarios of agentic reasoning frameworks, and how agents in each scenario are often designed. Lastly, Chapter \S\ref{sec:future} will discuss future directions, Chapter \S\ref{sec:limitations} discusses limitations, and Chapter \S\ref{sec:conclusion} states the conclusion of the survey.

%% file: chapters/02_related_surveys.tex
\section{Related Surveys}
\label{sec:related}
Recent surveys on agentic AI have explored the agentic reasoning landscape from several valuable perspectives. A major focus has been model-centric, examining how to enhance the reasoning capabilities of LLMs through training and inference-time methods~\cite{sun2025survey,xu2025towards}. Other surveys explore smaller, specialized models~\cite{belcak2025small}, planning abilities~\cite{huang2024understanding}, and role-and-workflow paradigms~\cite{li2025review}. These perspectives also extend toward agent systems: Ke et al.~\cite{ke2025survey} connect model-level reasoning to single- and multi-agent methods, while context-engineering surveys examine how information is organized for reasoning~\cite{mei2025survey}.

Alongside these developments, numerous agentic frameworks have emerged that use LLMs, external tools and interacting agents to organize multi-step reasoning. Existing framework surveys provide substantial coverage of this direction, including functional modules~\cite{wang2024survey}, agent architectures and communication~\cite{masterman2024landscape,guo2024large}, construction and evolution~\cite{luo2025large}, and reasoning, acting and interacting~\cite{plaat2025agentic}. Building on these perspectives, our survey specifically concentrates on how reasoning is organized at the framework level. We categorize methods along three compositional levels: single-agent, tool-based and multi-agent. We relate these methods through a unified notation and reasoning algorithm. A framework can combine mechanisms from these levels; the taxonomy describes their roles within the same reasoning process. This focus complements advances in model architectures and training techniques.

Closer to our application-oriented discussion, domain surveys examine how agentic technologies support scientific discovery~\cite{gridach2025agentic,ren2025towards}, software engineering and code reasoning~\cite{jin2024llms,codereasoningsurvey2025}, medicine~\cite{wang2025survey}, and financial trading~\cite{ding2024large}. These surveys provide depth within their respective fields, while their task-specific classifications make direct comparison across domains less straightforward. For instance, software tools can return executable feedback, clinical reasoning requires evidence and review, and social simulations preserve interacting objectives. Relating these requirements through a common taxonomy helps explain how framework designs are adapted to each scenario. We therefore connect the methodological account in Section~\ref{sec:method} to application-specific comparisons and the cross-scenario synthesis in Section~\ref{sec:scenario_synthesis}. Furthermore, our scenario-driven evaluation review traces the datasets, protocols and reported conditions used by representative systems. Our contribution lies in connecting methods, scenarios and evaluation so that readers can follow both the composition of a framework and the conditions under which its reported capabilities are assessed.

Table~\ref{tab:related_surveys} compares coverage under \textit{Method}, \textit{Scenarios} and \textit{Evaluation}. Under Method, shared formalism relates mechanisms through common notation or an algorithm. Under Scenarios, domain breadth spans application families; task--design links relate task requirements to framework choices across settings; deployment constraints address resources, integration, permissions and human oversight. Under Evaluation, benchmark mapping organizes evaluation resources; experiment protocols describe execution and scoring; comparison validity examines confounding settings, evaluation bias and limits to attribution or transfer. Table headers abbreviate these terms. Filled circles indicate focused treatment; open circles indicate aspects not developed at that scope.

\begin{table}[!t]
\centering
\captionsetup{skip=3pt}\caption{Coverage of related surveys published through 2025, ordered from older to newer, across methods, scenarios and evaluation. $\bullet$: focused coverage; $\circ$: not a focus. The comparison criteria are defined in the text.}
\label{tab:related_surveys}
\begingroup
\newcommand{\surveyCovered}{\tikz[baseline=-0.55ex]{\fill[fill=.] (0,0) circle (1.5pt);}}
\newcommand{\surveyOpen}{\tikz[baseline=-0.55ex]{\draw[draw=.,line width=0.4pt] (0,0) circle (1.3pt);}}
\fontsize{7.4}{8.4}\selectfont
\setlength{\tabcolsep}{1.15pt}\renewcommand{\arraystretch}{0.96}
\begin{tabularx}{\textwidth}{@{}lc>{\hsize=1.12\hsize\linewidth=\hsize\centering\arraybackslash}X>{\hsize=0.73\hsize\linewidth=\hsize\centering\arraybackslash}X>{\hsize=0.79\hsize\linewidth=\hsize\centering\arraybackslash}X>{\hsize=1.08\hsize\linewidth=\hsize\centering\arraybackslash}X>{\hsize=0.87\hsize\linewidth=\hsize\centering\arraybackslash}X>{\hsize=1.3\hsize\linewidth=\hsize\centering\arraybackslash}X>{\hsize=1.13\hsize\linewidth=\hsize\centering\arraybackslash}X>{\hsize=0.97\hsize\linewidth=\hsize\centering\arraybackslash}X>{\hsize=0.99\hsize\linewidth=\hsize\centering\arraybackslash}X>{\hsize=1.02\hsize\linewidth=\hsize\centering\arraybackslash}X@{}}
\toprule
\textbf{Survey} & \textbf{Year} & \multicolumn{4}{c}{\textbf{Method}} & \multicolumn{3}{c}{\textbf{Scenarios}} & \multicolumn{3}{c}{\textbf{Evaluation}} \\
\cmidrule(lr){3-6}\cmidrule(lr){7-9}\cmidrule(l){10-12}
& & \shortstack{Reasoning\\structure} & \shortstack{Tool\\use} & \shortstack{Agent\\coord.} & \shortstack{Shared\\formalism} & \shortstack{Domain\\breadth} & \shortstack{Task--design\\links} & \shortstack{Deploy.\\constraints} & \shortstack{Bench.\\mapping} & \shortstack{Expt.\\protocols} & \shortstack{Compar.\\validity} \\
\midrule
Ding et al.~\cite{ding2024large} & 2024 & \surveyCovered & \surveyCovered & \surveyCovered & \surveyOpen & \surveyOpen & \surveyOpen & \surveyCovered & \surveyCovered & \surveyCovered & \surveyCovered \\
Guo et al.~\cite{guo2024large} & 2024 & \surveyCovered & \surveyCovered & \surveyCovered & \surveyOpen & \surveyCovered & \surveyOpen & \surveyCovered & \surveyCovered & \surveyOpen & \surveyOpen \\
Huang et al.~\cite{huang2024understanding} & 2024 & \surveyCovered & \surveyCovered & \surveyOpen & \surveyCovered & \surveyOpen & \surveyOpen & \surveyOpen & \surveyCovered & \surveyCovered & \surveyCovered \\
Jin et al.~\cite{jin2024llms} & 2024 & \surveyCovered & \surveyCovered & \surveyCovered & \surveyOpen & \surveyOpen & \surveyOpen & \surveyCovered & \surveyCovered & \surveyCovered & \surveyCovered \\
Masterman et al.~\cite{masterman2024landscape} & 2024 & \surveyCovered & \surveyCovered & \surveyCovered & \surveyOpen & \surveyOpen & \surveyOpen & \surveyCovered & \surveyOpen & \surveyCovered & \surveyCovered \\
Wang et al.~\cite{wang2024survey} & 2024 & \surveyCovered & \surveyCovered & \surveyCovered & \surveyOpen & \surveyCovered & \surveyOpen & \surveyCovered & \surveyCovered & \surveyCovered & \surveyCovered \\
\addlinespace[2pt]
Gridach et al.~\cite{gridach2025agentic} & 2025 & \surveyCovered & \surveyCovered & \surveyCovered & \surveyOpen & \surveyOpen & \surveyOpen & \surveyCovered & \surveyCovered & \surveyCovered & \surveyCovered \\
Ke et al.~\cite{ke2025survey} & 2025 & \surveyCovered & \surveyCovered & \surveyCovered & \surveyCovered & \surveyOpen & \surveyOpen & \surveyOpen & \surveyOpen & \surveyCovered & \surveyCovered \\
Li~\cite{li2025review} & 2025 & \surveyCovered & \surveyCovered & \surveyOpen & \surveyCovered & \surveyOpen & \surveyOpen & \surveyOpen & \surveyCovered & \surveyOpen & \surveyOpen \\
Luo et al.~\cite{luo2025large} & 2025 & \surveyCovered & \surveyCovered & \surveyCovered & \surveyOpen & \surveyCovered & \surveyOpen & \surveyCovered & \surveyCovered & \surveyCovered & \surveyCovered \\
Mei et al.~\cite{mei2025survey} & 2025 & \surveyCovered & \surveyCovered & \surveyCovered & \surveyCovered & \surveyCovered & \surveyOpen & \surveyCovered & \surveyCovered & \surveyCovered & \surveyCovered \\
Plaat et al.~\cite{plaat2025agentic} & 2025 & \surveyCovered & \surveyCovered & \surveyCovered & \surveyOpen & \surveyCovered & \surveyOpen & \surveyCovered & \surveyCovered & \surveyOpen & \surveyOpen \\
Ren et al.~\cite{ren2025towards} & 2025 & \surveyCovered & \surveyCovered & \surveyCovered & \surveyOpen & \surveyCovered & \surveyCovered & \surveyCovered & \surveyCovered & \surveyOpen & \surveyCovered \\
Wang et al.~\cite{wang2025survey} & 2025 & \surveyCovered & \surveyCovered & \surveyCovered & \surveyOpen & \surveyOpen & \surveyOpen & \surveyCovered & \surveyCovered & \surveyCovered & \surveyCovered \\
\midrule
\rowcolor{lightblue}
\textbf{This survey} & -- & \surveyCovered & \surveyCovered & \surveyCovered & \surveyCovered & \surveyCovered & \surveyCovered & \surveyCovered & \surveyCovered & \surveyCovered & \surveyCovered \\
\bottomrule
\end{tabularx}
\endgroup
\end{table}

\paragraph{Paper Selection.}\label{par:literature_selection}Core coverage centers on 2023--2025, with foundational work and targeted updates through 9 September 2026. For methods (Section~\ref{sec:method}), we emphasize major AI venues, including ICLR, NeurIPS, ACL, EMNLP, AAAI and ICML. For scenarios (Section~\ref{sec:scenarios}), we also draw on domain journals, including \textit{Nature}, \textit{Science}, \textit{Cell}, \textit{Nature Machine Intelligence} and \textit{Nature Medicine}. Searches across venue archives, ACM/IEEE journals and arXiv combine LLM, agent, reasoning and tool-use terms with domain keywords; benchmark-led reference tracing supplements this coverage. For established work, we give priority to papers in leading conferences and journals, using citation uptake and documented reuse as signals of influence rather than fixed thresholds. Citation counts alone would disadvantage recent work in this rapidly developing field. We therefore also include selected, publicly inspectable preprints and industrial reports from influential academic groups, companies and research institutions when they have drawn field attention and contribute distinctive framework mechanisms, reusable designs or evaluation resources. We also discuss coverage and source limitations in Section~\ref{sec:limitations}.

\begin{figure*}[t]
    \centering
    \includegraphics[width=1\linewidth]{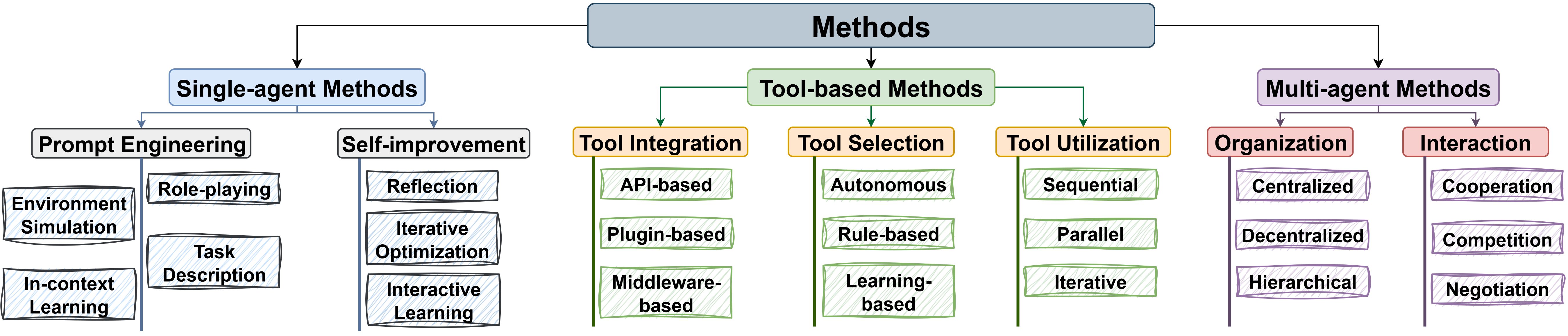}
    \caption{Taxonomy of agentic reasoning frameworks. We organize mechanisms at three compositional levels: \textbf{a)} single-agent methods, \textbf{b)} tool-based methods, and \textbf{c)} multi-agent methods. A framework can combine mechanisms across these levels.}
    \Description{This image presents a detailed, hierarchical breakdown of agentic reasoning frameworks. The taxonomy is organized into three main categories: Single-agent Methods, which focus on prompt engineering and self-improvement techniques; Tool-based Methods, which cover how agents integrate, select, and utilize external tools; and Multi-agent Methods, which describe how multiple agents are organized and how they interact with each other. The diagram effectively visualizes the various sub-categories within each method, such as reflection, iterative optimization, and different forms of tool integration and agent interaction. This structure provides a clear and comprehensive overview of the different approaches used in designing agentic reasoning systems.}
    \label{fig:taxonomy}
\end{figure*}

%% file: chapters/03_methods.tex
\section{Methods}
\label{sec:method}
Extended from foundation LLMs, agentic reasoning frameworks are a key development toward autonomous, environment-aware systems that can solve complicated problems in the real world. In this section, we propose a taxonomy to categorize these methods. At the top, we deconstruct the reasoning framework into three compositional levels, namely single-agent, external tool calling, and multi-agent, as shown in Figure \ref{fig:taxonomy}.  \textbf{\textit{Single-agent methods}} focus on enhancing the reasoning capability of individual agents; \textbf{\textit{tool-based methods}} extend the boundaries of agent reasoning through external tools; and \textbf{\textit{multi-agent methods}} enable more flexible reasoning through different paradigms of organization and interaction among multiple agents. 
We cover these levels in Chapters \S\ref{sec:single_agent}, \S\ref{sec:tool_based}, \S\ref{sec:multi_agent} respectively, after we introduce the notations in Chapter \S\ref{sec:preliminaries}.
Together, these methods at different levels can be integrated in different ways to fulfill specific scenarios, which is covered in Chapter \S\ref{sec:scenarios}.

\subsection{Notations}
\label{sec:preliminaries}
We highlight that agent systems can achieve their goals through a complete process that includes multiple reasoning steps. Multi-agent systems can further execute a complete reasoning process under the collaboration of different agents \cite{tran2025multi}. To clearly introduce this complex process, we propose a general reasoning algorithm (Alg. \ref{alg}) and a notation table (Table \ref{tab:notation}) in this section, which brings another level of abstraction. In the subsequent chapters (Chapters \S\ref{sec:single_agent}, \S\ref{sec:tool_based}, and \S\ref{sec:multi_agent}), we will further discuss how each representative line of works improve the reasoning performances by modifying or adjusting this general algorithm.

A key differentiator between an agentic system and a standard Large Language Model is the ability to coordinate multi-step reasoning with external actions \cite{guo2024large}. This capability relies on the active management of a persistent context throughout the lifecycle of a task within the agentic system \cite{mei2025survey}. While a model invocation produces an output from a supplied context, an agent system, based on its various action choices, iteratively updates its context to support multi-step reasoning. Each action, though it has different targets or intentions, follows a similar logic to tackle such input-output relations. Therefore, we formalize a single reasoning step as an operation where the agent executes an action $a$ based on its current context $\mathcal{C}$ to produce an output $ y$, expressed as $y=a(\mathcal{C})$. A full reasoning process will contain several such reasoning steps.

\begin{table*}[!t]
\centering
\resizebox{\surveycolumnwidth}{!}{%
\begin{minipage}{\surveycolumnwidth}
\centering
    \caption{Notations Used in This Chapter}
    \label{tab:notation}
    \begin{tabularx}{\columnwidth}{l >{\RaggedRight\arraybackslash}X}
        \toprule 
        \textbf{Notation} & \textbf{Description} \\
        \midrule 
        $P_U$ & The user's input query. \\
        $Q$ & The termination condition for the reasoning process. \\
        $g$ & The current goal or set of goals; indexed by $k$ when adapted. \\
        $t$ & An external tool available to the agent. \\
        $\mathcal{C}$ & The internal context of an agent. \\
        $\mathcal{R}$ & Optional retained state available across tasks; empty when no state is reused. \\
        $y$ & The output of an agent after an action. \\
        $k$ & A count of reasoning steps. \\
        \midrule 
        $\mathcal{A}$ & The output-producing action types considered in this abstraction. \\
        $a$ & A general action that produces an output from a given input. \\
        $a'$ & An action that updates the current context based on an input. \\
        $a_{\text{reason}}$ & An action that performs a step of deep reasoning. \\
        $a_{\text{tool}}$ & An action that involves an interaction with an external tool $t$. \\
        $a_{\text{reflect}}$ & An action to reflect on and evaluate previous reasoning steps. \\
        \bottomrule 
    \end{tabularx}
\end{minipage}%
}
\end{table*}

The outputs and insights from the preceding steps are preserved within this context, enabling the agent to build upon its prior work \cite{mei2025survey}. We explicitly distinguish the action of generating an output ($a$) from the action of updating the state ($a'$). This separation is crucial because the objective of a context update (e.g., summarizing history, integrating a tool's results) often differs from that of producing a final or intermediate answer \cite{mei2025survey}.

To execute these steps, the agent selects actions from a generalized action space $\mathcal{A}$, which we define for our purposes as $ \mathcal{A} = \{ a_{\text{reason}}, a_{\text{tool}}, a_{\text{reflect}}\}$. 
To maintain focus on the reasoning logic, our framework abstracts access to components that support reasoning, including persistent memory \cite{zhang2024survey}, knowledge retrieval \cite{zhao2024retrieval}, sandboxed environments \cite{durante2024agent}, and human feedback \cite{natarajan2025human} through a unified external interface $t$. The action $a_{\text{tool}}$ is specifically designed and used to invoke this tool, providing the agent with necessary external information or capabilities. The abstraction describes access to these capabilities; mandatory approval and interruption remain execution constraints that the agent cannot bypass. While this action space is sufficient for our analysis, it can be extended or tailored for specific domains \cite{wang23voyager}.

Consequently, a complete reasoning task is modeled as an iterative sequence of actions. The process is initiated by a user query $P_U$ and proceeds until a predefined termination condition $Q$ is met. This condition is essential for ensuring controlled execution and conserving computational resources \cite{tran2025multi}. Building on the notations in Table \ref{tab:notation}, we formalize this multi-step reasoning process in Algorithm \ref{alg}. The comments within the algorithm serve as forward references, indicating which of the methodologies discussed in subsequent sections modify a particular step of the general procedure.  Its optional goal-update step applies when the framework permits adaptation of $g$; otherwise the goal remains fixed. Unprimed actions belong to $\mathcal{A}$ and produce outputs; primed actions are separately defined context-update operators. The algorithm describes one task. When a framework reuses experience, $\mathcal{R}$ denotes the retained memory, skills or controller configuration available at its start; initialization and tool access expose the relevant parts to reasoning. Persisting changes to this state makes them available to later tasks. Such reuse can preserve a fixed $g$ and does not imply a change to model parameters.

\begin{algorithm}
\caption{General Algorithm for Framework-level Agentic Reasoning}
\label{alg}
\begin{algorithmic}[1] 
    \Require User Query $P_U$; Goal $g$; External Tool $t$; Action Space $\mathcal{A}$; Terminate Condition $Q$; Optional Retained State $\mathcal{R}$
    \Ensure Final Output $y_{\text{out}}$
    \State Initialize context $\mathcal{C}_0 \gets \text{Init}(P_U;\mathcal{R})$ \Comment{\S\ref{sec:single_agent} (eq.\ref{eq1},eq.\ref{eq3}); \S\ref{sec:multi_agent} (eq.\ref{eq11})}
    \State Initialize reasoning step $k \gets 0$
    \While{$\neg Q(\mathcal{C}_k, k)$}
    \Comment{\S\ref{sec:single_agent} (eq.\ref{eq4})}
        \State $y_{k+1} = a_k(\mathcal{C}_k, g, t), \quad a_k \in \mathcal{A}$
        \Comment{\S\ref{sec:tool_based} (eq.\ref{eq6},eq.\ref{eq7},eq.\ref{eq9})}
        \State $\mathcal{C}_{k+1} = a'_k(\mathcal{C}_k, y_{k+1}, g, t)$
        \Comment{\S\ref{sec:single_agent} (eq.\ref{eq2}); \S\ref{sec:tool_based} (eq.\ref{eq8},eq.\ref{eq10}); \S\ref{sec:multi_agent} (eq.\ref{eq12})}
        \State Optionally update $g$ using the current context and feedback
        \Comment{\S\ref{sec:single_agent} (eq.\ref{eq5}); \S\ref{sec:multi_agent} (eq.\ref{eq13},eq.\ref{eq14},eq.\ref{eq15})}
        \State $k \gets k + 1$
    \EndWhile
    \State \Return Final output derived from $\mathcal{C}_k$ 
\end{algorithmic}
\end{algorithm}

\subsection{Single-agent Methods}
\label{sec:single_agent}
Single-Agent methods focus on enhancing the cognitive and decision-making abilities of an individual agent. From the perspectives of external guidance and internal optimization, this part categorizes single-agent methods into two main types: \textit{\textbf{prompt engineering}} and \textit{\textbf{self-improvement}}. Prompt engineering emphasizes guiding the agent’s reasoning process by leveraging roles, environments, tasks, and examples, while self-improvement focuses on how the agent refines its reasoning strategies through reflection, iteration, and interaction. 

\subsubsection{Prompt Engineering}
\begin{figure}
    \centering
    \includegraphics[width=1\linewidth]{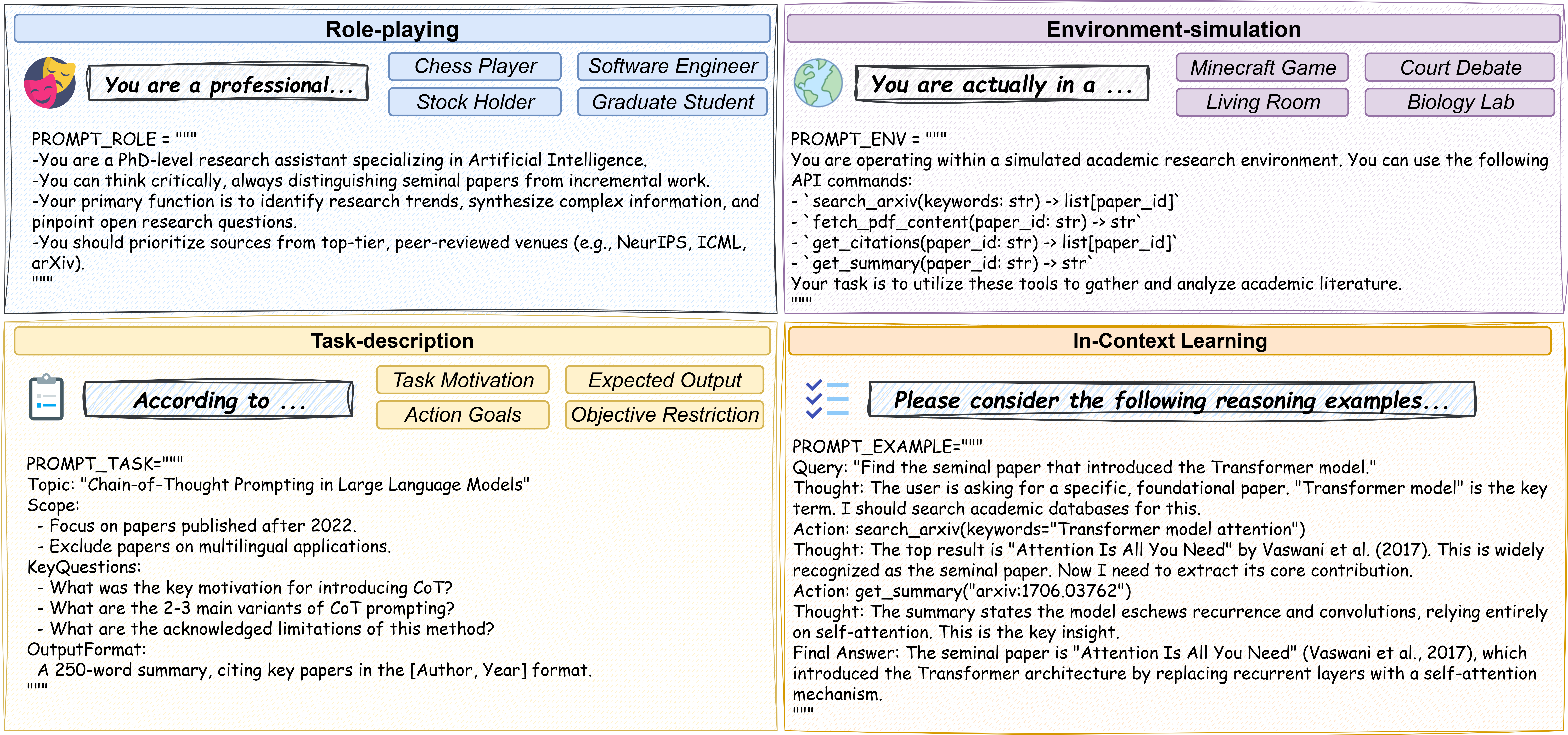}
    \caption{Prompt engineering for agentic reasoning frameworks. We summarize four types of methods: \textit{\textbf{a) Role-playing}}: an agent will be distributed with a specific role, to stimulate its specific performance; \textit{\textbf{b) Environmental-simulation}}: an agent will be told in a carefully designed environment, where it can reason with multi-modalities or external abilities; \textit{\textbf{c) Task-description}}: a task will be clearly reconstructed and expressed to an agent; \textit{\textbf{d) In-context Learning}}: several examples will be provided to an agent before or during its multi-step reasoning. 
    For each type of prompting method, we provide a short example prompt, with the theme of conducting agentic research.} 
    \Description{This image is a detailed diagram illustrating four types of Prompt Engineering methods for agentic reasoning frameworks. The diagram breaks down each method with a short description and a corresponding example.}
    \label{fig:prompt}
\end{figure}

Prompt engineering enhances the agent's performance by enriching its initial context, which corresponds to the context initialization step\textit{ (line 1 in Alg.\ref{alg}).} \cite{schulhoff2024prompt}. Instead of relying solely on the user's query ($ P_U $), this approach augments the initial context $\mathcal{C}_0$ with a meticulously crafted prompt, denoted as $P^*$. This conceptual shift can be represented as:
\begin{equation}\label{eq1}
\mathcal{C}_0 \gets \text{Init}(P_U;\mathcal{R}) \quad \xrightarrow{\text{Prom. Eng.}} \quad \mathcal{C}_0 \gets \text{Init}(P_U, P^*)
\end{equation}
Equation \ref{eq1} illustrates that the initialization process is transformed. Originally, the context 
$\mathcal{C}_0$ is derived exclusively from the user query $P_U$. With prompt engineering, it is initialized with both $P_U$ and the engineered prompt $P^*$. This additional prompt $P^*$ is often a composite of several components: a \textit{role-playing perspective} (\(P_{\text{role}}\)), an \textit{environment simulation} (\(P_{\text{env}}\)), a detailed \textit{task clarification} ($P_{\text{task}}$), and a set of \textit{in-context examples} ($P_{\text{icl}}$).
Unlike fine-tuning methods, which alter the LLM's parameters, prompt engineering guides the model's behavior non-intrusively, steering the agent towards more accurate and predictable reasoning outcomes \cite{liu2023pretrainpromptpredict}. Each component of $P^*$ contributes to this guidance in a distinct way, as detailed below and illustrated in Fig. \ref{fig:prompt}.
\paragraph{Role-Playing}
To instill a role-playing perspective ($P_{\text{role}}$), the prompt assigns the agent a specific persona or identity, such as ``You are an expert data scientist'' or ``Act as a seasoned historian" \cite{shanahan2023role}. This encourages the agent to leverage the expertise, cognitive frameworks, and linguistic styles associated with that role  \cite{salewski2023impersonation}. By adopting a persona, the model can better activate domain-specific knowledge and structure its responses from a more professional viewpoint during reasoning \cite{kong2024better}. 
This technique has become a widely adopted method in the agentic frameworks discussed in chapter \S\ref{sec:scenarios}, owing to its low deployment cost and high guidance efficiency. By assigning a clear role, it enables agents to better focus on their specific duties, thereby optimizing their reasoning and decision-making processes in complex tasks.
However, the efficiency of role assignment can be sensitive to the granularity of the persona design and the specifics of the task \cite{kim2024personadoubleedgedswordmitigating}. Furthermore, for fact-based questions, role-playing may introduce biases inherent to the persona, potentially leading to factually inaccurate outputs \cite{li2024benchmarkingbiaslargelanguage}.

\paragraph{Environment Simulation}
The environment simulation prompt ($P_{\text{env}}$) contextualizes the agent by describing the specific setting in which it operates. This provides task-relevant background information, rules, and constraints, enabling the agent to make decisions that are better aligned with the simulated world. These environments can range from mimicking real-world scenarios, such as a stock market \cite{fatemi2024finvision} or a medical clinic \cite{fan2025ai}, to entirely virtual settings like a video game world \cite{wang23voyager}, often with a action space that are carefully designed. A detailed and task-relevant environmental description is critical, as it prompts the agent to generate actions that are contextually appropriate and highly correlated with the scenario's objectives.

\paragraph{Task Description}
A clear task description ($P_{\text{task}}$), which outlines the primary goal $g$, constraints, and expected output format, is a cornerstone of virtually every agent system. A well-structured task description guides the agent in decomposing a complex problem into a sequence of manageable sub-tasks. 
By providing a precise description, agents can better comprehend the task's intent and execute it in the specified manner, which effectively reduces ambiguity during the reasoning process and leads to more accurate outcomes \cite{min2022rethinking}. 
However, the verbosity and structure of the task description can significantly impact the performance of the underlying LLM, often requiring careful optimization tailored to the specific model being used \cite{levy2024same}.

\paragraph{In-context Learning}

In-context learning (ICL) provides the agent with a set of few-shot examples, or demonstrations, within the prompt ($P_{\text{icl}}$). These examples typically take the form of pairs \(\{(x_1, z_1), (x_2, z_2), \dots, (x_n, z_n)\}\), where each pair \((x_j, z_j)\) consists of an exemplary input \(x_j\) and its corresponding desired output \(z_j\) \cite{brown2020language}. This allows the agent to discern patterns and generalize to new task instances without any gradient updates. 
Chain-of-thought prompting \cite{wei2022chain} further provides a paradigm in which intermediate reasoning steps are also supplied to the agent, teaching agents how to reason, plan and break down problems without internal tuning \cite{zhang2025igniting}. Least-to-most prompting~\cite{zhou2023leasttomostprompting} first decomposes a problem into subproblems, then solves them sequentially using previous answers.
However, the performance of ICL is highly sensitive to the quality and relevance of the provided examples; low-quality or irrelevant demonstrations can significantly degrade the agent's reasoning capabilities \cite{li2023finding}.

\subsubsection{Self-Improvement}
\begin{figure}
    \centering
    \includegraphics[width=1\linewidth]{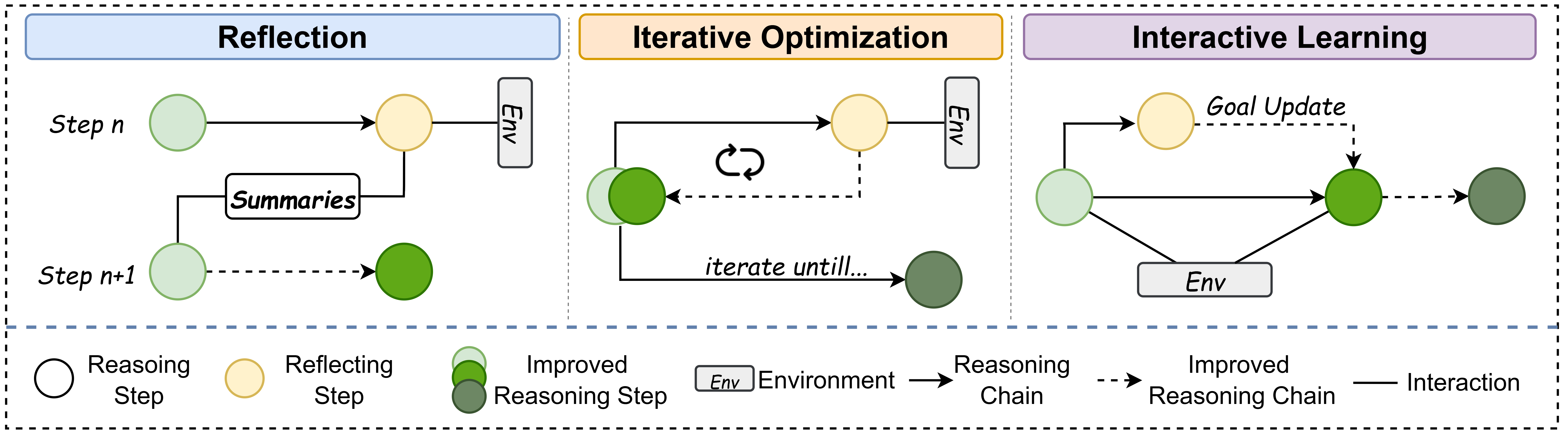}
    \caption{Paradigms of self-improvement for an LLM-based agent. We introduce three core mechanisms. \textit{\textbf{a) Reflection}}: The agent analyzes a completed trajectory to generate a textual summary, storing in its context. This summary will helps for the next reasoning step. \textit{\textbf{b) Iterative Optimization:}} Within a single task, the agent generates an initial output, compares it against a defined standard or feedback from others, and repeatedly refines it in following reasoning steps, until $Q$ is met. \textit{\textbf{c) Interactive Learning}}: The agent interacts with a dynamic environment, where experiences can update reusable strategies and, when permitted, high-level goals. Goal adjustment is optional; experience reuse can also support a fixed task goal.}
    \Description{This image illustrates three core paradigms for self-improvement in an LLM-based agent: Reflection, Iterative Optimization, and Interactive Learning.}
    \label{fig:reflection}
\end{figure}

Self-improvement mechanisms encourage an agent to enhance its reasoning capabilities through introspection and learning from experience. These methods enable the agent to dynamically adapt its strategies based on its own experiences. As summarized in Fig. \ref{fig:reflection}, this internal optimization process can be understood through three complementary paradigms: \textit{reflection}, which involves learning from past trajectories; \textit{iterative optimization}, which focuses on refining outputs within a single reasoning cycle; and \textit{interactive learning}, which allows for adaptation through environmental feedback, including goal adjustment when permitted. 

\paragraph{Reflection}
Reflection enables an agent to perform post-hoc analysis on its past actions and outcomes to extract valuable lessons for future tasks. This process involves generating a summary of its reasoning process, identifying flaws or inefficiencies, and articulating insights in natural language \cite{guo2025mirror}. This process refers to \textit{line 5 in Alg. \ref{alg}}, where the action $a'_k$ is specifically assigned as reflection $a'_{\text{reflect}}$: 
\begin{equation}\label{eq2}
    \mathcal{C}_{k+1} = a'_{\text{reflect}}(\mathcal{C}_k, y_{k+1}, g, t)
\end{equation}
The scope of reflection can vary. For instance, the Reflexion framework \cite{shinn2023reflexion} guides agents to verbally reflect on task failures, storing these reflections in an episodic memory to refine plans in subsequent trials. Other approaches use external feedback to guide reflection. For instance, Logic-LM \cite{pan2023logiclm} uses symbolic-solver error messages to revise a generated formalization. This self-correction capability allows the agent to learn from its mistakes and continuously adapt its strategies without requiring external intervention or parameter updates.

\paragraph{Iterative Optimization}
\label{IO}
In contrast to the post-hoc nature of reflection, iterative optimization utilize a whole reasoning process to complete a pre-defined standard or constraint, which we denoted as $\mathcal{S}$.  This mechanism introduces two key modifications to the agent's fundamental operation described in Alg. \ref{alg}. 

First, the standard $\mathcal{S}$ is incorporated into the agent's initial context. This modification of the initialization step\textit{ (line 1 in Alg. \ref{alg}}) ensures that agent is aware of the optimization target from the outset:
\begin{equation}\label{eq3}
\mathcal{C}_0 \gets \text{Init}(P_U;\mathcal{R}) \quad \xrightarrow{\text{Iter. Opt.}} \quad \mathcal{C}_0 \gets \text{Init}(P_U, \mathcal{S})
\end{equation}
Equation \ref{eq3} shows that the context initialization is augmented to include not just the user's query $P_U$, but also the explicit standard $\mathcal{S}$ that the final output must satisfy.
Second, the agent's autonomy to decide when to stop is replaced by $\mathcal{S}$. The general termination condition $Q$ is now precisely defined by whether the current output $y$ satisfies the standard $\mathcal{S}$. This can be expressed as a formal redefinition of $Q$ \textit{(line 3 in Alg.\ref{alg})}:
\begin{equation}\label{eq4}
Q \quad \xrightarrow{\text{Iter. Opt.}} \quad Q \triangleq (y \vDash \mathcal{S})
\end{equation}
As stated in Equation \ref{eq4}, the termination condition $Q$ is now defined as the predicate checking if the current output $y$ satisfies the standard $\mathcal{S}$. Consequently, after each reasoning step that produces an output, the agent checks it against $\mathcal{S}$, entering an iterative loop of refinement until the condition is met.
This iterative loop is central to frameworks like Self-Refine \cite{madaan2023self}, where a single LLM acts as its own generator, critic, and refiner to improve its output without external training data. This approach is particularly effective for tasks requiring high precision, such as code generation \cite{chenteaching} or mathematical reasoning \cite{ahmaditeshnizi2024optimus}. However, it can be computationally intensive and risks converging on a suboptimal solution if the feedback mechanism is flawed or the search space is too complex \cite{pan2025multiagent}.

\paragraph{Interactive Learning}
Interactive learning allows an agent to adapt its strategies and, when permitted, alter its high-level goals $g$ based on continuous interaction with a dynamic environment. This paradigm moves beyond optimizing a fixed plan to enabling the agent to decide what to do next on a strategic level. This corresponds to an enhancement of the goal-updating mechanism \textit{(line 6 in Alg. \ref{alg})}, where the goal $g$ can be re-evaluated after a reasoning step:
\begin{equation}\label{eq5}
g_{k+1} \gets a_k( \{(\mathcal{C}_i, y_i)\}_{i=1}^{k+1}, g_k, t)
\end{equation}
Equation \ref{eq5} shows that the new goal $g_{k+1}$ is derived from the entire history of contexts and outputs $ \{(\mathcal{C}_i, y_i)\}_{i=1}^{k+1}$, the current goal $g_k$, and available tools $t$. 
Voyager \cite{wang23voyager} exemplify this, where an agent in Minecraft autonomously proposes new goals based on its discoveries, gradually building a complex skill tree without human intervention. 
Experience can also be reused while keeping the task goal fixed. ExpeL \cite{zhao2024expel} enables an agent to learn from trial-and-error experiences, extracting insights from successful and failed attempts and retrieving previous experiences for future tasks. 
Further systematizing this process, Learn-by-Interact \cite{hongjin2025learn} introduces a data-centric framework where an agent autonomously collects interaction data and then uses the trajectories as retrieved in-context demonstrations or as training data. Here, we focus on its demonstration-retrieval mechanism.
Through experience reuse and optional goal adaptation, the agent demonstrates a higher form of autonomy, allowing it to navigate complex, evolving environments in a truly adaptive manner \cite{zheng2025lifelong}.
\subsection{Tool-based Methods}
\label{sec:tool_based}
\begin{figure*}
    \centering
    \includegraphics[width=1\linewidth]{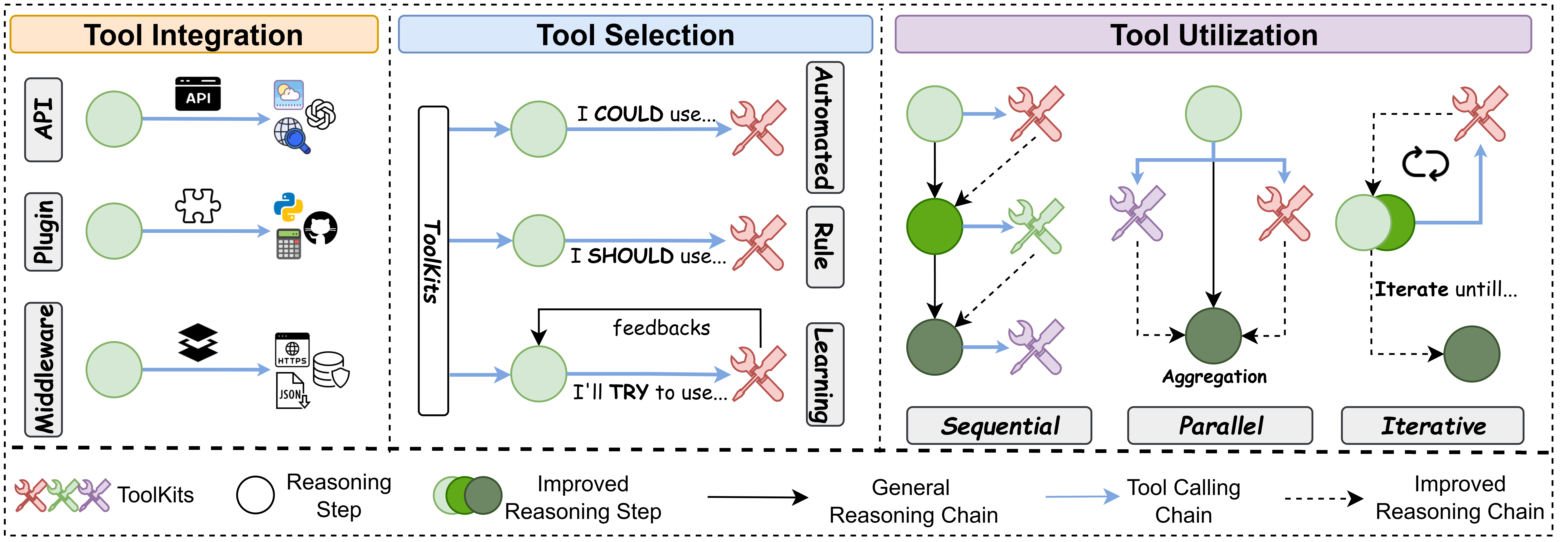}
    \caption{Tool-based reasoning frameworks of LLM-based agent. \textit{\textbf{a) Tool integration}} studies how to incorporate tools into the agent’s reasoning process; \textit{\textbf{b) Tool Selection}} addresses which tool from the toolkit $\mathcal{T}$ is most suitable for the current task or sub-problem;\textit{\textbf{ c) Tool Utilization}} concerns how to effectively operate the chosen tool to generate the desired output.}
    \Description{This image is a diagram that illustrates the three main components of a Tool-based reasoning framework for an LLM-based agent. The first component, Tool Integration, shows how tools can be incorporated into an agent's reasoning process through APIs, Plugins, or Middleware. The second, Tool Selection, describes the decision-making process for choosing the most suitable tool from a toolkit, using methods that are either automated, rule-based, or learning-based. The third, Tool Utilization, details how the selected tools are operated to generate a desired output, with examples of sequential, parallel, and iterative execution.}
    \label{fig:tool}
\end{figure*}

While the general agentic reasoning framework \textit{(Alg. \ref{alg})} conceptualizes tool use via a single entity $t$, this abstraction is insufficient for complex scenarios where reasoning is deeply intertwined with specific environmental capabilities. Here we expand this single entity $t$ into a comprehensive toolkit $\mathcal{T} = \{ t_1, t_2, ..., t_n\}$, where each $ t_i$ represents a distinct tool available to the agent. As illustrated in Fig. \ref{fig:tool}, we deconstruct the tool-based reasoning pipeline into three fundamental stages: \textit{\textbf{Tool Integration}}, \textit{\textbf{Tool Selection}}, and \textit{\textbf{Tool Utilization}}. Generally, the output from tool calling will be integrated into the context of agent by a specific action \textit{(line 5 in Alg. \ref{alg})}. These three steps together form the tool-based methods for complex multi-step reasoning, helping agents better exploit external resources to solve complex reasoning problems. 
\subsubsection{Tool Integration}

Before an agent selects and utilizes a tool, the tool must first be made accessible within the agent's operational environment. This architectural integration defines the interface and communication protocol between the agent and the tool \cite{ehtesham2025survey}. We categorize these integration patterns into three primary models: \textit{API-based Integration}, \textit{Plugin-based Integration}, and \textit{Middleware-based Integration}. APIs enable agents to easily interact with various tools without needing to understand their internal implementations; plugins dynamically extend the functionality of the agent system; while middleware focuses on aligning the interactions between the agent and the tools.

\paragraph{API-based Integration}
APIs \textit{(Application Programming Interfaces)} provide standard for integrating external tools \cite{zhang2025api}. APIs provide a stable, well-documented contract that allows an agent to interact with a tool (e.g., a web search engine \cite{chenmindsearch}) without needing to understand its internal implementation. The agent simply learns to formulate a request according to the API specification and parse the returned data.

Emerging Agent protocols such as MCP further develop the diversity of API tools. Under the corporation of service provider, agents now can easily use precise services such as map navigation to provide detailed information for the user \cite{hou2025model}. But such integration is subject to network latency, rate limits, and potential service outages. It also requires the agent to manage authentication and security credentials \cite{hou2025model}.

\paragraph{Plugin-based Integration}
Plugins are software components that are loaded and executed directly within the agent's own runtime environment. Unlike external API calls, plugins operate with lower latency and have deeper access to the agent's internal state \cite{lyu2023gitagentfacilitatingautonomousagent}. 

Retrieval-Augmented Generation (RAG) \cite{lewis2020retrieval} is a typical case of plugin-based integration. A vector database is integrated directly into the agent system, introducing domain-specific knowledge to the agent in the form of a tool call, thereby increasing the credibility of its answers \cite{gao2023retrieval}. Liu et al. \cite{liu2025drbioright} offers a more specific application of plugins. By integrating an interactive heatmap plugin and a scatter plot plugin, the agent system is enabled to dynamically process, analyze, and visualize domain-specific data during its reasoning process.
Thus, plugins offer a higher level of customization, extending the agentic framework's core abilities, but it may introduce complexity to the overall system \cite{hofstatter2023fid}.

\paragraph{Middleware-based Integration}
Middleware is a software layer situated between the Agent and tools \cite{gu2024middleware}. This layer acts as a universal adapter or an ``operating environment'' for the agent, abstracting away the complexities of direct tool interaction, shielding the LLM from environmental complexity \cite{gu2024middleware}. A middleware layer could manage API keys, standardize data formats across different tools, or provide a unified file system and execution environment for the agent \cite{xie2024osworld}. Therefore, middleware simplifies the agent's logic by offloading complex tasks, providing a consistent interface over a heterogeneous set of tools. Chen et al. \cite{cheninternet} further propose Internet of Agents, highlighting the advantages of middleware in complex reasoning processes. However, it adds another layer of abstraction that can complicate maintenance.

\subsubsection{Tool Selection}
Instead of generally using tool $t$ in each reasoning step \textit{(line 4 and 5 of Alg. \ref{alg})}, here we want to highlight the importance of the selecting action of tool within reasoning steps. Effective tool selection is pivotal when an agent is presented with a large and diverse toolkit $\mathcal{T}$. The challenge lies in accurately mapping the requirements of a given problem to the specific choice of a tool $t$, where $t \in \mathcal{T}$. Based on the degree of agent autonomy, we categorize tool selection strategies into three primary approaches: \textit{Autonomous Selection}, \textit{Rule-Based Selection}, and \textit{Learning-Based Selection}. 

\paragraph{Autonomous Selection}
This paradigm highlights the autonomy of agentic systems. The agent autonomously selects a tool based on its intrinsic reasoning capabilities, guided solely by the natural language descriptions of the available tools and the input query \cite{zhang2024mapgpt}. This process is often framed as a zero-shot reasoning task, where the agent must ``think'' to connect the problem to the right tool without explicit rules \cite{yao2023react}.

Just like a general reasoning step, the tool selection step will let agent reason, reflect, or even use tools to decide which tool $t_{k+1}$ is suited for current condition, this tool $t_{k+1}$ can be regarded as the output $y$ within this reasoning step:
\begin{equation}\label{eq6}
y_{k+1} \gets t_{k+1} = a_k(\mathcal{C}_k, g, \mathcal{T}), \quad a_k \in \mathcal{A}
\end{equation}
Following \textit{line 5 in Alg. \ref{alg}}, the selected tool $t_{k+1}$ is updated into the current context window, allowing the agent to use it in subsequent reasoning steps. This selection process may sometimes be repeated multiple times to gradually filter for the best tools from a large toolset \cite{li2025chemhts}. Since this method requires no task-specific examples or fine-tuning, it enables the agent to dynamically adapt to novel combinations of tools, tasks and scenarios. However, its performance is highly dependent on the quality of tool descriptions and the agent's inherent reasoning capacity, which challenges the robustness and efficiency of the agent system.

\paragraph{Rule-Based Selection}
This approach governs agent's tool selection through a set of predefined, explicit rules $\mathcal{R}$ that map specific tasks, intents, or states to designated tools \cite{luo2025intention}. These rules provide a deterministic and reliable mechanism for tool choice. The selection process is thus conditioned on these rules: 
\begin{equation}\label{eq7}
    t_{k+1} = a_k(\mathcal{C}_k, g, \mathcal{T}, \mathcal{R}), \quad a_k \in \mathcal{A}
\end{equation}
The rules in $\mathcal{R}$ can be implemented in various forms, from simple keyword matching \cite{liu2024agentbench} to structured formats like process description language (PDL) \cite{zhou2024isr}.

The main benefit of rule-based selection is its high reliability for well-defined tasks. It ensures that the agent consistently uses the correct tool for a known situation, minimizing errors \cite{li2025review}. However, manually crafting and maintaining a comprehensive set of rules is labor-intensive and scales poorly as the number of tools and the complexity of tasks grow. It struggles with unforeseen problems that do not match any existing rules, forcing a default failure or a fallback to a different selection mechanism.

\paragraph{Learning-Based Selection}
Learning-based selection can involve parameter training, as in Toolformer~\cite{schick2023toolformer}, or contextual adaptation during inference. Our framework-level focus here is the latter: an explicit, online process where the agent refines its tool selection strategy through experience. This adaptation occurs through a cycle of action, feedback, and reflection, improving its concurrent tool actions. As illustrated in Fig.~\ref{fig:tool}, the agent attempts a tool for task, receives feedback on its performance (e.g., from execution results, or human guidance), and then explicitly reflects on this outcome to update its context $\mathcal{C}$ for subsequent steps:
\begin{equation}\label{eq8}
\mathcal{C}_{k+1} = a'_{\text{reflect}}(\mathcal{C}_k, y_{k+1}, g, t)
\end{equation}
This reflective step allows the agent to learn from its own context by storing experiences of successful tool-task pairings or by generating explicit strategies to avoid repeating past mistakes \cite{qin2024tool}. This approach enables the agent to adapt to novel scenarios and user preferences without requiring model retraining. Learn-by-interact~\cite{hongjin2025learn} synthesizes trajectories of agent-environment interactions based on documentations; its in-context variant retrieves demonstrations using the current observation and interaction history. Agent Workflow Memory~\cite{wang2025awm} similarly supports reuse by inducing workflows from prior trajectories. However, this requires a sound feedback logic, and the exploration process can be costly.

\subsubsection{Tool Utilization}
Following the previous section, this section focus on how to make the best use of the selected tools \cite{masterman2024landscape}. Here we divide tool utilization into three modes: sequential use, parallel use, and iterative use. \textit{Sequential use} involves invoking multiple tools in a predetermined order, \textit{parallel use} focuses on the breadth of tool calls within the same reasoning step, while \textit{iterative use} aims at achieving the optimal task solution within certain limits through repeated cycles. 

\paragraph{Sequential Utilization}
In this mode, the agent invokes tools in a sequence, where the output of one tool often serves as the input for the next, forming a clear tool-chain \cite{li2025review}. This is well-suited for tasks that can be decomposed into a linear workflow. The results of tool calling are integrated into the current context, influencing the next calling \cite{basu2024nestful}. CRITIC \cite{goucritic} improve its output through a sequential use of external tools, including search engine and code interpreter. MCP-Zero \cite{fei2025mcp} further promote tool discovery based on the tool chain, where agent sequentially use different tool to solve complex problems. Its primary benefit is simplicity and predictability, making workflows easy to design, analyze and debug. But it can be inefficient for tasks with independent sub-problems and susceptible to cascading failures, where an error in an early step halts the entire chain.

\paragraph{Parallel Utilization}
To enhance efficiency, this mode involves invoking multiple tools concurrently within a single reasoning step. The Agent invokes multiple tools simultaneously to achieve synchronous processing of multidimensional information. For a selected tool set $\mathcal{T}^{'} = \{t_1^{'}, t_2^{'}, ..., t_m^{'}\}$ in any middle reasoning step $k$, the agent will generate a group of results in parallel using each tool within $\mathcal{T'}$. That is, for \textit{line 4 in Alg.\ref{alg}}, the output will becomes a set of output:

\begin{equation}\label{eq9}
\begin{gathered}
  \mathcal{Y}_{k+1} = \{ y_{k+1}^1, y_{k+1}^2, \dots, y_{k+1}^m \} \\
  \text{where } y^i_{k+1} = a_k(\mathcal{C}_k, g, t'_i), \quad a_k \in \mathcal{A}, \quad t'_i \in \mathcal{T'}
\end{gathered}
\end{equation}
After that, the update of context will further consider this output set $ \mathcal{Y}_{k+1}$\cite{zhu2025dividethenaggregateefficienttoollearning}, instead of a single output like before:
\begin{equation}\label{eq10}
    \mathcal{C}_{k+1} = a'_k(\mathcal{C}_k, \mathcal{Y}_{k+1}, g, t)
\end{equation}
The key advantage is a significant reduction in latency, as multiple time-consuming tool utilization can be executed at once. It is also efficient to explore several suitable tools simultaneously. 
For example, LLM Compiler \cite{kim2024llm} efficiently orchestrates multiple function calls by executing them in parallel during intermediate reasoning steps, while LLM-Tool Compiler \cite{singh2024llm} achieves tool parallelization by selectively fusing tools with similar functionalities. But such techniques may also introduce the challenge of aggregating potentially conflicting information from diverse outputs.

\paragraph{Iterative Utilization}
Iterative utilization involves a micro-level loop where an agent repeatedly interacts with a tool to achieve a fine-grained objective within a single step of the broader reasoning process \cite{madaan2023self}. This contrasts with macro-level iterative optimization of the entire solution in \S\ref{IO}. The focus here is on perfecting a single tool-use instance. A prime example is an agent using a code interpreter: if the first execution fails, the agent inspects the error, refines the code, and re-executes it until it runs successfully, all before moving to the next macro reasoning step \cite{he2024cocost}. 
This method enhances the robustness of tool execution, but may increase the latency of a single reasoning step and carries the risk of getting stuck in unproductive loops \cite{pan2025multiagent}. This necessitates carefully designed termination conditions or reflection mechanisms \cite{shinn2023reflexion}.

\subsection{Multi-agent Methods}
\label{sec:multi_agent}
While single-agent frameworks demonstrate considerable capabilities, they inherently face limitations when confronted with tasks demanding diverse expertise or complex problem decomposition. Multi-agent systems (MAS) emerge as a natural solution, leveraging the collective intelligence of multiple agents to tackle these challenges. The central principle of MAS is to ``divide and conquer'', but its core challenge lies in achieving effective coordination \cite{tran2025multi}. This challenge bifurcates into two fundamental questions: (1) \textit{How should the agents be organized?} This pertains to the system's organizational architecture, which dictates the patterns of control and information flow. (2) \textit{How should the agents interact with others?} This relates to the individual interaction protocols, which define how agents align their goals and behaviors.

\begin{figure*}
    \centering
    \includegraphics[width=1\linewidth]{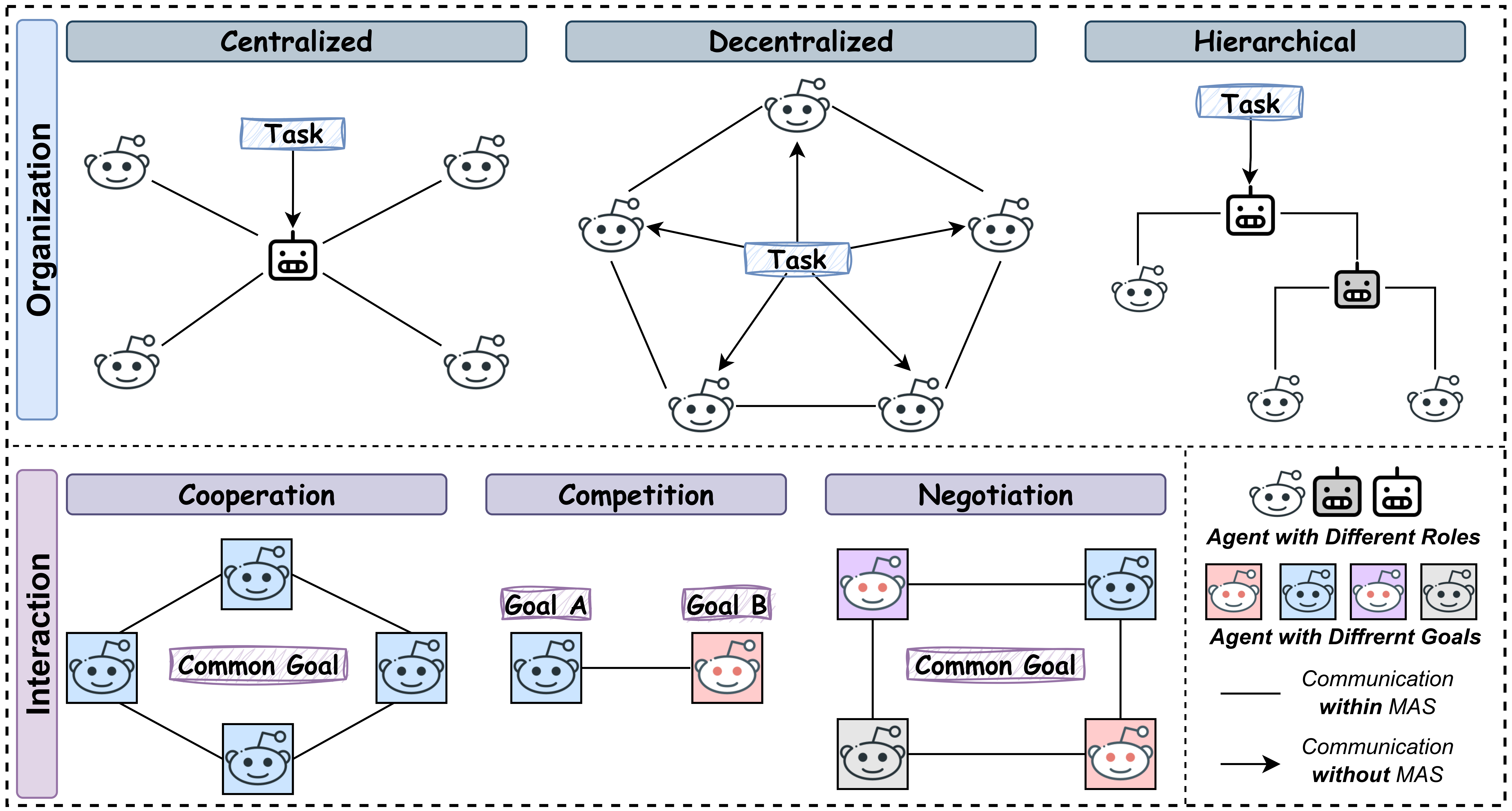}
    \caption{A taxonomy of Multi-agent reasoning frameworks, categorized in \textbf{\textit{a) Organizational Architectures}}: we summarize three paradigms of multi-agent frameworks to explain how such system can be organized to solve different kinds of challenges; \textit{\textbf{b) Individual Interactions}}: we demonstrated three types of interaction between different agents within one multi-agent system.}  
    \Description{This image is a taxonomy of multi-agent reasoning frameworks, divided into two primary categories: Organizational Architectures and Individual Interactions.}
    \label{fig:MAS}
\end{figure*}

As illustrated in Figure \ref{fig:MAS}, we analyze the multi-agent reasoning frameworks along these two axes. \textbf{\textit{Organizational architectures}} include centralized, distributed, and hierarchical forms, which determine the structural backbone of the system; while \textbf{\textit{individual interactions}} involve cooperation, competition, and negotiation, governing the dynamics between agents as they pursue their objectives. 

To formally discuss these paradigms, we represent a multi-agent system as a set of agents $\mathcal{M} = \{A^{r_1}, A^{r_2}, ..., A^{r_n}\}$, where $r_i$ denotes the specialized role of an agent $A^{r_i}$. While each agent adheres to the general reasoning loop outlined in \textit{Alg. \ref{alg}}, its behavior is individuated by its unique role, goal $g^{r_i}$, available actions $a^{r_i}$ and tool $t^{r_i}$. They also maintain different individual context $\mathcal{C}^{r_i}$. It is the distinctiveness of each agent's context and role that drives the heterogeneity in their reasoning, ultimately shaping the system's collective output.

\subsubsection{Organizational Architecture}
The organizational architecture defines the macro-level structure for coordination and control, which is often instantiated by assigning a specific role $r_i$ to each agent's initial context $\mathcal{C}_0^{r_i}$:
\begin{equation}\label{eq11}
        \mathcal{C}_0 \gets \text{Init}(P_U;\mathcal{R}) \Rightarrow \mathcal{C}_0^{r_i} \gets \text{Init}(P_U,r_i)
\end{equation}
Extending line~5 of Algorithm~\ref{alg}, an agent may incorporate outputs from other agents alongside its previous context $\mathcal{C}_k^{r_i}$. For a synchronous round, $\mathcal{Y}_{k+1}$ collects the newly produced outputs, with the organizational architecture determining which of them are accessible to the update action:
\begin{equation} \label{eq12}
\begin{gathered}
\mathcal{C}_{k+1}^{r_i} = a'^{r_i}_k(\mathcal{C}_k^{r_i}, \mathcal{Y}_{k+1}, g^{r_i}, t^{r_i})\\ \text{where } \mathcal{Y}_{k+1} = \{y_{k+1}^{r_1}, y_{k+1}^{r_2}, ..., y_{k+1}^{r_n}\}
\end{gathered}
\end{equation}
For synchronous rounds, $\mathcal{Y}_{k+1}$ represents the collective outputs of all the $n$ agents at reasoning step $k+1$. The specific subset of $\mathcal{Y}_{k+1}$ that an agent $A^{r_i}$ considers is determined by different organizational architecture. We further decompose the organization of multi-agent systems into three distinct paradigms: centralized, decentralized, and hierarchical. A \textit{centralized architecture} is suitable for scenarios requiring global optimization and strict control; a \textit{distributed architecture} offers greater autonomy for each agent; while a \textit{hierarchical architecture} is appropriate for tasks with clear goals and well-defined processes. These paradigms can be statically or dynamically integrated, reaching a flexible organization structure in specific scenarios \cite{liu2025advances}. Asynchronous systems use the messages available when each agent updates.

\paragraph{Centralized}
In general, a centralized architecture sets a central agent $A^{\text{cen}}$ to manage and coordinate the reasoning activities for other agents, $A^{r_i}$ where $ r_i \neq \text{cen}$ \cite{ghafarollahi2024protagents}. The roles differ, however. This central agent  $A^{\text{cen}}$ typically performs global planning, task decomposition, and result synthesis, requiring it to process the outputs from all other agents, as in Eq. \ref{eq12}. Subordinate agents, however, may only need to consider instructions from the manager, simplifying their context updates. This architecture supports coordination toward a global objective \cite{ghafarollahi2024protagents}. However, it introduces a potential performance bottleneck and a single point of failure at the central node \cite{pan2025multiagent}.

\paragraph{Decentralized}
In a decentralized architecture, there is no central authority. Each agent possesses equal status and makes decisions based on local information and direct peer-to-peer communication \cite{yang2025agentnet}. Consequently, the context update for every agent typically follows the general form in Eq. \ref{eq12}, where each agent must process the outputs of all its neighbors, or all other agents in a fully connected system. This discussion-like process fosters emergent collaboration and enhances system robustness and fault tolerance, as the failure of one agent does not cripple the entire system \cite{yang2025agentnet}. However, it may reduce the overall efficiency of resource utilization \cite{pan2025multiagent}.

\paragraph{Hierarchical}
A hierarchical architecture organizes agents into a structured tree or pyramid, decomposing a complex task into sub-problems at different levels of abstraction. As illustrated in MetaGPT \cite{hongmetagpt}, agents at higher levels are responsible for strategic planning and delegate tasks to lower-level agents, which execute more specific sub-tasks. Information typically flows vertically: instructions pass down from upper to lower levels, and results are passed back up. This structure excels at solving well-defined problems that can be clearly decomposed, promoting efficiency and consistency \cite{chen2025mdteamgpt}. However, such architectures can be rigid and may stifle the flexibility and creativity of individual agents. 

\subsubsection{Individual Interaction}
The interaction protocol governs how agents pursue their goals in response to others, directly influencing the system's emergent behavior. When the setting permits goal adaptation, this introduces a dynamic update to an agent's goal $g^{r_i} $, corresponding to the optional goal update in the basic reasoning loop \textit{(line 6 in Alg. \ref{alg})}. We further categorize these interactions as cooperation, competition, and negotiation. 
\textit{Cooperation} emphasizes maximizing collective interests, \textit{competition} focuses on maximizing individual interests, and \textit{negotiation} represents a compromise between the two. These three different paradigms can also be further combined to achieve specific reasoning objectives. 

\paragraph{Cooperation}
In cooperation mechanism, the primary objective for agents is to maximize collective interests. A common goal $\mathcal{G}$ is established to guide knowledge sharing and collaborative planning \cite{tran2025multi}. This goal can be predefined in the system prompt or dynamically formed during different reasoning steps. 

When goal adaptation is enabled, an agent can update its goal by reflecting on its performance and aligns its individual goal $g^{r_i}$ with the common goal $\mathcal{G}$. The updated goal will further influence the \textit{next} reasoning step:
\begin{equation}\label{eq13}
 g^{r_i} \gets a_{\text{reflect}}^{r_i}(\mathcal{C}_{k+1}^{r_i}, g^{r_i}, \mathcal{G}, t^{r_i}) 
\end{equation}
Therefore, the achievement of individual agentic goals often promotes the goals of other agents and the entire system \cite{guo2024large}. 

\paragraph{Competition}
In competitive interactions, agents pursue their individual goals, which are often in conflict. The objective is to maximize individual benefit, which may involve outperforming or strategically undermining opponents \cite{fan2024can}. An agent must not only advance its own agenda but also infer and counter the intentions of others based on their observable outputs $\mathcal{Y}_{k+1}$. When goals can change, the update process can therefore become adversarial:
\begin{equation}\label{eq14}
\begin{gathered}
     g^{r_i} \gets a_{\text{reflect}}^{r_i}(\mathcal{C}_{k+1}^{r_i}, g^{r_i}, \mathcal{Y}_{k+1}, t^{r_i}),
      \text{where } \mathcal{Y}_{k+1} = \{y_{k+1}^{r_1}, y_{k+1}^{r_2}, ..., y_{k+1}^{r_n}\}
\end{gathered}
\end{equation}
A related interaction pattern is exemplified by frameworks that use multi-agent debate, such as MAD \cite{liang2024encouraging}, where agents take on adversarial ``debater'' roles to challenge assumptions and uncover flaws in reasoning. Such adversarial interactions can significantly improve the robustness and quality of the final output by forcing a thorough exploration of the problem space \cite{islam2024mapcoder}. Opposing arguments can still serve a shared task goal. Table~\ref{tab:general_frame} uses \textit{Deb} for this debate protocol and \textit{Coo} for cooperation.

\paragraph{Negotiation}
Negotiation is a hybrid interaction that balances cooperation and competition. It enables agents with conflicting interests to reach a mutually acceptable consensus through communication and compromise \cite{gottweis2025towards}. During negotiation, agents exchange proposals. If they adapt their local goals, the update can depend on the common objective $\mathcal{G}$ and others' proposals in $\mathcal{Y}_{k+1}$, as illustrated in Figure \ref{fig:MAS}:
\begin{equation}\label{eq15}
\begin{gathered}
     g^{r_i} \gets a_{\text{reflect}}^{r_i}(\mathcal{C}_{k+1}^{r_i}, g^{r_i}, \mathcal{Y}_{k+1}, \mathcal{G}, t^{r_i}) 
     \\ \text{where } \mathcal{Y}_{k+1} = \{y_{k+1}^{r_1}, y_{k+1}^{r_2}, ..., y_{k+1}^{r_n}\}
\end{gathered}
\end{equation}
This process compels agents to weigh their own objectives against collective constraints and the perspectives of others. For instance, ChatEval \cite{chanchateval} utilizes a consensus-oriented discussion among multiple ``referee'' agents to autonomously evaluate the quality of AI-generated text, reaching a joint judgment. This approach is particularly effective for complex decision-making tasks where there is no single correct answer, but rather a spectrum of acceptable solutions.

\begin{table*}[!t]
\centering
\resizebox{\surveycolumnwidth}{!}{%
\begin{minipage}{\surveycolumnwidth}
\centering
\caption{A comprehensive comparison of mainstream agentic reasoning frameworks, illustrating how methods from our taxonomy are organized within each work, alongside their inspiration, evaluation, and code. The legend for the abbreviations is as follows.
\textbf{PE (Prompt Engineering)}: Role (role-playing), Task (task description).
\textbf{SI (Self-Improvement)}: R.F (reflection), I.O (iterative optimization), I.L (interactive learning).
\textbf{Tools}: T.I (tool integration); T.S (tool selection: `rule' for rule-based, `auto' for autonomous); T.U (tool utilization: `Seq' for sequential, `Iter' for iterative).
\textbf{Multi-agent}: M.O (organization: `Dec' for decentralized, `Cen' for centralized, `Hier' for hierarchical); M.I (interaction: `Deb' for debate, `Coo' for cooperation).
\\ $^*$Mix: The framework employs more than one organization method.
\\ \textbf{$\dagger$}Prompt engineering techniques are widely used, so we list only the most representative sub-methods employed in each work.
\\ \textbf{$\ddagger$}This column lists the primary inspirations (theories or prior works) stated in each paper.}
\label{tab:general_frame}
\resizebox{\textwidth}{!}{%
\renewcommand{\arraystretch}{1.3}
\setlength{\tabcolsep}{3pt}
\begin{tabular}{c|cc|ccc|cc|c|c|c}
\Xhline{1.4pt}
\rowcolor{lightblue}
\cellcolor{lightblue}&
\multicolumn{2}{c|}{\cellcolor{lightblue}\textbf{Single-agent}}&
\multicolumn{3}{c|}{\cellcolor{lightblue}\textbf{Tool-based}}&
\multicolumn{2}{c|}{\cellcolor{lightblue}\textbf{Multi-agent}}&
\cellcolor{lightblue}&
\cellcolor{lightblue}&
\cellcolor{lightblue}\\ \cline{2-8}
\rowcolor{lightgray}
\multirow{-2}{*}{\cellcolor{lightblue}\textbf{Work}}&
PE\textbf{$^\dagger$}&
SI&
T.I&
T.S&
T.U&
M.O&
M.I&
\multirow{-2}{*}{\cellcolor{lightblue}\textbf{Inspiration}\textbf{$^\ddagger$}}&
\multirow{-2}{*}{\cellcolor{lightblue}\textbf{Datasets}}&
\multirow{-2}{*}{\cellcolor{lightblue}\textbf{Code}}\\ \hline
Du et al.~\raisebox{0pt}[\height][\depth]{\cite{du2024improving}}&Role&I.O.&-&-&-&Dec.&Deb.&\begin{tabular}[c]{@{}c@{}}Society of Mind~\raisebox{0pt}[\height][\depth]{\cite{minsky1986society}}\end{tabular}&\begin{tabular}[c]{@{}c@{}}GSM8k~\raisebox{0pt}[\height][\depth]{\cite{cobbe2021training}},BigBench~\raisebox{0pt}[\height][\depth]{\cite{srivastava2023beyond}},\\MMLU~\raisebox{0pt}[\height][\depth]{\cite{hendrycksmeasuring}}\end{tabular}&\raisebox{0pt}[\height][\depth]{\href{https://github.com/composable-models/llm_multiagent_debate}{\faGithub}}\\\hline
MAD~\raisebox{0pt}[\height][\depth]{\cite{liang2024encouraging}}&Role/Task&-&-&-&-&Dec.&Deb.&\begin{tabular}[c]{@{}c@{}}Degeneration-of\\-Thought~\raisebox{0pt}[\height][\depth]{\cite{bortolotti2011does,keestra2017metacognition}}\end{tabular}&\begin{tabular}[c]{@{}c@{}}Kong et al.~\raisebox{0pt}[\height][\depth]{\cite{kong2022eliciting}},Website$^{**}$,\end{tabular}&\raisebox{0pt}[\height][\depth]{\href{https://github.com/Skytliang/Multi-Agents-Debate}{\faGithub}}\\\hline
SPP~\raisebox{0pt}[\height][\depth]{\cite{wang2024unleashing}}&Role&I.O.&-&-&-&Cen.&Coo./Deb.&\begin{tabular}[c]{@{}c@{}}Pretend play~\raisebox{0pt}[\height][\depth]{\cite{piaget2013construction,pellegrini2009role}}\end{tabular}&TriviaQA~\raisebox{0pt}[\height][\depth]{\cite{joshi2017triviaqa}},BigBench~\raisebox{0pt}[\height][\depth]{\cite{srivastava2023beyond}}&\raisebox{0pt}[\height][\depth]{\href{https://github.com/MikeWangWZHL/Solo-Performance-Prompting}{\faGithub}}\\\hline
AutoGen~\raisebox{0pt}[\height][\depth]{\cite{wu2024autogen}}&Task&RF.&API&Rule&Seq&Mix$^*$&Coo.&\begin{tabular}[c]{@{}c@{}}Society of Mind~\raisebox{0pt}[\height][\depth]{\cite{minsky1986society}}\end{tabular}&\begin{tabular}[c]{@{}c@{}}MATH~\raisebox{0pt}[\height][\depth]{\cite{hendrycks2measuring}},Kwiatkowski et al.~\raisebox{0pt}[\height][\depth]{\cite{kwiatkowski2019natural}},\\Adlakha et al.~\raisebox{0pt}[\height][\depth]{\cite{adlakha2024evaluating}},ALFworld~\raisebox{0pt}[\height][\depth]{\cite{shridharalfworld}}\end{tabular}&\raisebox{0pt}[\height][\depth]{\href{https://github.com/microsoft/autogen}{\faGithub}}\\\hline
AgentVerse~\raisebox{0pt}[\height][\depth]{\cite{chen2024agentverse}}&Role&I.L.&\begin{tabular}[c]{@{}c@{}}API/\\Plugin$^{\dagger\dagger}$\end{tabular}&Auto&Seq&Dec.&Coo.&Markov decision process&\begin{tabular}[c]{@{}c@{}}FED~\raisebox{0pt}[\height][\depth]{\cite{mehri2020unsupervised}},Commongen-Challenge~\raisebox{0pt}[\height][\depth]{\cite{madaan2023self}},\\MGSM~\raisebox{0pt}[\height][\depth]{\cite{shilanguage}},BigBench~\raisebox{0pt}[\height][\depth]{\cite{srivastava2023beyond}},\\HumanEval~\raisebox{0pt}[\height][\depth]{\cite{chen2021evaluating}}\end{tabular}&\raisebox{0pt}[\height][\depth]{\href{https://github.com/OpenBMB/AgentVerse}{\faGithub}}\\\hline
AutoAgents~\raisebox{0pt}[\height][\depth]{\cite{chen2024autoagents}}&Role&I.O.&API&Auto&Seq&Mix$^*$&Coo.&\begin{tabular}[c]{@{}c@{}}ReAct~\raisebox{0pt}[\height][\depth]{\cite{yao2023react}},AutoGPT$^{\ddagger\ddagger}$\end{tabular}&\begin{tabular}[c]{@{}c@{}}MT-Bench~\raisebox{0pt}[\height][\depth]{\cite{zheng2023judging}},FairEvals~\raisebox{0pt}[\height][\depth]{\cite{wang2024large}},\\HumanEval~\raisebox{0pt}[\height][\depth]{\cite{chen2021evaluating}},TriviaQA~\raisebox{0pt}[\height][\depth]{\cite{joshi2017triviaqa}}\end{tabular}&\raisebox{0pt}[\height][\depth]{\href{https://github.com/Link-AGI/AutoAgents}{\faGithub}}\\\hline
CAMEL~\raisebox{0pt}[\height][\depth]{\cite{li2023camel}}&Role/Task&I.O.&API&Auto&Seq&Dec.&Coo.&\begin{tabular}[c]{@{}c@{}}Society of Mind~\raisebox{0pt}[\height][\depth]{\cite{minsky1986society}}\end{tabular}&HumanEval~\raisebox{0pt}[\height][\depth]{\cite{chen2021evaluating}},HumanEval+~\raisebox{0pt}[\height][\depth]{\cite{liu2023your}}&\raisebox{0pt}[\height][\depth]{\href{https://github.com/camel-ai/camel}{\faGithub}}\\\hline
ChatDev~\raisebox{0pt}[\height][\depth]{\cite{qian2024chatdev}}&Role&I.O.&API&Auto/Rule&Seq&Hier.&Coo.&LLM Hallucination~\raisebox{0pt}[\height][\depth]{\cite{zhang2025siren}}&SRDD~\raisebox{0pt}[\height][\depth]{\cite{qian2024chatdev}}&\raisebox{0pt}[\height][\depth]{\href{https://github.com/OpenBMB/ChatDev}{\faGithub}}\\\hline
MetaGPT~\raisebox{0pt}[\height][\depth]{\cite{hongmetagpt}}&Role&I.O./I.L.&API&Auto&Iter.&Hier.&Coo.&\begin{tabular}[c]{@{}c@{}}ReAct~\raisebox{0pt}[\height][\depth]{\cite{yao2023react}},SOPs~\raisebox{0pt}[\height][\depth]{\cite{belbin2012team,manifestomanifesto,demarco2013peopleware,wooldridge1998pitfalls}}\end{tabular}&\begin{tabular}[c]{@{}c@{}}HumanEval~\raisebox{0pt}[\height][\depth]{\cite{chen2021evaluating}},MBPP~\raisebox{0pt}[\height][\depth]{\cite{austin2021program}},\\SoftwareDev~\raisebox{0pt}[\height][\depth]{\cite{hongmetagpt}}\end{tabular}&\raisebox{0pt}[\height][\depth]{\href{https://github.com/FoundationAgents/MetaGPT}{\faGithub}}\\
\Xhline{1.4pt}
\multicolumn{8}{l}{$^{**}$\url{https://www.geeksforgeeks.org/puzzles/}} \\
\multicolumn{8}{l}{$^{\dagger\dagger}$\url{https://github.com/OpenBMB/BMTools}} \\
\multicolumn{8}{l}{$^{\ddagger\ddagger}$\url{https://github.com/Significant-Gravitas/Auto-GPT}} \\
\end{tabular}
}
\end{minipage}%
}
\end{table*}

\subsection{Synthesis of Reasoning Methods}
\label{sec:emerging}
The methods discussed above can be combined within a single framework, linking individual reasoning to tool use and interaction among agents. For instance, an agent in a multi-agent system (\S\ref{sec:multi_agent}) can refine its reasoning through prompt engineering and self-reflection (\S\ref{sec:single_agent}), while invoking external tools (\S\ref{sec:tool_based}) to execute steps based on its designated role. Table~\ref{tab:general_frame} consolidates representative frameworks, detailing how they integrate methods from the different categories, along with their design motivations and evaluation datasets. AutoGen~\cite{wu2024autogen}, for example, combines task-description prompting and reflection at the single-agent level with rule-based sequential API tool calls at the tool level and a mixed multi-agent organization with cooperative interaction; a hybrid system is therefore classified by the mechanisms it composes at each level rather than by a single cell, and Table~\ref{tab:general_frame} records such combinations, with ``Mix'' marking more than one organizational pattern.

These compositions also support capabilities that cut across the three levels. \textbf{Long-term memory} connects reflection with persistent storage and retrieval, allowing experience to inform later reasoning~\cite{zhang2024survey}. The retained information may take the form of episodic feedback, extracted insights or linked notes, depending on how it will be revised and reused~\cite{shinn2023reflexion,zhao2024expel,xu2025amem,chhikara2025mem0}. Knowledge management similarly connects evidence acquisition to intermediate reasoning~\cite{singh2025agenticrag}: retrieval can supply context within one agent's loop, while specialized agents can gather and integrate complementary sources before passing evidence onward~\cite{lewis2020retrieval,lee2025rag,kang2024chatmof}.

\textbf{Continual learning} at the framework level extends reuse from observations to procedures. Interaction trajectories can inform later decisions, and recurring action sequences can become retrievable workflows, tools or skills that evolve across tasks~\cite{hongjin2025learn,wang23voyager,wang2025awm,zhou2026mementoskills}. Such adaptation changes external resources without necessarily updating the model's parameters; these two forms of adaptation have different mechanisms and limitations~\cite{li2025memos}. Context engineering connects these retained resources to subsequent reasoning by determining which instructions, observations and messages enter each step~\cite{mei2025survey}. Compression helps useful experience fit within the context budget, while incremental curation uses reflection on new experience to revise the material retained for future tasks~\cite{kang2026acon,zhang2026ace}.

The three levels describe how reasoning is organized, while these capabilities describe functions implemented within and across those levels. Algorithm~\ref{alg} makes their interfaces explicit: retained resources enter initialization, tool actions access external evidence, and context-update operators incorporate observations for subsequent reasoning. Reusing memory or revising a skill need not change the current goal. This structural account does not specify the internal policies of memory management or model-training procedures. Chapter~\ref{sec:scenarios} further discusses how these combinations support different application requirements, including task-oriented and self-improving systems in Section~\ref{sec:general_agents}; Table~\ref{tab:general_frame} provides the common methodological view.

\subsection{Discussion}
\label{sec:discussion}

In this chapter, we introduced a three-level, compositional taxonomy to demonstrate how methods from each level enhance various facets of an agentic framework's reasoning process. This classification is grounded in a unified formal language and a general reasoning algorithm (Alg. \ref{alg}). We contend that by combining methods across these levels, the capability boundaries and collaborative patterns of agent systems can be significantly expanded. 

Furthermore, while our taxonomy provides a comprehensive summary of the logical structures and collaborative patterns at the framework level, we acknowledge that researchers often incorporate optimization techniques like Supervised Fine-Tuning \cite{gao2025survey} and Reinforcement Learning \cite{ghasemi2024comprehensive} to achieve superior performance \cite{ke2025survey}. To maintain a clear focus on the foundational nature of these reasoning frameworks, we exclude these technical details from our classification. In the subsequent chapter, we will further showcase the value of these frameworks by examining their applications in specific scenarios.

%% file: chapters/04_scenarios.tex
\section{Scenarios}
\label{sec:scenarios}
\input{fig/scenario}
Building upon the previous chapter's foundational concepts, this chapter offers a panoramic view of agentic reasoning capabilities across diverse application scenarios. Our primary goal is to systematically compare and contrast the similarities and differences among these frameworks. We also conduct a comprehensive collection of the evaluation metrics, methodologies, and datasets across these domains. We categorize the application scenarios into \textit{\textbf{scientific research, healthcare, software engineering, social \& economic simulations, and general-purpose agents}}, as illustrated in Fig. \ref{fig:scenarios}. Two conventions apply throughout this chapter. First, the benchmark tables (Tables~\ref{tab:codegen}, \ref{tab:coderep}, \ref{tab:general_agents} and~\ref{tab:self_evolving}) compile results reported by different papers, which use different task models, prompts, tool budgets, benchmark versions and subsets. The tables identify task models, reporting sources and the benchmark versions, metrics or special settings specified in their headers and notes. Other details, including prompts and resource budgets, remain in the cited reports. Model grouping and a shared source aid interpretation but do not establish matched conditions: direct comparisons require the controlled configurations documented by the reporting study. Across sources, the numbers indicate reported capabilities, not a ranking of frameworks. Second, the scenario discussions relate task requirements to framework choices; Table~\ref{tab:cross_scenario} compares the constraints and choices, and \S\ref{sec:scenario_synthesis} draws out considerations that transfer across domains.

\subsection{Scientific Research}
\label{sec:sci}

Agent systems are increasingly becoming a cornerstone for automating various stages of scientific inquiry. Through the implementation of well-designed reasoning pipelines, these agents enhance the efficiency of the entire scientific workflow. We systematically review the design of agent frameworks aimed at accelerating research, with a focus on their applications in a range of disciplines including \textit{mathematics, astrophysics, geoscience, biochemistry, materials science}, as well as \textit{general scientific research} frameworks. 
\subsubsection{Math}

By leveraging the synergistic combination of their constituent components, agent systems go beyond traditional reasoning methods to achieve remarkable results in specific mathematical domains, including optimization and proof generation.

As an early work, LLM4ED \cite{du2024llm4ed} utilizes a symbolic library to aid in equation discovery, where the LLM iteratively proposes and refines novel equations based on natural language instructions to outperform purely text-based methods. Subsequent research has gravitated towards multi-agent systems that employ structured collaboration. In the realm of mathematical optimization, Optimus \cite{ahmaditeshnizi2024optimus} leverages a multi-agent system to autonomously manage the entire pipeline for mixed-integer linear programming, including task assignment, modeling, and evaluation, while using a central graph to track dependencies for iterative refinement. A similar hierarchical organization is also used in computational fluid dynamics (CFD) field by MetaOpenFOAM \cite{chen2024metaopenfoam}, where role-based agents collaboratively handle simulation design, setup, and review in a iterative closed loop. The concurrent work MetaOpenFOAM 2.0 \cite{chen2025metaopenfoam} further enhances robustness by introducing Chain of Thought (CoT) \cite{wei2022chain} and iterative CoT strategies for complex subtask decomposition. This orchestration method also proves effective in OptimAI \cite{thind2025optimai}, which solves natural language optimization problems by hierarchically decomposing user queries and automating the full cycle of model formulation, coding, and debugging through iterative feedback.

Other related works focus on formal theorem proving. MA-LoT \cite{wang2025ma} employs multi-agent collaboration to decouple the natural language cognitive task of proof generation from subsequent error analysis. In its framework, one agent generates a complete proof, while another corrects it, establishing a structured interaction between an LLM and the Lean4 verifier guided by a Long CoT. Addressing the challenge of continuous learning, LeanAgent \cite{kumarappanleanagent} optimizes its learning trajectory based on mathematical difficulty and manages evolving knowledge through a dynamic database, enabling stable yet plastic lifelong mathematical learning via progressive training. Besides, Prover Agent \cite{baba2025prover} uses an informal reasoning language model for high-level thinking and a separate formal proof model to execute the theorem-proving steps in Lean. During its reasoning process, the system strategically creates auxiliary intermediate theorems to guide the proof and leverages feedback from Lean to reflect upon and adjust its strategy. Furthermore, MathSensei \cite{das2024mathsensei} emphasizes the auxiliary role of tool invocation in mathematical reasoning. It equips its agent with a comprehensive suite of tools, including a knowledge retriever (powered by an LLM or Bing Web Search), a Python-based program generator and executor, and a symbolic problem solver (Wolfram-Alpha), thereby significantly extending the boundaries of the system's reasoning capabilities.

\subsubsection{Astrophysics}
In astrophysics, agent systems are being developed to assist the research process by managing vast, proprietary datasets through automated and scalable collaboration. AstroAgents \cite{saeediastroagents} generates hypotheses from spectral data. It employs a team of eight specialized agents that work in sequence to interpret the data, perform deep analysis on specific segments, formulate hypotheses, conduct literature searches, evaluate the hypotheses, and propose refinements. Expanding the scope to the entire scientific lifecycle, The AI Cosmologist \cite{moss2025ai} implements an end-to-end pipeline encompassing ideation, experimental evaluation, and research dissemination. It utilizes dedicated agents for planning, coding, execution, analysis, and synthesis, aiming to automate the complex workflows of data analysis and machine learning research in cosmology and astronomy. Focusing on the cosmological parameter analysis, Laverick et al.'s work \cite{laverick2024multi} is built upon AutoGen framework \cite{wu2024autogen} and integrates Retrieval-Augmented Generation (RAG) to facilitate the auxiliary analysis of cosmological data.

\subsubsection{Geo-science}
The integration of Geographic Information Systems (GIS) with agentic reasoning frameworks can significantly enhance a system's ability to autonomously reason, deduce, innovate, and advance geospatial solutions \cite{li2025giscience}. As a pioneering work, Autonomous GIS \cite{li2023autonomous} introduces an agent-based framework for geospatial problem analysis. The system decomposes user requirements into ordered operational steps, constructs a flowchart, and generates Python code to sequentially execute tasks such as data loading, spatial joins, statistical analysis, and plotting to produce the final output. Concurrent works have specialized in particular aspects of the workflow. Ning et al. \cite{ning2025autonomous} enhances the reasoning process by focusing on data retrieval. It performs autonomous data discovery based on task understanding and a data-source manual, while generating Python retrieval code via in-context learning that is iteratively executed, debugged, and refined by the framework. Moreover, Pantiukhin et al. \cite{pantiukhin2025accelerating} leverages a centralized Multi-Agent System and a suite of earth science tools for data processing, analysis, and visualization. Crucially, it incorporates a reflection module to contemplate evaluation outcomes and drive iterative improvements to its plan. Besides, GeoLLM-Squad \cite{lee2025multi} targets on Remote Sensing (RS) workflows. Built upon the AutoGen \cite{wu2024autogen} and GeoLLM-Engine \cite{singh2024geollm} frameworks, it modularizes RS applications by decomposing complex tasks and assigning them to specialized sub-agents, covering areas such as urban monitoring, forestry conservation, climate analysis, and agricultural research.

Further research has focused on improving the quality and scope of agent-based geospatial analysis. To mitigate subjective bias in domain-specific question answering, Wang et al. \cite{wang2024mitigating} utilizes RAG and online search to comprehensively gather relevant information. The system then employs a CoT \cite{wei2022chain} process to integrate and reflect upon this information, ensuring reliable geospatial analysis. Pushing the boundaries of task complexity, GeoAgent \cite{huang2024geoagent} builds upon RAG by incorporating Monte Carlo Tree Search (MCTS) to plan and execute multi-step analyses. Starting from a natural language description, it iteratively generates, runs, and debugs multi-step code. GeoAgent \cite{huang2024geoagent} also introduced the Geocode Benchmark, a comprehensive suite of single and multi-turn tasks involving data acquisition, analysis, and visualization to evaluate agents in diverse geospatial contexts. Venturing into multi-modal understanding, GeoMap-Agent \cite{huang2025peace} pioneers the use of a Multi-modal Large Language Model (MLLM) to interpret geological maps. It performs hierarchical information extraction to segment the map and identify salient elements. This is followed by retrieving domain knowledge from an expert database, which is integrated into an enhanced prompt to enable precise question answering. GeoMap-Agent \cite{huang2025peace} also introduce GeoMap-Bench, the first benchmark designed to assess the geological map understanding capabilities of MLLMs across a full spectrum of skills, including extraction, referring, localization, reasoning, and analysis.
Building on these tool-oriented workflows, Spatial-Agent~\cite{bao2026spatialagent} makes spatial concepts and their dependencies explicit in a GeoFlow Graph, which is mapped to executable operations through templates and ordering constraints. This representation connects the selection of geospatial tools to the meaning and order of the transformations required by the question.

\subsubsection{Biochemistry and Material Science}
\label{sec:biochem}

Agentic systems connect domain knowledge and computational tools to research workflows~\cite{abramson2024accurate}. The five application areas below differ in the evidence available to reasoning: retrieved records, computed estimates, execution feedback and experimental observations. Table~\ref{tab:biochem} compares how representative systems obtain and integrate these inputs; Table~\ref{tab:bio_mat} lists evaluation resources. Related applications in diagnostic assistance and clinical workflows are discussed in greater detail in \S\ref{sec:healthcare}.

\input{fig/table_biochem}

\paragraph{Drug Discovery}
In drug discovery and design, agentic systems must connect user requirements with domain knowledge and computational evidence. Existing frameworks explore different ways to organize this information within individual reasoning loops or specialized teams. ChatDrug~\cite{liu2024conversational} combines retrieved examples, domain feedback and a conversational refinement loop. CLADD~\cite{lee2025rag} instead distributes evidence gathering among a planning team, a knowledge-graph team and a molecular-understanding team, whose outputs support a prediction agent. Both ground reasoning in external knowledge, but they organize access differently: ChatDrug incorporates retrieved examples into an iterative dialogue, whereas CLADD coordinates complementary evidence sources before forming a response. Computational tools can also supply numerical estimates of molecular properties for subsequent reasoning. LIDDIA~\cite{averly2025liddia} separates the reasoner, executor and evaluator, retaining these estimates in memory to inform later decisions. Here, feedback comes from computational assessment of a candidate, rather than from retrieved descriptions alone. DrugAgent~\cite{inoue2025drugagent} uses a different organization: a coordinator manages specialists that provide machine-learning predictions, knowledge-graph evidence and literature evidence, while a reasoning agent integrates their outputs into an explained assessment. Its central mechanism is evidence aggregation across sources. PharmAgents~\cite{gao2025pharmagents} instead assigns specialist responsibilities around successive research stages, linking their outputs through a broader workflow. ChatChemTS~\cite{ishida2025large} emphasizes the interface between user intentions and computational tools: a ReAct agent coordinates dedicated components that translate requirements into executable objectives and configurations. These examples distinguish using computed feedback to guide later actions, coordinating evidence for a conclusion, and configuring an external tool through language.

\begin{figure}
    \centering
    \includegraphics[width=1\linewidth]{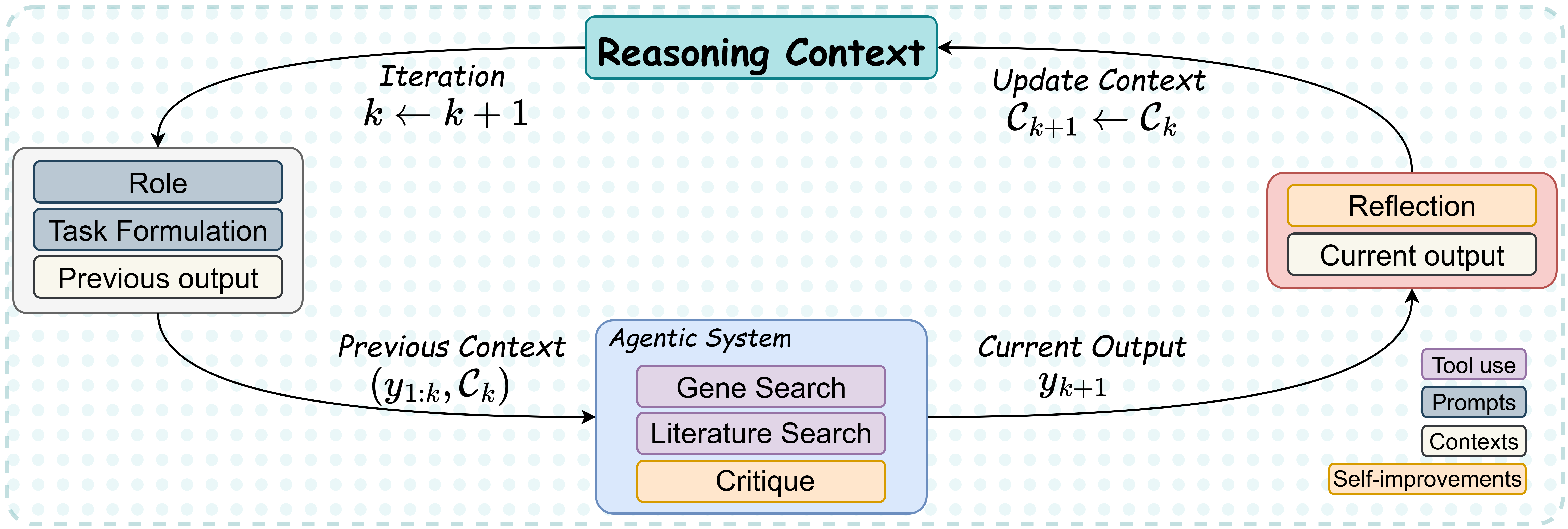}
    \caption{The BioDiscoveryAgent pipeline~\cite{roohanibiodiscoveryagent}, interpreted using our taxonomy. Experimental outcomes inform context updates and subsequent experiment selection. Adapted from~\cite{roohanibiodiscoveryagent}.}
    \Description{A flowchart of BioDiscoveryAgent showing iterative experimental design and context updates from observed outcomes.}
    \label{fig:biopipe}
\end{figure}

\paragraph{Genomics and Biological Experiment Design}
In genomics and biological experiment design, agentic systems assist researchers in interpreting heterogeneous information and organizing complex research workflows. Their reasoning therefore depends on how prior knowledge and new observations inform subsequent decisions. BioDiscoveryAgent~\cite{roohanibiodiscoveryagent} carries previous outcomes and external knowledge into the context used for subsequent research decisions (Fig.~\ref{fig:biopipe}). Knowledge-guided assistants also support research planning~\cite{qu2025crispr}. For bioinformatics assistance, BioAgents~\cite{mehandru2025bioagents} combines a specialist trained on tool documentation with a specialist that retrieves workflow documentation; a separate reasoning agent integrates their responses. This design gives complementary knowledge sources distinct roles within the framework. When the task centers on computational analysis, feedback can enter through execution and checking: AutoBA~\cite{zhou2024ai} plans an analysis, executes generated code and revises it from execution feedback; BioMaster~\cite{su2025biomaster} distributes decomposition, execution and validation across agents. These applications share a context-update loop, but scientific observations, retrieved evidence and code checks provide different kinds of justification for revising a conclusion.

\begin{wraptable}{R}{0.5\surveycolumnwidth}
\centering
\caption{Evaluation strategies in biochemistry and material science. Metrics assess response or property quality; datasets support repeatable task evaluation; case studies examine complete research workflows. The categories provide complementary evidence.}
\label{tab:bio_mat}
\resizebox{0.5\surveycolumnwidth}{!}{%
\begin{minipage}{\surveycolumnwidth}
\centering
\begin{minipage}{\textwidth}
    \vspace{-2mm}
    \centering
    
    \resizebox{\linewidth}{!}{%
        \setlength{\heavyrulewidth}{1.5pt}%
        \renewcommand{\arraystretch}{1.05}\normalsize%
        \begin{tabular}{@{} c c | >{\centering\arraybackslash}p{8cm} @{}}
            \toprule
            \multicolumn{3}{c}{\cellcolor[HTML]{EDEDED}\itshape\bfseries Metrics Level} \\
            \midrule
            \textbf{Focus} & \textbf{Related Work} & \textbf{Metrics} \\
            \midrule

            \makecell[c]{Lab-involved \\ Biomedical Research}
            & BioResearcher~\raisebox{0pt}[\height][0pt]{\cite{luo2025intention}}
            & \makecell[c]{Completeness, Level of Detail, Correctness, \\ Logical Soundness, Structural Soundness} \\
            \cmidrule(lr){1-3}

            \makecell[c]{High-fidelity Materials \\ Knowledge Retrieval}
            & LLaMP~\raisebox{0pt}[\height][0pt]{\cite{chiang2025llamp}}
            & \makecell[c]{Precision, coefficient of precision, confidence, \\ self-consistency of response, MAE} \\
            \cmidrule(lr){1-3}

            Drug Discovery
            & LIDDIA~\raisebox{0pt}[\height][0pt]{\cite{averly2025liddia}}
            & \makecell[c]{drug-likeness~\raisebox{0pt}[\height][0pt]{\cite{bickerton2012quantifying}}, Lipinski’s Rule of Five~\raisebox{0pt}[\height][0pt]{\cite{lipinski1997experimental}}, \\ synthetic accessibility~\raisebox{0pt}[\height][0pt]{\cite{ertl2009estimation}}, binding affinities~\raisebox{0pt}[\height][0pt]{\cite{trott2010autodock}}} \\
            \midrule
            \multicolumn{3}{c}{\cellcolor[HTML]{DDEBF7}\itshape\bfseries Benchmark/Dataset Level} \\
            \midrule
            \textbf{Focus} & \textbf{Related Work} & \textbf{Benchmark/Dataset} \\
            \midrule
            \makecell[c]{Genetic Perturbation \\ Experiment Design}
            & BioDiscoveryAgent~\raisebox{0pt}[\height][0pt]{\cite{roohanibiodiscoveryagent}}
            & \makecell[c]{Schmidt et al.~\raisebox{0pt}[\height][0pt]{\cite{schmidt2022crispr}}, Carnevale et al.~\raisebox{0pt}[\height][0pt]{\cite{carnevale2022rasa2}}, \\ Scharenberg et al.~\raisebox{0pt}[\height][0pt]{\cite{scharenberg2023spns1}}, Sanchez et al.~\raisebox{0pt}[\height][0pt]{\cite{sanchez2021genome}}, \\ Horlbeck et al.~\raisebox{0pt}[\height][0pt]{\cite{horlbeck2018mapping}}, OpenTargets~\raisebox{0pt}[\height][0pt]{\cite{ochoa2023next}}} \\
            \cmidrule(lr){1-3}

            Quantum Chemistry
            & El Agente Q~\raisebox{0pt}[\height][0pt]{\cite{zou2025agente}}
            & \makecell[c]{University-level \\ course exercises~\raisebox{0pt}[\height][0pt]{\cite{zou2025agente}}} \\
            \cmidrule(lr){1-3}

            Drug Discovery
            & LIDDIA~\raisebox{0pt}[\height][0pt]{\cite{averly2025liddia}}
            & \makecell[c]{PDB~\raisebox{0pt}[\height][0pt]{\cite{berman2000protein}}, ChEMBL~\raisebox{0pt}[\height][0pt]{\cite{bento2014chembl}}} \\
            \cmidrule(lr){1-3}

            \makecell[c]{Chemical Tool Invocation \\ Optimization}
            & ChemHTS~\raisebox{0pt}[\height][0pt]{\cite{li2025chemhts}}
            & \makecell[c]{ChemLLMBench~\raisebox{0pt}[\height][0pt]{\cite{guo2023can}}, ChEBI-20-MM~\raisebox{0pt}[\height][0pt]{\cite{liu2025quantitative}}, \\ USPTO-MIT~\raisebox{0pt}[\height][0pt]{\cite{jin2017predicting}}, MoleculeNet~\raisebox{0pt}[\height][0pt]{\cite{wu2018moleculenet}}} \\
            \cmidrule(lr){1-3}

            \makecell[c]{Tool-assisted \\ Molecular discovery}
            & CACTUS~\raisebox{0pt}[\height][0pt]{\cite{mcnaughton2024cactus}}
            & CACTUS~\raisebox{0pt}[\height][0pt]{\cite{mcnaughton2024cactus}} \\
            \cmidrule(lr){1-3}

            Mendelian Disease Diagnosis
            & MD2GPS~\raisebox{0pt}[\height][0pt]{\cite{zhou2025llm}}
            & \makecell[c]{SCH~\raisebox{0pt}[\height][0pt]{\cite{huang2023diseasegps}}, JN~\raisebox{0pt}[\height][0pt]{\cite{zhou2025llm}}, DDD~\raisebox{0pt}[\height][0pt]{\cite{firth2009decipher}}, RD~\raisebox{0pt}[\height][0pt]{\cite{zhang2018compendium}}} \\
            \cmidrule(lr){1-3}

            \makecell[c]{Automated Bioinformatics \\ Analysis}
            & BioMaster~\raisebox{0pt}[\height][0pt]{\cite{su2025biomaster}}
            & \makecell[c]{Morey et al.~\raisebox{0pt}[\height][0pt]{\cite{morey2013rybp}}, \raisebox{0pt}[\height][0pt]{\cite{10002012integrated}}, Rao et al.~\raisebox{0pt}[\height][0pt]{\cite{rao20143d}}} \\
            \cmidrule(lr){1-3}

            RAG-based Drug Discovery
            & CLADD~\raisebox{0pt}[\height][0pt]{\cite{lee2025rag}}
            & \makecell[c]{MoleculeNet~\raisebox{0pt}[\height][0pt]{\cite{wu2018moleculenet}}, Drug Repurposing Hub~\raisebox{0pt}[\height][0pt]{\cite{corsello2017drug}},\\ DrugBank~\raisebox{0pt}[\height][0pt]{\cite{wishart2018drugbank}},STITCH~\raisebox{0pt}[\height][0pt]{\cite{szklarczyk2016stitch}}, hERG~\raisebox{0pt}[\height][0pt]{\cite{wang2016admet}},\\DILI~\raisebox{0pt}[\height][0pt]{\cite{xu2015deep}},Skin~\raisebox{0pt}[\height][0pt]{\cite{alves2015predicting}},Lagunin et al.~\raisebox{0pt}[\height][0pt]{\cite{lagunin2009computer}}} \\
            \cmidrule(lr){1-3}

            full drug discovery workflow
            & PharmAgents~\raisebox{0pt}[\height][0pt]{\cite{gao2025pharmagents}}
            & \makecell[c]{CrossDocked~\raisebox{0pt}[\height][0pt]{\cite{francoeur2020three}}, PharmAgents~\raisebox{0pt}[\height][0pt]{\cite{gao2025pharmagents}}} \\
            \cmidrule(lr){1-3}

            \makecell[c]{Drug-target Interaction \\ Prediction}
            & DrugAgent~\raisebox{0pt}[\height][0pt]{\cite{inoue2025drugagent}}
            & Anastassiadis et al.~\raisebox{0pt}[\height][0pt]{\cite{anastassiadis2011comprehensive}} \\
            \cmidrule(lr){1-3}
            
            Drug Editing
            & \makecell[c]{Liu et al.~\raisebox{0pt}[\height][0pt]{\cite{liu2024conversational}}}
            & \makecell[c]{Rao et al.~\raisebox{0pt}[\height][0pt]{\cite{rao2019evaluating}}, MHCflurry 2.0~\raisebox{0pt}[\height][0pt]{\cite{o2020mhcflurry}},ZINC~\raisebox{0pt}[\height][0pt]{\cite{irwin2012zinc}}} \\
            \cmidrule(lr){1-3}

            Gene-Editing
            & CRISPR-GPT~\raisebox{0pt}[\height][0pt]{\cite{qu2025crispr}}
            & Gene-editing bench~\raisebox{0pt}[\height][0pt]{\cite{qu2025crispr}} \\
            \cmidrule(lr){1-3}

            Chemical Hypothesis Discovery
            & MOOSE-Chem~\raisebox{0pt}[\height][0pt]{\cite{yangmoose}}
            & MOOSE-Chem~\raisebox{0pt}[\height][0pt]{\cite{yangmoose}} \\
            \midrule
            \multicolumn{3}{c}{\cellcolor[HTML]{FCE4D6}\itshape\bfseries Case Study} \\
            \midrule
            \multicolumn{3}{c}{%
                \parbox{16cm}{\centering ProtAgents~\raisebox{0pt}[\height][0pt]{\cite{ghafarollahi2024protagents}}, Stewart et al.~\raisebox{0pt}[\height][0pt]{\cite{stewart2025molecular}}, PharmAgents~\raisebox{0pt}[\height][0pt]{\cite{gao2025pharmagents}}, CLADD~\raisebox{0pt}[\height][0pt]{\cite{lee2025rag}}, TourSynbio-Search~\raisebox{0pt}[\height][0pt]{\cite{liu2024toursynbio}}, AutoBA~\raisebox{0pt}[\height][0pt]{\cite{zhou2024ai}}, AI-HOPE~\raisebox{0pt}[\height][0pt]{\cite{yang2025ai}}, ChatMOF~\raisebox{0pt}[\height][0pt]{\cite{kang2024chatmof}}, Chemist-X~\raisebox{0pt}[\height][0pt]{\cite{chen2023chemist}}, Ruan et al.~\raisebox{0pt}[\height][0pt]{\cite{ruan2024automatic}}, ChatChemTS~\raisebox{0pt}[\height][0pt]{\cite{ishida2025large}}, El Agente Q~\raisebox{0pt}[\height][0pt]{\cite{zou2025agente}}, ChemCrow~\raisebox{0pt}[\height][0pt]{\cite{bran2023chemcrow}}, AtomAgents~\raisebox{0pt}[\height][0pt]{\cite{ghafarollahi2025automating}}.
                }%
            } \\
            \bottomrule
        \end{tabular}%
    }
    \vspace{-5mm}
\end{minipage}
\end{minipage}%
}
\end{wraptable}

\paragraph{Chemical Synthesis and Design}
In chemical synthesis and design, agentic systems connect domain-specific tools and literature to broader research workflows. A central design choice is how to coordinate these sources of evidence across successive reasoning steps. ChemCrow~\cite{bran2023chemcrow} equips an LLM with expert-designed chemistry tools, allowing intermediate reasoning to determine which tool to invoke and how its output enters the next step. CACTUS~\cite{mcnaughton2024cactus} similarly interleaves question analysis with tool selection and interpretation of returned results. El Agente Q~\cite{zou2025agente} develops a hierarchical organization: a top-level agent plans the task, while specialized agents execute assigned computations and respond to execution errors. Separating these responsibilities also limits the context each agent needs to handle. Staged workflows make a different division. Chemist-X~\cite{chen2023chemist} connects retrieval to recommendation through successive components, while Ruan et al.~\cite{ruan2024automatic} assign dedicated agents to different research activities. The handoff between stages, rather than only the next tool call, becomes part of the reasoning structure. Literature-driven frameworks additionally use retrieved work to support new hypotheses or collaborative analysis~\cite{yangmoose,stewart2025molecular}. Thus, similar information sources can support an iterative tool loop, hierarchical delegation or staged collaboration.

\paragraph{Material Science}
In materials science, agentic systems bring together structured databases, domain knowledge and computational models to support analysis and design. These resources motivate both tool-centered reasoning and collaboration among specialized agents. ChatMOF~\cite{kang2024chatmof} combines database and internet retrieval with specialized prediction and generation tools, using returned observations to revise its actions. LLaMP~\cite{chiang2025llamp} separates high-level coordination from tool-specific reasoning: a supervisor delegates requests to ReAct assistants, each responsible for a particular tool or API. The assistants can revise calls using returned feedback, while the supervisor integrates their results. This hierarchy reduces the number of tool schemas presented to any one agent. AtomAgents~\cite{ghafarollahi2025automating} instead coordinates specialized roles across knowledge retrieval, physics-based simulation and multimodal analysis. The distinction concerns how responsibilities are divided: around individual tool interfaces in LLaMP and around complementary activities in a broader research workflow in AtomAgents. Both illustrate how tool integration and multi-agent organization can be composed (\S\ref{sec:tool_based}).

\paragraph{Biochemical Automated Research}
Beyond specialized tasks, several works propose systematic designs from the broader perspective of automating biochemical research. These systems place greater emphasis on agent orchestration and the integration of evidence across research stages. BioResearcher~\cite{luo2025intention} organizes retrieval, planning and analysis hierarchically, with reviewer feedback at intermediate stages. ProtAgents~\cite{ghafarollahi2024protagents} uses a chat manager to select participating agents and coordinate communication among specialized roles. TourSynbio-Search~\cite{liu2024toursynbio} places more emphasis on routing user requests to suitable agents and asking the user to clarify ambiguous inputs. These are distinct control choices: coordinating an ongoing collaboration and resolving which capability a request requires. ChemHTS~\cite{li2025chemhts} places the emphasis on the tool workflow itself, combining exploration, hierarchical composition and selective retention of intermediate information. Across these applications, the shared question is which component holds the task state, selects the next capability and checks a result before reuse. Table~\ref{tab:biochem} separates the source of evidence from the mechanism that incorporates it into reasoning.

\subsubsection{General Research}
The problem-solving capabilities of agentic frameworks can be generalized from specialized, domain-specific tasks to broader research inquiries. These systems are designed to operate from an initial prompt to the final deliverable of a detailed research report. Here we classify these scenarios into \textit{literature survey, end-to-end research automation, research collaboration and refinement}. Table \ref{tab:general} summarize different metrics and datasets used in each selected work, with their specific focus about general research. We also conclude the work that use case study or subjective methods to evaluate their work.

\begin{wraptable}{R}{0.5\surveycolumnwidth}
\centering
\caption{Evaluation strategies for general research agents: output metrics, benchmark and dataset studies, research case studies, and model or human judgments. Corpus sizes describe the listed resources, rather than a common number of evaluated tasks.}
\label{tab:general}
\resizebox{0.5\surveycolumnwidth}{!}{%
\begin{minipage}{\surveycolumnwidth}
\centering
\begin{minipage}{\textwidth}
\vspace{-2mm}
\centering

\resizebox{\linewidth}{!}{%
\begingroup %
\setlength{\heavyrulewidth}{1.5pt}%
\renewcommand{\arraystretch}{1.05}\normalsize
\setlength{\tabcolsep}{5pt}
\begin{tabular}{cc|c|cc} 
\toprule
\multicolumn{5}{c}{\cellcolor[HTML]{E2EFDA}\textit{\textbf{Metrics Level}}} \\
\textbf{Focus} &
  \textbf{Related Work} &
  \multicolumn{3}{c}{\textbf{Metrics}} \\ 
\midrule
 &
  AutoSurvey~\raisebox{0pt}[\height][0pt]{\cite{wang2024autosurvey}} &
  \multicolumn{3}{c}{Survey Creation Speed, Content Quality} \\
 &
  Gao et al.~\raisebox{0pt}[\height][0pt]{\cite{gao2023enabling}} &
  \multicolumn{3}{c}{Citation Quality} \\
 &
  SurveyX~\raisebox{0pt}[\height][0pt]{\cite{liang2025surveyx}} &
  \multicolumn{3}{c}{Insertion over Union, semantic-based reference relevance} \\
 &
  SurveyForge~\raisebox{0pt}[\height][0pt]{\cite{yan2025surveyforge}} &
  \multicolumn{3}{c}{Reference, Outline, and Content Quality (SAM Metrics)} \\
 &
  Agent Laboratory~\raisebox{0pt}[\height][0pt]{\cite{schmidgall2025agent}} &
  \multicolumn{3}{c}{Inference cost, Inference time, Success Rate} \\
 &
  Su et al.~\raisebox{0pt}[\height][0pt]{\cite{su2025icml}} &
  \multicolumn{3}{c}{Proxy MSE, Proxy MAE} \\
\multirow{-7}{*}{\makecell{Paper Generation\\Quality}} &
  VirSci~\raisebox{0pt}[\height][0pt]{\cite{su-etal-2025-many}} &
  \multicolumn{3}{c}{\makecell{Historical Dissimilarity, Contemporary Dissimilarity,\\ Contemporary Impact}} \\ 
\midrule
\multicolumn{5}{c}{\cellcolor[HTML]{D9E1F2}\textit{\textbf{Benchmark/Dataset Level}}} \\
\textbf{Focus} &
  \textbf{Related Work} &
  \multicolumn{1}{c}{\textbf{Type}} &
  \textbf{Benchmark/Dataset} &
  \textbf{\#Papers} \\ 
\midrule
Idea Generation &
  Research Agent~\raisebox{0pt}[\height][0pt]{\cite{baek2025researchagent}} &
   &
  ResearchAgent~\raisebox{0pt}[\height][0pt]{\cite{baek2025researchagent}} &
  300 \\
Survey Automation &
  SurveyForge~\raisebox{0pt}[\height][0pt]{\cite{yan2025surveyforge}} &
   &
  SurveyBench~\raisebox{0pt}[\height][0pt]{\cite{yan2025surveyforge}} &
  100 \\
Research Assistants &
  Agent Laboratory~\raisebox{0pt}[\height][0pt]{\cite{schmidgall2025agent}} &
   &
  Mle-bench~\raisebox{0pt}[\height][0pt]{\cite{chanmle}} &
  - \\
Research Improvement &
  CycleResearcher~\raisebox{0pt}[\height][0pt]{\cite{wengcycleresearcher}} &
   &
  \makecell{Review-5k/ research-14k~\raisebox{0pt}[\height][0pt]{\cite{wengcycleresearcher}}} &
  $\sim$5k/$\sim$14k \\
Scientific Innovation &
  AI-Researcher~\raisebox{0pt}[\height][0pt]{\cite{tang2025ai}} &
  \multirow{-5}{*}{\makecell{Off-\\line}} &
  Scientist-Bench~\raisebox{0pt}[\height][0pt]{\cite{tang2025ai}} &
  22 \\ 
\cmidrule(lr){1-5}
 &
  ResearchAgent~\raisebox{0pt}[\height][0pt]{\cite{baek2025researchagent}} &
   &
  \makecell{Semantic Scholar Academic Graph API$^*$}&
  - \\
 &
  &
   &
  \makecell{AMiner Computer Science dataset$^\dagger$}&
  $\sim$2M \\
\multirow{-3}{*}{Idea Generation} &
  \multirow{-2}{*}{VirSci~\raisebox{0pt}[\height][0pt]{\cite{su-etal-2025-many}}} &
  \multirow{-3}{*}{\makecell{On-\\line}} &
  \makecell{Open Academic Graph 3.1$^\ddagger$}&
  $\sim$131M \\ 
\midrule
\multicolumn{5}{c}{\cellcolor[HTML]{FFF2CC}\textit{\textbf{Case Study}}} \\
\multicolumn{5}{c}{\makecell{IdeaSynth~\raisebox{0pt}[\height][0pt]{\cite{pu2025ideasynth}}, The AI Scientist~\raisebox{0pt}[\height][0pt]{\cite{lu2024ai}}, Co-Scientist~\raisebox{0pt}[\height][0pt]{\cite{gottweis2025towards}}, Robin~\raisebox{0pt}[\height][0pt]{\cite{ghareeb2025robin}}, SciAgents~\raisebox{0pt}[\height][0pt]{\cite{ghafarollahi2025sciagents}}}} \\ 
\midrule
\multicolumn{5}{c}{\cellcolor[HTML]{FCE4D6}\textit{\textbf{LLM-as-a-Judge/Human Evaluation}}} \\
\multicolumn{5}{c}{\makecell{ResearchAgent~\raisebox{0pt}[\height][0pt]{\cite{baek2025researchagent}}, AutoSurvey~\raisebox{0pt}[\height][0pt]{\cite{wang2024autosurvey}}, SurveyX~\raisebox{0pt}[\height][0pt]{\cite{liang2025surveyx}},Agent Laboratory~\raisebox{0pt}[\height][0pt]{\cite{schmidgall2025agent}},The AI Scientist v2~\raisebox{0pt}[\height][0pt]{\cite{lu2024ai}},\\CycleResearcher~\raisebox{0pt}[\height][0pt]{\cite{wengcycleresearcher}},Co-Scientist~\raisebox{0pt}[\height][0pt]{\cite{gottweis2025towards}}}} \\ \bottomrule
\multicolumn{4}{l}{$^*$\url{https://www.semanticscholar.org/product/api}} \\
\multicolumn{4}{l}{$^\dagger$\url{https://www.aminer.cn/aminernetwork}} \\
\multicolumn{4}{l}{$^\ddagger$\url{https://open.aminer.cn/open/article?id=65bf053091c938e5025a31e2}} \\
\end{tabular}%
\endgroup
}
\vspace{-5mm}
\end{minipage}
\end{minipage}%
}
\end{wraptable}

\paragraph{Literature Survey}
Automating literature surveys requires systems to process and orchestrate vast amounts of domain literature to generate comprehensive and coherent reviews. AutoSurvey \cite{wang2024autosurvey} first retrieves relevant literature via semantic search and generates a preliminary outline. It then employs multiple agents to concurrently draft each section, retrieving additional literature to produce text with accurate citations. After drafting, the system integrates and refines the content, using multi-LLM-as-judge setup to score the survey's quality and coverage for iterative improvement. Similarly, SurveyX \cite{liang2025surveyx} automates this process in two phases: a preparation phase that uses an attribute tree structure to acquire and preprocess literature, and a generation phase that performs both coarse and fine-grained content creation from an initial plan. RAG is leveraged to ensure citation accuracy during rewriting and to support multi-media content generation. Besides, SurveyForge \cite{yan2025surveyforge} combines a library of human-written outlines with domain-specific papers to generate a new outline via in-context learning. It introduces a memory-driven framework with multi-layer sub-query and retrieval memories to refine the search process, followed by filtering and reconstruction stages for content integration and parallel text generation. This work also contributes SurveyBench, a benchmark for quantitatively evaluating survey quality across multiple dimensions.

\paragraph{End-to-End Research Automation}
Moving beyond knowledge synthesis, end-to-end research automation aims to emulate the entire scientific lifecycle for a given topic, often by mimicking real-world research workflows. As a representative work, Agent Laboratory \cite{schmidgall2025agent} actualizes a user's research idea through a multi-stage, hierarchical architecture covering literature retrieval, experimental planning, code generation, result analysis, and report writing, integrating human feedback at each stage for quality assurance. Many similar works decompose the research process to achieve automation. The AI Scientist \cite{lu2024ai} decompose scientific discovery into five stages (ideation, experiment design, execution, paper writing, and peer review), guiding agents with structured instructions. AI Scientist-v2 \cite{lu2024ai} further introduces an experimental parameter space and a tree-search algorithm to optimize the research process, incorporating a vision-language model (VLM) to provide feedback on generated figures and text. Notably, one of three manuscripts generated by this system met the acceptance threshold in an ICLR 2025 workshop experiment; the submissions were withdrawn after review under the agreed experimental protocol\footnote{\scriptsize{\url{https://github.com/SakanaAI/AI-Scientist-ICLR2025-Workshop-Experiment/}}}. Moreover, AI-Researcher \cite{tang2025ai} adopt a four-stage decomposition of the research pipeline from literature review, hypothesis generation, algorithm implementation to manuscript writing. AI-Researcher \cite{tang2025ai} also propose Scientist-Bench, which comprises recent papers spanning diverse AI research areas and includes both guided-innovation and open-ended exploration tasks. 
Moreover, Robin \cite{ghareeb2025robin} facilitates semi-autonomous discovery by progressing through four stages: hypothesis generation, experimental design, result interpretation, and hypothesis updating.
Besides, PaperCoder \cite{seo2025paper2code} automates the reproduction of code from machine learning papers. It uses three dedicated agents for a three-stage process: a planning stage to create a high-level roadmap, an analysis stage to interpret specific implementation details, and a generation stage to produce modular, dependency-aware code.
Extending this iterative organization, DOLPHIN~\cite{yuan2025dolphin} connects literature-informed proposals with template-based implementation, execution-error feedback and analysis of experimental outcomes. The results inform subsequent proposals, allowing the research process to revise its direction using evidence from earlier experiments.

\paragraph{Research Collaboration and Refinement}
There are some emergent research focuses on introducing specific mechanisms for collaboration and refinement to enhance the quality, creativity, and efficiency of automated science. A significant direction is improving idea and hypothesis generation through multi-agent interaction. ResearchAgent \cite{baek2025researchagent} iteratively refines research ideas by integrating core papers, their citation network, and feedback from a panel of ``reviewer agents'' to mimic peer review and foster innovation. Similarly, VirSci \cite{su-etal-2025-many} uses multi-agent dialogue -- spanning collaborator selection, topic discussion, and novelty assessment -- to generate ideas, while AI co-scientist \cite{gottweis2025towards} employs a hierarchical agent team for critical debate to propose, critique, and refine hypotheses. Scientists set the research goals, review the proposals and guide subsequent exploration; experimental validation remains distinct from the agents' automated discussion and refinement.

Another key direction is the refinement of the research process and agent capabilities. CycleResearcher \cite{wengcycleresearcher} proposes an iterative training framework that simulates the ``research-review-improve'' academic closed loop to enhance generated paper quality, contributing two large datasets for this purpose. Furthermore, SciAgents \cite{ghafarollahi2025sciagents} leverages an ontology knowledge graph for structured reasoning, while SciMaster \cite{chai2025scimaster} uses a decentralized, stacked workflow to scale reasoning depth and breadth. At the ecosystem level, AgentRxiv \cite{schmidgall2025agentrxiv} constructs a scientific ecosystem where multiple \textit{Agent Laboratories} \cite{schmidgall2025agent} can collaborate asynchronously. By using a centralized preprint server, independent agent teams can upload, share, and retrieve research, forming a dynamic knowledge commons that enables cumulative, collaborative scientific discovery.

\leavevmode\phantomsection\label{par:design_science}Across these research settings, the timing and interpretation of feedback help explain framework choices. Verifiers and execution environments support iterative refinement when checks are available within a task~\cite{wang2025ma,baba2025prover,zhou2024ai,huang2024geoagent}. When evidence must be gathered and interpreted across research stages, specialized roles can separate retrieval, analysis and review, while introducing coordination and handoff costs~\cite{lee2025rag,schmidgall2025agent,lu2024ai}. Retrieval may thus operate inside one agent's reasoning loop or supply evidence exchanged between roles~\cite{liu2024conversational,kang2024chatmof,chen2023chemist,ruan2024automatic}. Human involvement remains important at goal setting and proposal review when validation or decision authority extends beyond the automated workflow~\cite{gottweis2025towards,schmidgall2025agent}.

\subsection{Healthcare}
\label{sec:healthcare}

\begin{wraptable}{R}{0.5\surveycolumnwidth}
\centering
\caption{Evaluation strategies for healthcare agents, grouped into dataset tests, system-level evaluation, simulated environments and case studies. Clinical validity requires evidence beyond benchmark or simulation performance.}
\label{tab:healthcare}
\resizebox{0.5\surveycolumnwidth}{!}{%
\begin{minipage}{\surveycolumnwidth}
\centering
\begin{minipage}{\textwidth}
    \vspace{-2mm}
    \centering
    
    \setlength{\heavyrulewidth}{1.5pt}%
    \renewcommand{\arraystretch}{1.05}\normalsize
    \setlength{\tabcolsep}{3.5pt}
    
    \resizebox{\linewidth}{!}{%
        \begin{tabular}{cl|cc} 
            \toprule
            \multicolumn{4}{c}{\cellcolor[HTML]{D6DCE4}\textit{\textbf{Benchmark/Dataset Level}}} \\
            \textbf{Focus} & \textbf{Related Work} & \multicolumn{2}{c}{\textbf{Benchmark/Dataset}} \\
            \midrule 
            Clinical Consultation Flow & Wang et al.~\raisebox{0pt}[\height][0pt]{\cite{wang2025empowering}} & \multicolumn{2}{c}{MVME~\raisebox{0pt}[\height][0pt]{\cite{fan2025ai}}} \\
            \midrule
            \makecell[c]{Zero-shot Medical Reasoning} & MedAgents~\raisebox{0pt}[\height][0pt]{\cite{tang2024medagents}} & \multicolumn{2}{c}{\begin{tabular}[c]{@{}c@{}}Jin et al.~\raisebox{0pt}[\height][0pt]{\cite{jin2021disease}},Pal et al.~\raisebox{0pt}[\height][0pt]{\cite{pal2022medmcqa}},PubMedQA~\raisebox{0pt}[\height][0pt]{\cite{jin2019pubmedqa}},\\Hendrycks et al.~\raisebox{0pt}[\height][0pt]{\cite{hendrycksmeasuring}}\end{tabular}} \\
            \midrule
            \makecell[c]{Automated supervision of\\Healthcare Safety} & TAO~\raisebox{0pt}[\height][0pt]{\cite{kim2025tiered}} & \multicolumn{2}{c}{\begin{tabular}[c]{@{}c@{}}Safetybench~\raisebox{0pt}[\height][0pt]{\cite{zhang2024safetybench}},Medsafetybench~\raisebox{0pt}[\height][0pt]{\cite{han2024medsafetybench}},\\Chang et al.~\raisebox{0pt}[\height][0pt]{\cite{chang2024red}}, Hu et al.~\raisebox{0pt}[\height][0pt]{\cite{hu2024language}},Wang et al.~\raisebox{0pt}[\height][0pt]{\cite{wang2025can}}\end{tabular}} \\
            \midrule
            Evolvable Medical Agents & MDAgents~\raisebox{0pt}[\height][0pt]{\cite{kim2024mdagents}} & \multicolumn{2}{c}{\begin{tabular}[c]{@{}c@{}}MedQA~\raisebox{0pt}[\height][0pt]{\cite{jin2021disease}}, PubMedQA~\raisebox{0pt}[\height][0pt]{\cite{jin2019pubmedqa}},MedBullets~\raisebox{0pt}[\height][0pt]{\cite{chen2025benchmarking}},\\JAMA~\raisebox{0pt}[\height][0pt]{\cite{chen2025benchmarking}},DDXPlus~\raisebox{0pt}[\height][0pt]{\cite{fansi2022ddxplus}},SymCat~\raisebox{0pt}[\height][0pt]{\cite{al2023nlice}}, \\PathVQA~\raisebox{0pt}[\height][0pt]{\cite{he2020pathvqa}},PMC-VQA~\raisebox{0pt}[\height][0pt]{\cite{zhang2023pmc}},MedVidQA~\raisebox{0pt}[\height][0pt]{\cite{gupta2023dataset}},MIMIC-CXR~\raisebox{0pt}[\height][0pt]{\cite{bae2023ehrxqa}}\end{tabular}} \\
            \midrule
            Healthcare Intent Awareness & MedAide~\raisebox{0pt}[\height][0pt]{\cite{wei2024medaide}} & \multicolumn{2}{c}{\begin{tabular}[c]{@{}c@{}}Pre-Diagnosis, Diagnosis, Medicament,\\Post-Diagnosis Bench~\raisebox{0pt}[\height][0pt]{\cite{wei2024medaide}}\end{tabular}} \\
            \midrule
            Multimodal Tool-integration Diagnosis & MMedAgent~\raisebox{0pt}[\height][0pt]{\cite{li2024mmedagent}} & \multicolumn{2}{c}{\begin{tabular}[c]{@{}c@{}}VQA-RAD~\raisebox{0pt}[\height][0pt]{\cite{lau2018dataset}}, Slake~\raisebox{0pt}[\height][0pt]{\cite{liu2021slake}},PMC-VQA~\raisebox{0pt}[\height][0pt]{\cite{zhang2023pmc}},PathVQA~\raisebox{0pt}[\height][0pt]{\cite{he2020pathvqa}}\end{tabular}} \\
            \midrule
            Tool-assist Therapeutic Reasoning & TxAgent~\raisebox{0pt}[\height][0pt]{\cite{gao2025tx}} & \multicolumn{2}{c}{\begin{tabular}[c]{@{}c@{}}DrugPC, BrandPC, GenericPC, \\ DescriptionPC, TreatmentPC~\raisebox{0pt}[\height][0pt]{\cite{gao2025tx}}\end{tabular}} \\
            \midrule
            Rare Disease Curation & RareAgent~\raisebox{0pt}[\height][0pt]{\cite{chen2024rareagents}} & \multicolumn{2}{c}{\begin{tabular}[c]{@{}c@{}}RareBench~\raisebox{0pt}[\height][0pt]{\cite{chen2024rarebench}},MIMIC-IV~\raisebox{0pt}[\height][0pt]{\cite{johnson2023mimic}}\end{tabular}} \\
            \midrule
            Clinical Trial & ClinicalAgent~\raisebox{0pt}[\height][0pt]{\cite{yue2024clinicalagent}} & \multicolumn{2}{c}{\begin{tabular}[c]{@{}c@{}}DrugBank 5.0~\raisebox{0pt}[\height][0pt]{\cite{wishart2018drugbank}},Himmelstein~\raisebox{0pt}[\height][0pt]{\cite{himmelstein2017systematic}}\end{tabular}} \\
            \midrule
            Multi-modal Diagnosis & MedAgent-Pro~\raisebox{0pt}[\height][0pt]{\cite{wang2025medagent}} & \multicolumn{2}{c}{\begin{tabular}[c]{@{}c@{}}Refuge2 challenge~\raisebox{0pt}[\height][0pt]{\cite{fang2022refuge2}},MITEA~\raisebox{0pt}[\height][0pt]{\cite{zhao2023mitea}}, MIMIC-IV~\raisebox{0pt}[\height][0pt]{\cite{johnson2023mimic}},\\ Nejm image challenge$^*$\end{tabular}} \\
            \midrule
            Mendelian Disease Diagnosis & MD2GPS~\raisebox{0pt}[\height][0pt]{\cite{zhou2025llm}} & \multicolumn{2}{c}{\begin{tabular}[c]{@{}c@{}}SCH~\raisebox{0pt}[\height][0pt]{\cite{huang2023diseasegps}}, JN~\raisebox{0pt}[\height][0pt]{\cite{zhou2025llm}},DDD~\raisebox{0pt}[\height][0pt]{\cite{firth2009decipher}}, RD~\raisebox{0pt}[\height][0pt]{\cite{zhang2018compendium}}\end{tabular}} \\
            \midrule
            Clinical Environment Simulation & MedAgentSim~\raisebox{0pt}[\height][0pt]{\cite{almansoori2025self}} & \multicolumn{2}{c}{\begin{tabular}[c]{@{}c@{}}NEJM~\raisebox{0pt}[\height][0pt]{\cite{schmidgall2024agentclinic}}, MedQA~\raisebox{0pt}[\height][0pt]{\cite{jin2021disease}},MIMIC-IV~\raisebox{0pt}[\height][0pt]{\cite{johnson2023mimic}}\end{tabular}} \\ 
            \midrule
            
            \multicolumn{4}{c}{\cellcolor[HTML]{DDEBF7}\textit{\textbf{System Level}}} \\
            \textbf{Focus} & \textbf{Related Work} & \multicolumn{2}{c}{\textbf{Methods}} \\
            \midrule
            \makecell{Knowledge Graph Enhancement\\for Medical Diagnosis} & KG4Diagnosis~\raisebox{0pt}[\height][0pt]{\cite{zuo2025kg4diagnosis}} & \multicolumn{2}{c}{} \\
            Multi-modal Medical Diagnosis & MedAgent-Pro~\raisebox{0pt}[\height][0pt]{\cite{wang2025medagent}} & \multicolumn{2}{c}{} \\
            Benchmarking multimodal medical agent & AgentClinic~\raisebox{0pt}[\height][0pt]{\cite{schmidgall2024agentclinic}} & \multicolumn{2}{c}{\multirow{-3}{*}{Human Evaluation}} \\
            \midrule
            Multimodal tool-integration Diagnosis & MMedAgent~\raisebox{0pt}[\height][0pt]{\cite{li2024mmedagent}} & \multicolumn{2}{c}{Open-ended Medical Dialogue~\raisebox{0pt}[\height][0pt]{\cite{li2023llava}}} \\
            \midrule
            Medical Necessity Justification & Pandey et al.~\raisebox{0pt}[\height][0pt]{\cite{pandey2024advancing}} & \multicolumn{2}{c}{\makecell{Parent and Leaf node\\ Judgement with accuracy}} \\ 
            \midrule

            \multicolumn{4}{c}{\cellcolor[HTML]{E2EFDA}\textit{\textbf{Environmental Level}}} \\
            \multicolumn{2}{c}{\textbf{Environment}} & \textbf{Related Work} & \textbf{Simulation Focus} \\
            \midrule
            \multicolumn{2}{c}{Entire Illness Treating} & Agent Hospital~\raisebox{0pt}[\height][0pt]{\cite{li2024agent}} & Evolvable Medical Agents \\
            \midrule
            \multicolumn{2}{c}{Doctor-Patient Interaction} & AI Hospital~\raisebox{0pt}[\height][0pt]{\cite{fan2025ai}} & Medical Interaction Simulator \\
            \midrule
            \multicolumn{2}{c}{Multimodal clinical interaction} & AgentClinic~\raisebox{0pt}[\height][0pt]{\cite{schmidgall2024agentclinic}} & Multimodal agent benchmark \\ 
            \midrule
            
            \multicolumn{4}{c}{\cellcolor[HTML]{FCE4D6}\textit{\textbf{Case Study}}} \\
            \multicolumn{4}{c}{\makecell[c]{AMIE~\raisebox{0pt}[\height][0pt]{\cite{tu2025towards}}, AI-HOPE~\raisebox{0pt}[\height][0pt]{\cite{yang2025ai}}, ClinicalAgent~\raisebox{0pt}[\height][0pt]{\cite{yue2024clinicalagent}}}} \\ 
            \bottomrule
            \multicolumn{4}{l}{$^*$\url{https://www.nejm.org/image-challenge}}
        \end{tabular}%
    }
    \vspace{-5mm}
\end{minipage}
\end{minipage}%
}
\end{wraptable}

The advent of powerful foundational Large Language Models is reshaping the landscape of healthcare by empowering agentic systems with new capabilities. This transformation facilitates a critical shift from reactive, predictive functions to proactive, interactive engagement in clinical workflows. 
These advanced agents are increasingly leveraged to resolve chronic issues surrounding clinical efficiency, diagnostic precision, and the quality of patient care. 
As summarized in Table \ref{tab:healthcare}, a wide range of evaluation datasets as well as methods are used to evaluate these framework.
Accordingly, we survey these works  through \textit{diagnostic assistance, clinical management, and environmental simulation}.

\subsubsection{Diagnosis Assistance}

To augment diagnostic capabilities, one primary research focus involves creating multi-agent dialogue frameworks that deconstruct the complex diagnostic process into manageable, collaborative phases. As an early work, MedAgents \cite{tang2024medagents} established a role-playing environment where agents representing different medical experts achieve a consensus through independent analysis and iterative discussion. Concurrently, Wang et al. \cite{wang2025empowering} designed a virtual medical team that includes a physician, a patient, and an examiner in order to model the consultation flow across inquiry, examination, and diagnosis, guided by a hierarchical action set for dynamic responses. Similarly, Chen et al. \cite{chen2025enhancing} employed an admin agent for user information, doctor agents for diagnosis via dialogue, and a supervisor agent to ensure diagnostic consistency. Besides, RareAgents \cite{chen2024rareagents} focuses on rare diseases, featuring a primary physician agent that collaborates with multiple specialist agents over several rounds of discussion, integrating dynamic long-term memory and medical tools for personalized diagnostics.

Beyond simulating dialogue, other frameworks enhance diagnostic reasoning by integrating external knowledge bases and data-driven debate mechanisms. KG4Diagnosis \cite{zuo2025kg4diagnosis} utilizes a hierarchical multi-agent structure where a general practitioner agent first conducts triage before coordinating with specialist agents who perform in-depth diagnosis leveraging a knowledge graph. This approach features an end-to-end pipeline for semantic knowledge extraction and human-guided reasoning, therefore improving system extensibility. MD2GPS \cite{zhou2025llm} introduces a multi-agent debate system driven by both existing literature and patient data to diagnose Mendelian diseases effectively.

A parallel research direction empowers agents with tools for autonomous data analysis and evidence synthesis, transitioning them from communicators to actors. AI-HOPE \cite{yang2025ai} can interpret natural language commands into executable code, enabling it to autonomously analyze locally stored data for precision medicine research through tasks like association studies and survival analysis. Moreover, TxAgent \cite{gao2025tx} introduces ToolUniverse, a comprehensive suite of 211 specialized medical tools. By invoking these tools, the agent can retrieve and synthesize evidence from multiple sources, consider drug interactions and patient history, and iteratively refine treatment recommendations.

Addressing the heterogeneous nature of real-world medical data, recent efforts have focused on developing multi-modal diagnostic agents. MMedAgent \cite{li2024mmedagent} constructs a system where a Multi-modal Large Language Model (MLLM) acts as a planner, orchestrating a four-step process of user input interpretation, action planning, tool execution, and result aggregation, enhancing its tool-use proficiency via in-context learning. Similarly, MedAgent-Pro \cite{wang2025medagent} adopts a hierarchical agentic workflow. It first retrieves clinical guidelines using Retrieval-Augmented Generation (RAG) to formulate a diagnostic plan. It then employs sequential tool calls to analyze the patient's multi-modal data, generating a final report that includes diagnostic evidence.
To make disagreements within medical reasoning more explicit, MedLA~\cite{ma2026medla} connects premise extraction and question decomposition to logic trees constructed by medical agents. A credibility agent revisits uncertain nodes, so that collaboration can focus on the reasoning steps requiring further examination.

\leavevmode\phantomsection\label{par:design_healthcare}These approaches address two related demands of clinical reasoning: reconciling incomplete information and making the basis of a conclusion inspectable. Specialist deliberation exposes disagreements and supports consensus, at the cost of additional model calls~\cite{tang2024medagents,zhou2025llm}; retrieval and tool planning connect conclusions to records, guidelines and analyses~\cite{zuo2025kg4diagnosis,gao2025tx,wang2025medagent}. The choice depends on whether the main difficulty lies in coordinating interpretations or acquiring the evidence they require.

\subsubsection{Clinical Management and Automation}
Recently, in order to efficiently manage and analyze complex medical information, a growing number of research are focusing on adapting general-purpose agentic management and automation systems. For instance, ClinicalAgent \cite{yue2024clinicalagent} utilizes a hierarchical multi-agent architecture to predict clinical trial outcomes, assessing drug efficacy, safety, and patient recruitment feasibility based on external data sources and predictive models. Other systems target the automation of healthcare services; MedAide \cite{wei2024medaide} performs query rewriting via RAG and uses contextual encoding to identify fine-grained user intents. This process activates relevant agents to collaborate based on role assignments, delivering personalized diagnostic suggestions and department recommendations. Furthermore, Pandey et al. \cite{pandey2024advancing} explores the justification of prior authorizations by first reasoning over clinical guidelines and then employing a two-stage collaborative framework to decompose the problem into solvable sub-tasks for each agent.

However, directly implementing multi-agent collaboration can lead to challenges such as incompatible medical information flows and low component efficiency \cite{bedi2025optimization}. To enhance adaptability and dynamism, MDAgents \cite{kim2024mdagents} introduces a moderator agent that dynamically assembles appropriate multi-agent structures based on the complexity of the medical problem. This framework can configure specialized teams, such as a Primary Care Clinician, a Multidisciplinary Team, or an Integrated Care Team, and selectively employs reflection, iterative optimization, and collaborative methods to improve response accuracy. In order to organize additional scrutiny of difficult cases, Tiered Agentic Oversight (TAO)~\cite{kim2025tiered} routes cases through levels of automated review. Its separate clinician-in-the-loop study incorporates corrective feedback from physicians, distinguishing human review from the automated oversight stages discussed in Section~\ref{sec:oversight}. This trend towards structured and secure workflows is underscored by the push for systems that can navigate the complexities of sensitive data and automated decisions while adhering to regulatory standards \cite{neupane2025towards}.

\subsubsection{Environmental Simulation}
\begin{figure}
    \centering
    \includegraphics[width=1\linewidth]{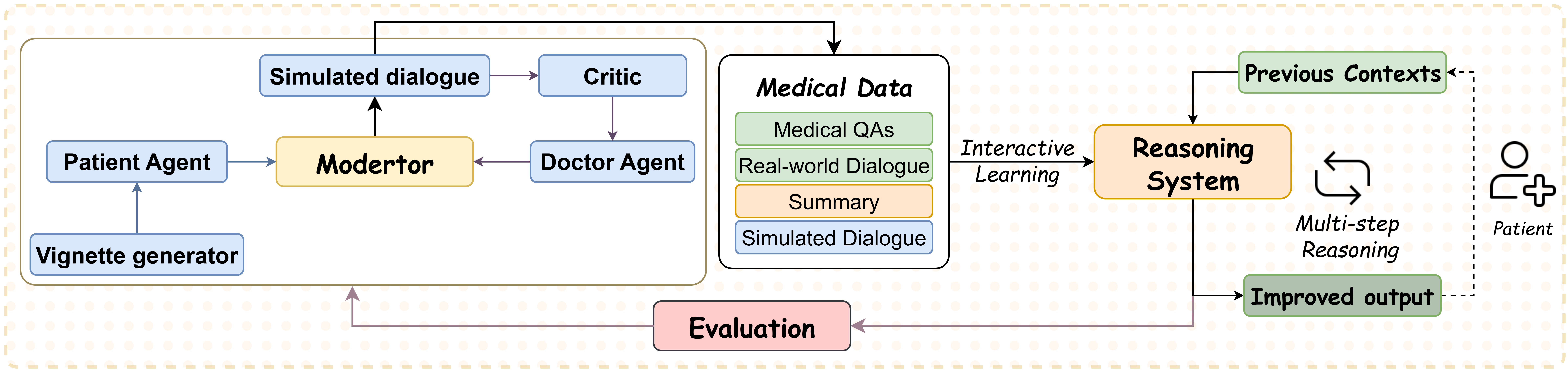}
\caption{AMIE's self-play and inference workflow~\cite{tu2025towards}. The inner loop refines simulated dialogues using critic feedback; the outer loop incorporates those dialogues into subsequent fine-tuning. At inference time, the system uses the conversation context and a chain-of-reasoning strategy to construct its next response. Adapted from Tu et al.~\cite{tu2025towards}.}
    \Description{AMIE uses an inner loop of simulated patient-doctor dialogue with critic feedback and an outer loop of model fine-tuning. The right-hand panel shows the separate inference-time conversation workflow.}

    \label{fig:AIME}
\end{figure}
Given that the healthcare domain is characterized by heterogeneous data which could be hard to collect, a significant line of research focuses on simulating realistic medical environments to enable agents to continuously optimize their performance through interactive learning. As a representative work, AMIE \cite{tu2025towards} simulates a diagnostic environment with physician, patient, and referee roles. By leveraging role-playing and CoT \cite{wei2022chain} strategies, it generates dialogue data for subsequent fine-tuning, with a separate chain-of-reasoning strategy at inference time, as illustrated in Fig. \ref{fig:AIME}. 

Moreover, AgentClinic \cite{schmidgall2024agentclinic} constructs a more complex multi-modal clinical simulation, focusing on patient interaction, data collection with incomplete information, and medical tool usage. The paradigm takes a major leap forward with Agent Hospital \cite{li2024agent}, which creates a simulated hospital where agents can evolve autonomously based on task resolution, learning from both successful treatments and failed cases without reliance on manually annotated data.

Beyond agent training, these simulated environments also serve as sophisticated testbeds for evaluation and as platforms for medical education. For instance, AI Hospital \cite{fan2025ai} establishes a dynamic evaluation environment by simulating four distinct roles (physician, patient, inspector, and director) and introduces the corresponding MVME benchmark. Its decentralized, role-based setup allows for the assessment of an agent's capabilities in symptom collection, examination recommendation, diagnosis, and dispute resolution. As for medical education, MEDCO \cite{wei2024medco} builds a collaborative learning system by simulating multi-disciplinary scenarios where student agents can interact with patient agents, expert physicians, and radiologists to proactively gather information and refine their diagnostic decisions. Besides, MedAgentSim \cite{almansoori2025self} presents a comprehensive simulation that requires agents to engage in multi-round, multi-modal interactions. It incorporates a self-improvement mechanism based on historical context and, notably, supports direct human interaction with the agents within the simulated environment.

These simulations supply interaction and evaluation settings where access to clinical data is limited~\cite{tu2025towards,schmidgall2024agentclinic,li2024agent}. However, success in simulated cases does not establish clinical validity. The separate clinician study in TAO illustrates how corrective human feedback can complement automated review~\cite{kim2025tiered}; the systems above remain research examples, with oversight and delegated authority requiring further examination (\S\ref{sec:oversight}).

\subsection{Software Engineering}
\label{se}

In contrast to Large Language Models specialized in code generation, agentic systems leverage a rich ecosystem of external tools to address a broader spectrum of software engineering tasks. This subsection examines the application of these agents in three key areas: \textit{code generation and testing, program repair,} and \textit{full-lifecycle software development.} 

\subsubsection{Code Generation and Testing}
\input{fig/table_codegen}
In the domain of code generation and testing, agentic systems significantly amplify the capabilities of LLMs beyond simple fine-tuning. These systems introduce structured collaboration and external tools, enhancing both the efficiency and reliability of code generation. Table~\ref{tab:codegen} demonstrates the performance of the selected agentic coding frameworks on popular benchmarks. A primary approach involves decomposing the coding process using multi-agent frameworks that emulate human programmer workflows. MapCoder \cite{islam2024mapcoder} decouples code generation into four collaborating LLM agents for retrieval, planning, coding, and debugging. The framework features a dynamic agent traversal model that adapts based on confidence scores from the planning phase, alongside plan-guided debugging and autonomous retrieval. Similarly, Almorsi et al. \cite{almorsi2024guided} implements a deliberately structured and fine-grained approach, utilizing LLMs as fuzzy searchers and approximate information retrievers. This multi-agent system effectively compensates for the inherent limitations of LLMs in long-sequence reasoning and long-context understanding. Moreover, AgentCoder \cite{huang2023agentcoder} pioneers a test-driven development (TDD) approach. It employs three distinct agents responsible for initial code generation, test case creation, and test execution with feedback. This test-driven iterative refinement loop enables the generation of higher-quality code with more efficient token usage.

Another major focus is augmenting LLMs with external tools to improve code quality and mitigate hallucination. CoCoST \cite{he2024cocost} introduces a framework where a task planner decomposes complex problems, an online search module acquires external knowledge, and a code generator iterates with a correctness tester to fix latent bugs. Moreover, CodeAgent \cite{zhang2024codeagent} integrates five distinct programming tools for repository-level code generation. Through rule-based tool usage, the system interacts with various software artifacts, iteratively performing information retrieval, code symbol navigation, and code testing. The authors also introduced CodeBench, a comprehensive benchmark for repository-level code generation, featuring code repositories across multiple domains.

\FloatBarrier

Further work has refined what the framework examines when a coding attempt fails. CodeSIM~\cite{islam2025codesim} simulates input--output behavior to check plans, while CollabCoder~\cite{doan2026collabcoder} uses joint diagnosis to decide whether to revise the plan or its implementation. MavenCoder~\cite{mavencoder2026} connects planning to estimated task difficulty and combines step-level verification with execution-based debugging. These developments bring feedback closer to the reasoning decision responsible for an error. The feedback itself also needs examination: Chen et al.~\cite{selfdebugtests2025} show how self-generated tests can misdirect repair, motivating a distinction between checks used during refinement and independent evaluation.

\subsubsection{Program Repair}
Complementary to code generation, automated program repair (APR) is another cornerstone of agent systems in software engineering. During the multi-step reasoning process, they can systematically understand code, localize faults, generate patches and validate fixes, often through sophisticated tool use and collaborative strategies. Table \ref{tab:coderep} shows the performance of selected APR agentic frameworks on popular software repair benchmarks.

\input{fig/table_repair}

A common strategy is to decompose the complex repair process into a structured workflow. RepairAgent \cite{bouzenia2025repairagent} formalizes this into four stages: defect information collection, fault localization, patch generation, and validation, using a state machine to dynamically select tools and adapt its strategy. Similarly, AgentFL \cite{qin2024agentfl} organizes method-level fault localization into comprehension, navigation and fault confirmation. Its final stage confirms a suspicious method rather than validating a repair. INTERVENOR \cite{wang2024intervenor} introduced an interactive ``repair chain'' concept, where a ``code teacher'' agent analyzes compiler errors to generate natural language suggestions, guiding a ``code learner'' agent in an iterative repair process. Agentless \cite{xia2024agentless} simplifies the process into three phases -- locate, fix, and validate -- using a staged workflow for hierarchical localization, patch generation and validation.

A critical sub-task within this workflow is precise fault localization. Several specialized agents have been developed for this purpose. OrcaLoca \cite{yuorcaloca} focuses on efficient localization by using dynamic priority scheduling and relevance scoring to prioritize actions, along with distance-aware context pruning to filter irrelevant code. LocAgent \cite{chen2025locagent} introduces a novel graph-based approach, parsing the codebase into a directed heterogeneous graph that captures structural dependencies, enabling an LLM agent to perform effective multi-hop reasoning for entity localization. Meanwhile, VulDebugger \cite{liu2025agent} utilizes both static and dynamic program analysis, continuously comparing the actual program state (observed via a debugger) with the expected state (inferred from constraints) to identify and rectify errors.

Building on these principles, recent frameworks aim to provide end-to-end solutions for real-world scenarios like resolving GitHub issues. MAGIS \cite{tao2024magis} employs a centralized architecture with four agents orchestrated by a central controller to manage task decomposition, file retrieval, code modification, and review. AutoCodeRover \cite{zhang2024autocoderover} further tackles GitHub issues by integrating structured code search with spectrum-based fault localization (SBFL) to pinpoint buggy methods. It then iteratively retrieves context via API calls and refines the issue description to synthesize a patch, demonstrating a robust solution in a practical setting. Besides, Rondon et al. \cite{rondon2025evaluating} explored the viability of agent-based program repair in an enterprise environment, contributing a valuable dataset of both human and machine reported bugs and offering insights into real-world applicability.

The organization of search provides a further choice within repository repair. SWE-Search~\cite{antoniades2025swesearch} combines Monte Carlo tree search with state-dependent feedback and a discriminator to explore alternative repair trajectories. Table~\ref{tab:coderep} records its comparison with an adapted base workflow and distinguishes localization hits, correct repairs and repository-level issue resolution. These measurements concern different stages of the repair process and should be interpreted with their reporting conditions.

\FloatBarrier
\subsubsection{Full-cycle Development}
Beyond discrete tasks like generation and repair, agentic systems are increasingly engineered to automate the entire software development lifecycle (SDLC), from initial requirements analysis to final testing and documentation. These systems often simulate human software teams and adopt established development methodologies.
Early explorations constructed virtual teams composed of agents with distinct roles. Dong et al. \cite{dong2024self} organize analyst, coder and tester agents managed by a waterfall model, communicating via a shared blackboard. Moreover, ChatDev \cite{qian2024chatdev} formalizes the process into design, coding, and testing phases, enabling role-specific agents to collaborate through natural language. It introduces a \textit{chat chain} for task refinement and a \textit{communicative de-hallucination} mechanism to ensure requirement clarity before coding. A significant advancement came with MetaGPT \cite{hongmetagpt}, which mimics the Standard Operating Procedures (SOPs) of human software companies. By assigning roles and enforcing structured workflows, MetaGPT facilitates hierarchical multi-agent collaboration across the full development pipeline, from requirements analysis to system design and testing, ensuring accurate information flow and reducing communication overhead. 

Furthermore, RepoAgent \cite{luo2024repoagent} specializes in automated code documentation. It analyzes project-level hierarchy and code dependencies to enrich LLM prompts and integrates with Git to maintain consistency between code and documentation in real-time. Similarly, DocAgent \cite{yang2025docagent} employs a multi-agent team -- comprising a reader, searcher, writer, validator, and coordinator, and uses topological code processing to automate the generation of comprehensive software documentation .

A more fundamental line of innovation focuses on how agents interact with their development environment. SWE-agent \cite{yang2024swe} pioneers an Agent-Computer Interface (ACI) that allows an agent to perform operations directly on a code repository, such as creating and editing files, navigating the filesystem, and executing test suites. Concurrently, OpenHands \cite{wangopenhands} imitates human developer interactions by providing agents with a sandboxed environment where they can write code, use a command-line interface, browse the web, and coordinate complex tasks, granting them a higher degree of autonomy. Besides, at the level of software system, MAAD \cite{zhang2025knowledge} tackles the complex process of software architecture design by creating a multi-agent system that learns from existing design knowledge, academic literature, and expert experience to generate and optimize architectures. Meanwhile SyncMind \cite{guosyncmind} identifies and addresses core challenges within these systems, such as the ``belief inconsistency'' problem in collaborative software engineering, and proposes SyncBench benchmarks to validate the solutions.

\leavevmode\phantomsection\label{par:design_software}Across these development tasks, executable feedback supports both a single repository controller and interacting generation/checking roles~\cite{yang2024swe,huang2023agentcoder,wang2024intervenor}. Full-cycle work adds dependencies among requirements, design, code and documentation, making explicit artifact handoffs useful for coordinating specialized responsibilities~\cite{hongmetagpt,qian2024chatdev,yang2025docagent}. These choices trade local control against communication and checking costs. Their benefits require the matched configurations documented by each reporting source in Tables~\ref{tab:codegen} and~\ref{tab:coderep}. The Darwin G\"odel Machine extends executable feedback to changes in the coding agent itself, with frozen foundation models~\cite{zhang2026dgm}; Section~\ref{sec:self_improving_systems} discusses this form of harness adaptation.

\subsection{Social and Economic Simulation}
\label{sec:socioeconomic}
\label{sec:so&eco}

Apart from the previous sections, the advent and rapid advancement of Large Language Models have also established a revolutionary paradigm for simulating social and economic behaviors. LLM-based agents, endowed with sophisticated, human-like capacities for reasoning, perception, and action, serve as the foundational elements of this new approach. Ranging from single-agent decision-making models to complex multi-agent systems, these frameworks are designed to construct and simulate critical socioeconomic dynamics at various scales. Consequently, they provide a powerful and versatile methodology for researchers to explore complex phenomena within the social and economic sciences. 

\subsubsection{Social Simulation}
\begin{figure}
    \centering
    \includegraphics[width=1\linewidth]{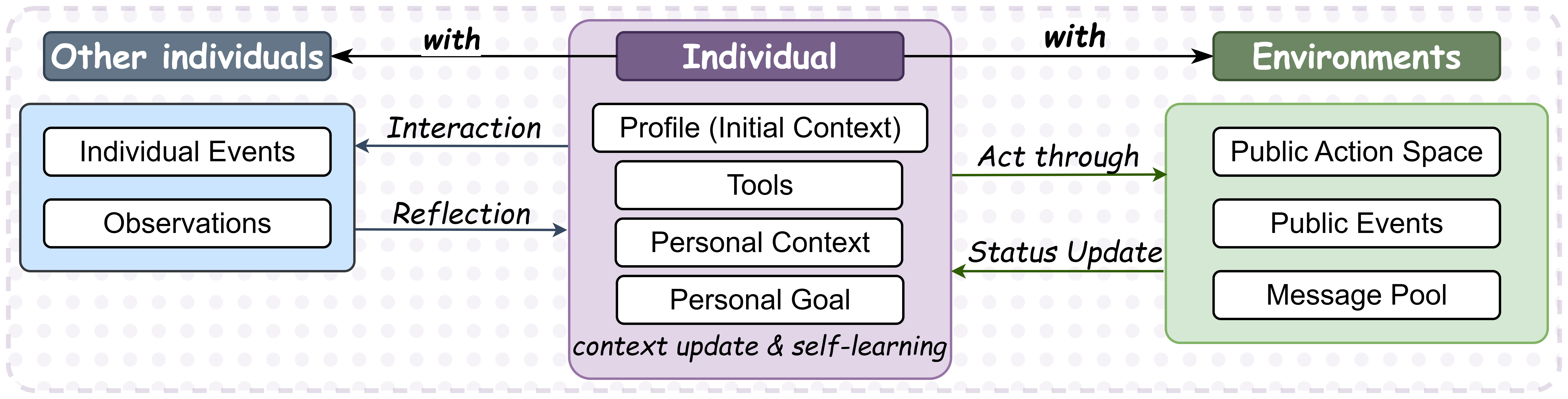}
\caption{The social-simulation paradigm based on an agentic reasoning framework. \textbf{\textit{(a) Individual (purple, center):}} Each agent has an initial profile, personal goals, tools and a context that changes as it observes and acts. \textbf{\textit{(b) Interaction with other individuals (blue, left):}} The agent observes other individuals and interaction events, then reflects on them to update its context and subsequent decisions. \textbf{\textit{(c) Interaction with the environment (green, right):}} Actions enter a public action space, where they can trigger public events or leave messages; status updates return the resulting changes to the agent. Together, the social and environmental feedback loops allow agents with distinct goals and contexts to behave differently under the same conditions and generate collective dynamics for evaluation.}

    \Description{This image presents the fundamental paradigm for social simulation using an Agentic Reasoning Framework. The diagram illustrates a central Individual agent with a profile, personal goals, and tools, which interacts with both Other Individuals and Environments.}
    \label{fig:socialpipe}
\end{figure}
LLM-based social simulation leverages the autonomy of agents to model a wide spectrum of emergent social behaviors. They could interact with different individuals which are also powered by LLMs, or dynamically act, learn and improve themselves from the simulated environments, as demonstrated in Fig. \ref{fig:socialpipe}, 
Early and influential research in this area often operates within text-based sandbox environments, where agents are assigned specific profiles and action spaces to drive interaction. One of the pioneer work is Generative Agents \cite{park2023generative}, which established this paradigm by creating an interactive sandbox with 25 agents, each possessing the capacity to observe, plan, and reflect, thus exhibiting distinct personalities. This setup allowed for the qualitative assessment of individual and collective behaviors through natural language interaction. Following this paradigms, subsequent studies have explored critical social dynamics. For instance, MetaAgents \cite{li2023metaagents} assessed coordination skills in a simulated job fair, while GOVSIM \cite{piatti2024cooperate} investigated whether agents could negotiate sustainable equilibria in a commons dilemma. Other research has used similar frameworks to study the formation of information cocoons, conformity \cite{wang2025user}, and voting behaviors in election scenarios \cite{zhang2024electionsim}.

Moreover, the focus expanded from specific simulations to the development of generalizable platforms and the modeling of online interactions. To enhance reusability and extensibility, frameworks like GenSim \cite{tang2025gensim} and Sotopia-S4 \cite{zhou2025sotopia} were introduced. These platforms provide configurable systems where researchers can define roles, contextual information, and action spaces to customize diverse multi-agent scenarios and test hypotheses with tailored evaluation metrics. Besides, research began to address the complexities of online social networks. BASES \cite{ren2024bases} modeled the emergent web search behaviors of diverse user profiles, while Mou et al. \cite{mou2024unveiling} simulated the propagation of influence in social movements by differentiating between core agent-driven users and peripheral users. The challenge of simulating malicious online environments was tackled by BotSim \cite{qiao2025botsim}, which created a mixed network of agents and human users. Similarly, Y-Social \cite{rossetti2024social} employed digital twin technology to replicate the dynamics of user interactions on social media platforms.

\begin{table}[!t]
\centering
\captionsetup{skip=3pt}
\caption{Social simulation methods and their reported scale. \# denotes the reported number of agents or users, retaining each study's original unit (simulated agents, accounts or source-user pools). k = $10^3$, M = $10^6$; an en dash indicates an unspecified count.}
\label{tab:social}
\begingroup
\fontsize{7}{8.4}\selectfont
\setlength{\tabcolsep}{4pt}\renewcommand{\arraystretch}{1.12}
\begin{tabularx}{\linewidth}{@{}>{\RaggedRight\arraybackslash}p{0.30\linewidth}>{\RaggedRight\arraybackslash}X>{\centering\arraybackslash}p{0.18\linewidth}>{\raggedleft\arraybackslash}p{0.09\linewidth}@{}}
\toprule
\rowcolor{lightblue}
\textbf{Related work} & \textbf{Simulation focus} & \textbf{Data source} & \textbf{\#} \\
\midrule
GOVSIM~\cite{piatti2024cooperate} & Sustainable cooperation & Simulated & -- \\
MetaAgents~\cite{li2023metaagents} & Job fair & Simulated & -- \\
Generative Agents~\cite{park2023generative} & Interactive human behavior & Simulated & 25 \\
RecAgent~\cite{wang2025user} & Recommendation system & Simulated & 20 \\
BotSim~\cite{qiao2025botsim} & Malicious social botnet & Simulated & $\sim$3k \\
S3~\cite{gao2023s} & Population-level interaction & Simulated & $\sim$10k \\
ElectionSim~\cite{zhang2024electionsim} & Massive-population election & Real-world & 10k \\
BASES~\cite{ren2024bases} & Web-search user simulation & Real-world & 200k \\
SocioVerse~\cite{zhang2025socioverse} & World model & Real-world & 10M \\
Sotopia-S4~\cite{zhou2025sotopia} & General social simulation & Simulated & 150 \\
GenSim~\cite{tang2025gensim} & General social simulation & Simulated & $\sim$100k \\
Y-Social~\cite{rossetti2024social} & Social media & Real-world & -- \\
Mou et al.~\cite{mou2024unveiling} & Social media & Real-world & 1k \\
OASIS~\cite{yang2024oasis} & Social media & Real-world & 100k \\
\bottomrule
\end{tabularx}
\endgroup
\end{table}

A primary challenge in social simulation is achieving realism at a large scale, and recent systems have made solid progress in this direction. OASIS \cite{yang2024oasis} implements large-scale user simulation by integrating dynamic context updates and an interest-based recommendation system modeled after real-world platforms. SocioVerse \cite{zhang2025socioverse} further enhances scalability by introducing and restructuring real-world information to create distinct contextual environments that drive divergent individual behaviors. In addition, S3~\cite{gao2023s} models individual users and derives population-level information, emotion and attitude dynamics from their interactions. GA-S3~\cite{zhang2025gas3}, a distinct system, instead constructs agents representing groups of behaviorally similar individuals, providing a more economical unit of simulation. MF-LLM~\cite{mi2025mfllm} introduces a complementary mean-field structure in which individual decisions update a population signal that then conditions subsequent individual actions. These approaches extend the original individual-agent setting through different representations of collective behavior. Table~\ref{tab:social} summarizes these works by simulation focus, data source and reported scale.

\subsubsection{Economic Perception and Simulation}
Agentic frameworks are increasingly being leveraged to perceive, analyze, and simulate complex economic markets, a domain with immense real-world value \cite{cao2025deep}. The evolution of these frameworks can be understood through three advancing frontiers: enhancing agent-native cognitive abilities, optimizing collaborative structures, and creating large-scale market simulations.

Initially, efforts focused on equipping individual agents with sophisticated cognitive modules. Early work like FinMem \cite{yu2024finmem} established a foundational agent system with analysis, memory, and decision modules, where the memory component utilized reflection and interactive learning to process historical context. Similarly, EconAgent \cite{li2024econagent} situated an LLM-based agent within a simulated macroeconomic environment, employing perception, memory, and reflection modules to analyze its behavior. This focus on reflection as a core cognitive faculty was further advanced by subsequent systems. FinVision \cite{fatemi2024finvision} places systematic reflection at its center, using a multi-agent system to analyze historical trading signals and generate feedback to improve future decisions. More recently, Zhang et al. \cite{zhang2024multimodal} introduced a sophisticated two-level reflection module. This module processes multimodal market information to establish distinct causal links between market data and price movements, while also reflecting on historical trading performance. Its memory system separately stores parsed market information and the insights from both levels of reflection to inform decision-making.

Building upon individual agent capabilities, another line of research explores the optimization of collaborative structures. As a pioneer work, FinCon \cite{yu2024fincon} addresses stock trading and portfolio management using a hierarchical manager-analyst multi-agent system. This structure facilitates synchronous collaboration and employs self-criticism to monitor market risks and update investment theses, thereby achieving robust risk control. Notably, conceptualized beliefs are only selectively communicated among agents, effectively reducing the communication overhead typical in multi-agent systems. The importance of structured collaboration is also highlighted in TradingAgents \cite{xiao2025tradingagents}, which enhance automated trading performance through explicit role and objective allocation combined with streamlined information integration.

The most ambitious application in this domain involves creating comprehensive, closed-loop platforms and simulating entire economic environments. FinRobot \cite{yang2024finrobot} implements a full-cycle financial analysis platform. At its agent layer, it uses CoT \cite{wei2022chain} to deconstruct complex financial problems, dynamically selecting or fine-tuning different LLMs and applying varied algorithms based on the task, thus enabling rapid market response. Pushing the boundaries further, large-scale simulations now aim to replicate real-world market dynamics. StockAgent \cite{zhang2024ai} deploys a massive multi-agent system to simulate a stock trading environment, allowing users to assess how external factors influence investor behavior and profitability. Similarly, Gao et al. \cite{gao2024simulating} simulates the stock market by creating agents equipped with unique profiles, observational capabilities, and tool-based learning. By integrating these agents with a realistic order-matching system, these frameworks achieve a high-fidelity simulation of actual stock market operations, opening new avenues for economic research and policy testing.
Building on these market environments, TwinMarket~\cite{yang2025twinmarket} couples investor decisions with a trading environment and a simulated social network. Information exchange can therefore affect market actions, while market outcomes feed back into individual decisions, extending the simulation from isolated trading policies to interactions between social and economic dynamics.

\leavevmode\phantomsection\label{par:design_simulation}Across social and economic simulations, agents' behavior is itself the object of study. Preserving private goals and local information is therefore essential; imposing consensus would change the phenomenon being modeled~\cite{park2023generative,piatti2024cooperate,zhang2024ai,gao2024simulating}. Individual agents with memory suit questions about actors' decisions, whereas group or population representations support the study of aggregate dynamics~\cite{yu2024finmem,li2024econagent,zhang2025gas3,mi2025mfllm}. Trading systems additionally use delayed returns to support reflection and risk control~\cite{fatemi2024finvision,yu2024fincon}. Evaluation must consequently examine agreement with observed behavior or market data. The scale and data sources in Table~\ref{tab:social} describe the simulation setting, but do not by themselves establish its validity.

\FloatBarrier
\subsection{General-Purpose Agents}
\label{sec:general_agents}
General-purpose agents extend agentic reasoning across information seeking, computation and digital workflows within a reusable system. We discuss these capabilities through \textit{task-oriented agentic systems}, which organize reasoning and action to complete a task, and \textit{self-improving agentic systems}, which adapt their resources and procedures through accumulated experience. These two capabilities can coexist in the same framework.

\subsubsection{Task-Oriented Agentic Systems}
\label{sec:task_oriented}
General-purpose tasks often require agents to gather information beyond the knowledge available in their initial context. The reasoning process therefore needs to determine both what remains unknown and how to obtain the missing evidence. WebDancer~\cite{wu2025webdancer} interleaves search and browsing in a ReAct-style loop, allowing observations from one page to guide the next search or navigation action. This interaction makes information acquisition part of the evolving solution process, extending retrieval beyond a preliminary step that supplies a fixed set of documents~\cite{singh2025agenticrag}. As tasks expand from answering questions to producing research reports, the emerging artifact can itself reveal gaps that require further investigation. WebThinker~\cite{li2025webthinker} incorporates a deep web explorer and report drafting into its reasoning process, enabling the agent to alternate between resolving knowledge gaps and developing the report. In this way, evidence gathering and synthesis inform one another throughout execution, rather than following a fixed retrieve-then-write sequence.

\input{fig/table_general_agents}
When a task involves several sources, tools and intermediate deliverables, a single sequence of reasoning steps may obscure which parts can proceed independently and which depend on earlier results. A common approach is to separate planning over the whole task from execution within individual subtasks. Workforce~\cite{hu2025owl}, the inference framework associated with OWL, adopts this organization through a domain-agnostic planner, a coordinator that manages subtasks, and workers responsible for specialized tool interactions. This division allows the overall plan to remain focused on task dependencies while workers handle the details of their local environments. Building on the need to manage these dependencies explicitly, Flash-Searcher~\cite{qin2026flashsearcher} represents execution as a directed acyclic graph, allowing independent reasoning paths to run concurrently and revising the graph using intermediate results. FlowSearch~\cite{hu2026flowsearch} further organizes exploration through an expanding knowledge flow, combining hierarchical decomposition with feedback-driven revisions to the search structure. Across these designs, coordination concerns how partial results become available to subsequent reasoning, as well as how work is distributed among agents.

As agents gather information and complete subtasks, later reasoning depends on the claims and candidate solutions they produce. Verification helps determine whether these intermediate results can be used as they stand or need revision. DeepVerifier~\cite{wan2026deepverifier} derives verification rubrics from a taxonomy of deep-research failures and returns detailed feedback for iterative answer refinement. Here, the verifier directs revision toward identifiable deficiencies in an existing response. A complementary approach is to preserve alternative reasoning routes before deciding which result to accept. ReThinker~\cite{zhang2026rethinker} organizes this process through solver, critic and selector stages, combining guided reflection with confidence control. PoTRE~\cite{kankariya2026potre} introduces diversity at the level of reasoning structure, drawing on agents with different exploration and refinement strategies before task-adaptive aggregation. These designs connect revision to either a diagnosed weakness or an alternative route through the task. When candidate solutions are executable, the environment can provide a more direct basis for this decision. AlphaEvolve~\cite{novikov2025alphaevolve} combines LLM-generated code changes with automated evaluators and a database of previous programs, allowing evaluation results to inform subsequent proposals. Its evolving programs illustrate how iterative reasoning can be organized around verifiable task artifacts; improving such a solution does not require modifying the agent's own controller.

Bringing these capabilities into a reusable system requires a common execution framework that maintains task state across planning, tool use and revision. JoyAgent-JDGenie~\cite{liu2025joyagent} combines planning and execution agents with critic voting, supported by working, semantic and procedural memory. This organization connects the information needed for the current step to knowledge and procedures that can support a longer workflow. MiroFlow~\cite{wang2026miroflow} provides flexible orchestration through an agent graph, an optional deep reasoning mode and workflow execution support, allowing different reasoning configurations to operate within the same framework. Long-running research also places pressure on the context supplied to each model call. Tongyi DeepResearch~\cite{tongyi2025deepresearch} supports extended information seeking through its agentic model and accompanying execution framework, while ACON~\cite{kang2026acon} examines the context-management problem directly by compressing observations and history and refining its compression guidelines through failure analysis. These systems show how memory organization and context construction support the continuity of reasoning across tool interactions. Table~\ref{tab:general_agents} presents the corresponding system evaluations across information seeking, reasoning, terminal operation and professional-artifact tasks~\cite{mialon2024gaia,wei2025browsecomp,cais2026hle,merrill2026terminalbench,sun2026ale}.

\subsubsection{Self-Improving Agentic Systems}
\label{sec:self_improving_systems}
While task-oriented systems revise their actions as a task unfolds, accumulated experience also provides a basis for adapting subsequent reasoning. An initial approach is to retain useful cases and recover them when a similar decision arises. Memento~\cite{zhou2025memento} incorporates case memory into its planning process, while ExpeL~\cite{zhao2024expel} extracts insights from past trajectories and retrieves relevant experience for new tasks. As this memory grows, adaptation also depends on how new information relates to what the system already retains. A-MEM~\cite{xu2025amem} constructs linked notes and allows new memories to revise related entries, whereas Mem0~\cite{chhikara2025mem0} separates the extraction of salient information from updates to existing memory. These mechanisms support the maintenance of useful context across interactions. Extending this process to reusable instructions and strategies, Agentic Context Engineering~\cite{zhang2026ace} combines generation, reflection and curation to update context incrementally. Together, these approaches connect experience reuse to the organization of the next reasoning context: adaptation occurs through selecting, revising and recovering external knowledge, without necessarily changing the underlying language model.

\input{fig/table_self_evolving}
Beyond retaining what happened in a previous task, agents can also abstract how a useful sequence of actions was carried out. Agent Workflow Memory~\cite{wang2025awm} induces reusable workflows from interaction trajectories, enabling later reasoning to draw on procedures that span multiple steps. This shifts reuse from isolated observations toward recurring patterns of action. Memento-Skills~\cite{zhou2026mementoskills} develops this direction through a read--write loop: the system selects skills for the current state and updates or expands its skill library using new experience. When a reusable procedure requires additional external capabilities, tool construction can become part of the same adaptation process. Alita~\cite{qiu2025alita} equips the agent to construct reusable tools, allowing successful solutions to contribute resources for future tasks. Related work extends the construction and revision of agents, skills and tools across broader task settings~\cite{tang2026autoagent,qiu2025alitag,cheng2026mem2evolve,liu2026roga,tang2026msce}. The resulting framework must connect the discovery of a useful procedure to its later selection and execution, so that the growing library can influence how subsequent tasks are solved.

Adaptation can further extend from the resources used during reasoning to the controller that organizes their use. This requires exposing parts of the framework as objects that can be evaluated and revised. TPGO~\cite{he2026tpgo} represents agents, tools and workflows as a textual parameter graph and uses execution feedback to guide changes to its components. EvoRoute~\cite{zhang2026evoroute} addresses model allocation within the agent system, while the Darwin G\"odel Machine~\cite{zhang2026dgm} expands the editable space to the agent's own code with frozen foundation models. Program evolution can also search directly over reasoning procedures. In its mathematical reasoning application, ShinkaEvolve~\cite{lange2026shinkaevolve} discovers a scaffold that combines independent solution generation, critical review and synthesis. Unlike evolving a program that solves one computational problem, the candidate here specifies a reusable organization of reasoning. As the editable space expands, identifying which part of an execution caused a failure becomes central to proposing useful changes. Recent harness optimizers therefore use execution traces to guide controller revision~\cite{chen2026harnessfix,zhang2026selfharness,chen2026harnessx}. Meta-Harness~\cite{lee2026metaharness} lets a proposer selectively inspect previous candidates' code, evaluation results and traces through a filesystem. AHE~\cite{lin2026ahe} organizes trajectory evidence for inspection and exposes harness components as editable files, pairing proposed changes with predicted effects that are checked against later outcomes. Both connect execution evidence to concrete program edits. Meta Context Engineering~\cite{ye2026mce} carries this adaptation into the improvement procedure itself by co-evolving context artifacts and the skills used to construct them, linking harness optimization to the recursive reasoning agenda in \S\ref{sec:recursive}.

An evolving framework must also determine which updates to retain and when to reuse them. Self-Harness~\cite{zhang2026selfharness} uses held-out tasks in its promotion gate, making validation part of the procedure that selects a revised controller. When labeled validation data are unavailable, RHO~\cite{pan2026rho} instead revisits challenging past tasks and compares candidate updates through self-validation, self-consistency and pairwise self-preference. Maintaining alternatives offers another way to accommodate differences between tasks, including semantic search over stored candidates and adaptive selection from a harness tree~\cite{luo2026harnessbank,liu2026adaptiveharness,zhang2026aevo}. This introduces a distinction between constructing an improved configuration and selecting one that fits the next workload. Transfer raises a related question about which parts remain useful when the underlying model changes; HarnessCompass~\cite{zhang2026harnesscompass} examines this through reuse of an unchanged harness across task models. Evaluation environments for self-evolving agents make these update and selection procedures explicit~\cite{zheng2026seagym,gao2026evoagentbench,huang2026evobench}, while broader interactive and professional-task benchmarks provide additional settings in which to examine their applicability~\cite{froger2026gaia2,barres2025tau2,patwardhan2025gdpval,yao2026harnessbench}. Table~\ref{tab:self_evolving} retains the reported model, feedback and evaluation conditions needed to interpret adaptation and transfer across these settings~\cite{kapoor2025hal}.

\leavevmode\phantomsection\label{par:design_general}This variety of update mechanisms also creates a common evaluation problem: improvements on tasks used to select a memory, skill or controller need not transfer to later workloads. Executable checks suit some tasks, while open-ended information and artifact tasks require other forms of review~\cite{merrill2026terminalbench,sun2026ale}. Tables~\ref{tab:general_agents} and~\ref{tab:self_evolving} therefore retain reporting sources, task models and selected evaluation conditions. Differences in search budgets, update selection and transfer settings must be resolved before attributing gains to a framework~\cite{kapoor2025hal}.

\subsection{Others}
\label{sec:others}
Beyond the applications in mainstream scenarios, \textit{Agentic Reasoning Frameworks} also demonstrate distinct potential in several other significant domains. A growing body of work explores the frontiers of their capabilities in complex tasks like \textit{Embodied Interaction}, \textit{GUI Operation}, and \textit{Strategic Reasoning}.

In \textit{Embodied Interaction}, an agent's reasoning process typically follows a perceive-plan-execute-memory cycle. The primary goal in this field is to empower agents to learn and complete tasks autonomously through continuous interaction with a virtual or physical environment \cite{liu2025aligning}. For example, based on training-free framework, Voyager \cite{wang23voyager} enables an agent to achieve lifelong learning in the game \textit{Minecraft} by designing a specialized action space and a reasoning framework that integrates iterative feedback and continuous evolution. However, existing general foundational Multimodal Large Language Models (MLLMs) still show deficiencies in handling complex, multi-level task planning. Consequently, a prevailing research direction is to fine-tune base models for specific embodied scenarios to enhance the agent's perception and planning capabilities in a tailored manner \cite{liu2025aligning}. Beyond individual skill storage, GATE~\cite{luo2026gate} organizes tools in a graph with semantic retrieval and invocation links. Its tool library evolves during acquisition and is frozen in its transfer tests.

\textit{GUI (Graphical User Interface) Operation} aims to enable agents to operate applications on phones, computers, and the web with human-like proficiency \cite{zheng2024gpt}. Early frameworks were built on vision-language models such as GPT-4V \cite{yan2023gpt}. They enhance reasoning frameworks by integrating visual memory and knowledge bases and introducing multi-agent collaboration mechanisms \cite{zhang2025appagent,lee2024mobilegpt,wang2024mobile,wang2025mobile}. These are combined with conventional components like reflection and context updating to effectively decompose complex GUI navigation tasks and enable continuous learning. However, as task complexity increases, the research focus is shifting from relying on the zero-shot capabilities of models towards more intensive, specialized training via Supervised Fine-Tuning (SFT) or Reinforcement Learning (RL) \cite{wang2024gui}. Subsequent systems further develop how execution feedback revises plans. Agent S2~\cite{agashe2025agents2} coordinates specialists with replanning at multiple time scales, while Agent S3~\cite{gonzalez2025agents3} adds a Behavior Judge over observed action effects. Agents' Last Exam~\cite{sun2026ale} extends evaluation to professional workflows through artifact checks, including GUI-assisted Codex, OpenClaw and ALE-Claw configurations.

\textit{Strategic Reasoning} requires an agent to understand and predict the behavior of opponents and dynamically adapt its own strategy, with the core challenge being the management of dynamics and uncertainty in multi-agent interactions. An important research direction in this area is creating diverse test environments to accurately evaluate and enhance the strategic capabilities of LLMs \cite{zhangllm}. These environments span a wide range, as evidenced by the development of benchmarks based on real-time strategy (RTS) games like \textit{StarCraft II} \cite{ma2024large}; the use of LLMs to generate expert-level decision explanations for board games \cite{kim2025bridging}; and the systematic analysis of their behavioral rationality through classic game theory models \cite{fan2024can}.

\subsection{Synthesis Across Scenarios}
\label{sec:scenario_synthesis}
\input{fig/table_cross_scenario}
The preceding scenarios show how task requirements influence the organization of agentic reasoning. Software agents can check proposed patches through execution, whereas research agents may need experimental observations before revising a hypothesis. Clinical workflows also require decisions about who reviews consequential actions. Table~\ref{tab:cross_scenario} brings these requirements together and relates them to organization, tool use and oversight.

The feedback available to an agent shapes how it revises its reasoning. Repository execution and symbolic solvers provide checks within a task~\cite{yang2024swe,pan2023logiclm}, while scientific observations can inform later research rounds~\cite{roohanibiodiscoveryagent}. The presence of a check, however, does not ensure reliable feedback. Generated tests can themselves misdirect code repair~\cite{selfdebugtests2025}. How feedback becomes available also helps explain the tool-utilization modes in \S\ref{sec:tool_based}. Checks that are cheap to repeat support iterative utilization. When each result must be interpreted before the next call, sequential utilization preserves that dependency. Independent sources can instead be queried in parallel and their results combined afterwards, as in the subtask graph of Flash-Searcher~\cite{qin2026flashsearcher}. The choice depends on what feedback can establish, when it becomes available and what another check costs.

These dependencies also affect how work is distributed within a framework. SWE-agent gives one controller access to repository tools~\cite{yang2024swe}, whereas MetaGPT exchanges structured artifacts among specialized roles and Workforce separates subtask coordination from local tool use~\cite{hongmetagpt,hu2025owl}. Separate roles are useful when information or control responsibilities can be clearly assigned and intermediate artifacts checked. An executable tool graph offers another way to make dependencies explicit~\cite{bao2026spatialagent}. Tightly coupled steps may instead favor a single controller, since dividing them introduces communication and verification costs.

Distributing work across agents then raises the question of whether they are pursuing the same outcome. Medical logic trees organize disagreements around a shared reasoning task~\cite{ma2026medla}. In social simulations and strategic games, however, private objectives and unequal information may be essential to the behavior being studied~\cite{piatti2024cooperate,fan2024can}. Requiring these agents to reach consensus would alter that behavior. Coordination therefore takes on a different meaning across these settings, and the choice of individual, group or population representations must preserve the distinctions relevant to the task~\cite{gao2023s,zhang2025gas3,mi2025mfllm}.

Whether agents collaborate or pursue separate objectives, the consequences of their actions also determine how much oversight is needed. Sandboxed execution and tool permissions bound the effects of autonomous choices~\cite{wangopenhands,shi2025progent}. Decisions that exceed delegated authority require human review (\S\ref{sec:oversight}), so the allocation of responsibilities must also specify when an agent should hand control to a person. Sections~\ref{sec:safety}--\ref{sec:safety_eval} develop these safeguards and their evaluation. These relationships offer conditional guidance for framework design, with performance benefits requiring assessment under matched tasks and resource conditions~\cite{pan2025multiagent,kapoor2025hal}.

%% file: fig/scenario.tex
\definecolor{paired-light-blue}{RGB}{198, 219, 239}
\definecolor{paired-dark-blue}{RGB}{49, 130, 188}
\definecolor{paired-light-orange}{RGB}{251, 208, 162}
\definecolor{paired-dark-orange}{RGB}{230, 85, 12}
\definecolor{paired-light-green}{RGB}{199, 233, 193}
\definecolor{paired-dark-green}{RGB}{49, 163, 83}
\definecolor{paired-light-purple}{RGB}{218, 218, 235}
\definecolor{paired-dark-purple}{RGB}{117, 107, 176}
\definecolor{paired-light-gray}{RGB}{217, 217, 217}
\definecolor{paired-dark-gray}{RGB}{99, 99, 99}
\definecolor{paired-light-pink}{RGB}{222, 158, 214}
\definecolor{paired-dark-pink}{RGB}{123, 65, 115}
\definecolor{paired-light-red}{RGB}{231, 150, 156}
\definecolor{paired-dark-red}{RGB}{131, 60, 56}
\definecolor{paired-light-yellow}{RGB}{231, 204, 149}
\definecolor{paired-dark-yellow}{RGB}{141, 109, 49}
\definecolor{light-green}{RGB}{118, 207, 180}
\definecolor{raspberry}{RGB}{228, 24, 99}

\definecolor{paired-light-gray}{RGB}{232, 234, 237}
\definecolor{paired-dark-blue}{RGB}{167, 205, 225}
\definecolor{paired-dark-orange}{RGB}{236, 179, 144}
\definecolor{paired-dark-green}{RGB}{188, 214, 191}
\definecolor{paired-light-red}{RGB}{209, 196, 233}

\tikzset{%
    root/.style =
    {align=center,text width=3cm,rounded corners=3pt, line width=0.5mm, fill=paired-light-gray!50,draw=paired-dark-gray!90},
    Scientific/.style =
    {align=center,text width=4cm,rounded corners=3pt, fill=paired-light-blue!50,draw=paired-dark-blue!80,line width=0.4mm},
    Health/.style =
    {align=center,text width=4cm,rounded corners=3pt, fill=paired-light-orange!50,draw=paired-dark-orange!80,line width=0.4mm},
    Software/.style =
    {align=center,text width=4cm,rounded corners=3pt, fill=paired-light-green!50,draw=paired-dark-green!80, line width=0.4mm},
    Social/.style =
    {align=center,text width=4cm,rounded corners=3pt, fill=paired-light-red!35,draw=paired-light-red!90, line width=0.4mm},
    General/.style =
    {align=center,text width=4cm,rounded corners=3pt,fill=light-green!18,draw=light-green!80,line width=0.4mm},
    Others/.style =
    {align=center,text width=4cm,rounded corners=3pt,fill=paired-light-yellow!30,draw=paired-dark-yellow!60,line width=0.4mm},
    subsection/.style =
    {align=center,text width=3.5cm,rounded corners=3pt},
}

\begin{figure*}[!t]
\centering
\newsavebox{\scenarioinventorybox}
\savebox{\scenarioinventorybox}{%
\resizebox{\textwidth}{!}{%
\begin{forest}
for tree={
    edge path={\noexpand\path[\forestoption{edge}] (!u.east) -- ++(6pt,0) |- (.west)\forestoption{edge label};},
    grow'=0, draw, rounded corners, node options={align=center},
    font=\fontsize{9}{9.5}\selectfont,
    s sep=3pt, l sep=10pt, inner sep=2pt,
    calign=child edge, calign child=(n_children()+1)/2,
},
[Application\\Scenarios, root, text width=2.7cm
    [{Scientific Research (\S\ref{sec:sci})}, Scientific, text width=3.6cm
        [{Math}, Scientific, text width=3.6cm
            [{LLM4ED~\cite{du2024llm4ed}, Optimus~\cite{ahmaditeshnizi2024optimus}, MetaOpenFOAM~\cite{chen2024metaopenfoam}, OptimAI~\cite{thind2025optimai}, MA-LoT~\cite{wang2025ma}, LeanAgent~\cite{kumarappanleanagent}, Prover Agent~\cite{baba2025prover}, MathSensei~\cite{das2024mathsensei}, MetaOpenFOAM 2.0~\cite{chen2025metaopenfoam}}, Scientific, text width=14cm, node options={align=left}]
        ]
        [{Astrophysics}, Scientific, text width=3.6cm
            [{AstroAgents~\cite{saeediastroagents}, The AI Cosmologist~\cite{moss2025ai}, Laverick et al.~\cite{laverick2024multi}, AutoGen~\cite{wu2024autogen}}, Scientific, text width=14cm, node options={align=left}]
        ]
        [{Geoscience}, Scientific, text width=3.6cm
            [{Li et al.~\cite{li2025giscience}, Autonomous GIS~\cite{li2023autonomous}, Wang et al.~\cite{wang2024mitigating}, Ning et al.~\cite{ning2025autonomous}, Pantiukhin et al.~\cite{pantiukhin2025accelerating}, GeoAgent~\cite{huang2024geoagent}, GeoMap-Agent~\cite{huang2025peace}, GeoLLM-Squad~\cite{lee2025multi}, Spatial-Agent~\cite{bao2026spatialagent}, GeoLLM-Engine~\cite{singh2024geollm}, AutoGen~\cite{wu2024autogen}}, Scientific, text width=14cm, node options={align=left}]
        ]
        [{Biochemistry \& Material Science}, Scientific, text width=3.6cm
            [{ChatDrug~\cite{liu2024conversational}, LIDDIA~\cite{averly2025liddia}, DrugAgent~\cite{inoue2025drugagent}, PharmAgents~\cite{gao2025pharmagents}, CLADD~\cite{lee2025rag}, ChatChemTS~\cite{ishida2025large}, BioDiscoveryAgent~\cite{roohanibiodiscoveryagent}, CRISPR-GPT~\cite{qu2025crispr}, BioAgents~\cite{mehandru2025bioagents}, AutoBA~\cite{zhou2024ai}, BioMaster~\cite{su2025biomaster}, Chemist-X~\cite{chen2023chemist}, Ruan et al.~\cite{ruan2024automatic}, ChemCrow~\cite{bran2023chemcrow}, El Agente Q~\cite{zou2025agente}, MOOSE-Chem~\cite{yangmoose}, Stewart and Buehler~\cite{stewart2025molecular}, ChatMOF~\cite{kang2024chatmof}, LLaMP~\cite{chiang2025llamp}, AtomAgents~\cite{ghafarollahi2025automating}, ProtAgents~\cite{ghafarollahi2024protagents}, TourSynbio-Search~\cite{liu2024toursynbio}, BioResearcher~\cite{luo2025intention}, CACTUS~\cite{mcnaughton2024cactus}, ChemHTS~\cite{li2025chemhts}}, Scientific, text width=14cm, node options={align=left}]
        ]
        [{General Research}, Scientific, text width=3.6cm
            [{AutoSurvey~\cite{wang2024autosurvey}, SurveyX~\cite{liang2025surveyx}, SurveyForge~\cite{yan2025surveyforge}, Agent Laboratory~\cite{schmidgall2025agent}, The AI Scientist~\cite{lu2024ai}, The AI Scientist v2~\cite{lu2024ai}, AI-Researcher~\cite{tang2025ai}, Robin~\cite{ghareeb2025robin}, PaperCoder~\cite{seo2025paper2code}, ResearchAgent~\cite{baek2025researchagent}, VirSci~\cite{su-etal-2025-many}, Co-Scientist~\cite{gottweis2025towards}, SciAgents~\cite{ghafarollahi2025sciagents}, DOLPHIN~\cite{yuan2025dolphin}, CycleResearcher~\cite{wengcycleresearcher}, SciMaster~\cite{chai2025scimaster}, AgentRxiv~\cite{schmidgall2025agentrxiv}, IdeaSynth~\cite{pu2025ideasynth}, Gao et al. (citation-grounded generation)~\cite{gao2023enabling}}, Scientific, text width=14cm, node options={align=left}]
        ]
    ]
    [{Healthcare (\S\ref{sec:healthcare})}, Health, text width=3.6cm
        [{Diagnosis Assistance}, Health, text width=3.6cm
            [{MedAgents~\cite{tang2024medagents}, Wang et al.~\cite{wang2025empowering}, Chen et al.~\cite{chen2025enhancing}, RareAgents~\cite{chen2024rareagents}, KG4Diagnosis~\cite{zuo2025kg4diagnosis}, AI-HOPE~\cite{yang2025ai}, TxAgent~\cite{gao2025tx}, MMedAgent~\cite{li2024mmedagent}, MedAgent-Pro~\cite{wang2025medagent}, MedLA~\cite{ma2026medla}, MD2GPS~\cite{zhou2025llm}}, Health, text width=14cm, node options={align=left}]
        ]
        [{Clinical Management and Automation}, Health, text width=3.6cm
            [{ClinicalAgent~\cite{yue2024clinicalagent}, MedAide~\cite{wei2024medaide}, Pandey et al.~\cite{pandey2024advancing}, MDAgents~\cite{kim2024mdagents}, TAO~\cite{kim2025tiered}, Neupane et al. (HIPAA-compliant agents)~\cite{neupane2025towards}, Bedi et al. (clinical multi-agent coordination)~\cite{bedi2025optimization}}, Health, text width=14cm, node options={align=left}]
        ]
        [{Environmental Simulation}, Health, text width=3.6cm
            [{AMIE~\cite{tu2025towards}, AgentClinic~\cite{schmidgall2024agentclinic}, Agent Hospital~\cite{li2024agent}, AI Hospital~\cite{fan2025ai}, MEDCO~\cite{wei2024medco}, MedAgentSim~\cite{almansoori2025self}}, Health, text width=14cm, node options={align=left}]
        ]
    ]
    [{Software Engineering (\S\ref{se})}, Software, text width=3.6cm
        [{Code Generation and Testing}, Software, text width=3.6cm
            [{MapCoder~\cite{islam2024mapcoder}, Almorsi et al.~\cite{almorsi2024guided}, AgentCoder~\cite{huang2023agentcoder}, CoCoST~\cite{he2024cocost}, CodeAgent~\cite{zhang2024codeagent}, Self-planning~\cite{jiang2024self}, Self-debugging~\cite{chenteaching}, Self-collaboration~\cite{dong2024self}, CodeSIM~\cite{islam2025codesim}, CollabCoder~\cite{doan2026collabcoder}, MavenCoder~\cite{mavencoder2026}, CoT~\cite{wei2022chain}, ReAct~\cite{yao2023react}, Reflexion~\cite{shinn2023reflexion}, AgentVerse~\cite{chen2024agentverse}, MetaGPT~\cite{hongmetagpt}, ChatDev~\cite{qian2024chatdev}, Chen et al. (self-generated-test debugging)~\cite{selfdebugtests2025}}, Software, text width=14cm, node options={align=left}]
        ]
        [{Program Repair}, Software, text width=3.6cm
            [{RepairAgent~\cite{bouzenia2025repairagent}, AgentFL~\cite{qin2024agentfl}, INTERVENOR~\cite{wang2024intervenor}, OrcaLoca~\cite{yuorcaloca}, Agentless~\cite{xia2024agentless}, LocAgent~\cite{chen2025locagent}, VulDebugger~\cite{liu2025agent}, MAGIS~\cite{tao2024magis}, AutoCodeRover~\cite{zhang2024autocoderover}, Rondon et al.~\cite{rondon2025evaluating}, SWE-Search~\cite{antoniades2025swesearch}, Moatless-Adapted~\cite{moatless2026tools}, SWE-agent~\cite{yang2024swe}}, Software, text width=14cm, node options={align=left}]
        ]
        [{Full-cycle Development}, Software, text width=3.6cm
            [{ChatDev~\cite{qian2024chatdev}, MetaGPT~\cite{hongmetagpt}, RepoAgent~\cite{luo2024repoagent}, DocAgent~\cite{yang2025docagent}, SWE-agent~\cite{yang2024swe}, OpenHands~\cite{wangopenhands}, MAAD~\cite{zhang2025knowledge}, SyncMind~\cite{guosyncmind}, Darwin G\"odel Machine~\cite{zhang2026dgm}, Self-collaboration~\cite{dong2024self}}, Software, text width=14cm, node options={align=left}]
        ]
    ]
    [{Social and Economic Simulation (\S\ref{sec:socioeconomic})}, Social, text width=3.6cm
        [{Social Simulation}, Social, text width=3.6cm
            [{Generative Agents~\cite{park2023generative}, MetaAgents~\cite{li2023metaagents}, GOVSIM~\cite{piatti2024cooperate}, RecAgent~\cite{wang2025user}, ElectionSim~\cite{zhang2024electionsim}, GenSim~\cite{tang2025gensim}, Sotopia-S4~\cite{zhou2025sotopia}, BASES~\cite{ren2024bases}, Mou et al.~\cite{mou2024unveiling}, BotSim~\cite{qiao2025botsim}, Y-Social~\cite{rossetti2024social}, OASIS~\cite{yang2024oasis}, SocioVerse~\cite{zhang2025socioverse}, S3~\cite{gao2023s}, GA-S3~\cite{zhang2025gas3}, MF-LLM~\cite{mi2025mfllm}}, Social, text width=14cm, node options={align=left}]
        ]
        [{Economic Perception and Simulation}, Social, text width=3.6cm
            [{FinMem~\cite{yu2024finmem}, EconAgent~\cite{li2024econagent}, FinVision~\cite{fatemi2024finvision}, FinAgent~\cite{zhang2024multimodal}, FinCon~\cite{yu2024fincon}, TradingAgents~\cite{xiao2025tradingagents}, FinRobot~\cite{yang2024finrobot}, StockAgent~\cite{zhang2024ai}, Gao et al.~\cite{gao2024simulating}, TwinMarket~\cite{yang2025twinmarket}}, Social, text width=14cm, node options={align=left}]
        ]
    ]
    [{General-Purpose Agents (\S\ref{sec:general_agents})}, General, text width=3.6cm
        [{Task-Oriented Agentic Systems}, General, text width=3.6cm
            [{WebDancer~\cite{wu2025webdancer}, WebThinker~\cite{li2025webthinker}, Workforce / OWL~\cite{hu2025owl}, Flash-Searcher~\cite{qin2026flashsearcher}, FlowSearch~\cite{hu2026flowsearch}, DeepVerifier~\cite{wan2026deepverifier}, ReThinker~\cite{zhang2026rethinker}, AlphaEvolve~\cite{novikov2025alphaevolve}, PoTRE~\cite{kankariya2026potre}, JoyAgent-JDGenie~\cite{liu2025joyagent}, MiroFlow~\cite{wang2026miroflow}, Tongyi DeepResearch~\cite{tongyi2025deepresearch}, ACON~\cite{kang2026acon}, OpenHands~\cite{wangopenhands}, Codex~\cite{openai2026codex}, Claude Code~\cite{anthropic2026claudecode}, ALE-Claw~\cite{sun2026ale}, OpenClaw~\cite{openclaw2026software}, Cursor~\cite{cursor2026agent}, AOrchestra~\cite{ruan2026aorchestra}, AutoGen~\cite{wu2024autogen}, FRIDAY / OS-Copilot~\cite{wu2024oscopilot}, Manus~\cite{qiu2025alita}, OAgents~\cite{zhu2025oagents}, OctoTools~\cite{lu2026octotools}, Search-o1~\cite{li2025searcho1}}, General, text width=14cm, node options={align=left}]
        ]
        [{Self-Improving Agentic Systems}, General, text width=3.6cm
            [{Memento~\cite{zhou2025memento}, ExpeL~\cite{zhao2024expel}, A-MEM~\cite{xu2025amem}, Mem0~\cite{chhikara2025mem0}, Agent Workflow Memory~\cite{wang2025awm}, Memento-Skills~\cite{zhou2026mementoskills}, Alita~\cite{qiu2025alita}, AutoAgent~\cite{tang2026autoagent}, Alita-G~\cite{qiu2025alitag}, Mem$^2$Evolve~\cite{cheng2026mem2evolve}, ROGA~\cite{liu2026roga}, TPGO~\cite{he2026tpgo}, EvoRoute~\cite{zhang2026evoroute}, Darwin G\"odel Machine~\cite{zhang2026dgm}, ShinkaEvolve~\cite{lange2026shinkaevolve}, Meta-Harness~\cite{lee2026metaharness}, AHE~\cite{lin2026ahe}, RHO~\cite{pan2026rho}, MCE~\cite{ye2026mce}, HarnessFix~\cite{chen2026harnessfix}, Self-Harness~\cite{zhang2026selfharness}, HarnessBank~\cite{luo2026harnessbank}, HarnessCompass~\cite{zhang2026harnesscompass}, HarnessX~\cite{chen2026harnessx}, MSCE~\cite{tang2026msce}, AEvo~\cite{zhang2026aevo}, Adaptive Auto-Harness~\cite{liu2026adaptiveharness}, ACE~\cite{zhang2026ace}, TF-GRPO~\cite{cai2025tfgrpo}, GEPA~\cite{agrawal2026gepa}, SCOPE~\cite{pei2025scope}, ReCreate~\cite{hao2026recreate}, Dynamic Cheatsheet~\cite{suzgun2025dynamic}, ReasoningBank~\cite{ouyang2026reasoningbank}, Sleep-time Compute~\cite{lin2025sleeptime}, ADAS~\cite{hu2025adas}, AFlow~\cite{zhang2025aflow}, SPO~\cite{xiang2025spo}, EverOS~\cite{hu2026evermemos}, EvoSkill~\cite{alzubi2026evoskill}, OpenSpace~\cite{hkuds2026openspace}, MemSkill~\cite{zhang2026memskill}, SkillFlow-Evolve~\cite{zhang2026skillflow}}, General, text width=14cm, node options={align=left}]
        ]
    ]
    [{Others (\S\ref{sec:others})}, Others, text width=3.6cm
        [{Voyager~\cite{wang23voyager}, GATE~\cite{luo2026gate}, Chain of Summarization~\cite{ma2024large}, concept-guided chess commentary~\cite{kim2025bridging}, Fan et al. (game-theoretic agents)~\cite{fan2024can}, SeeAct~\cite{zheng2024gpt}, GPT-4V in Wonderland~\cite{yan2023gpt}, AppAgent~\cite{zhang2025appagent}, MobileGPT~\cite{lee2024mobilegpt}, Mobile-Agent-v2~\cite{wang2024mobile}, Mobile-Agent-E~\cite{wang2025mobile}, Agent S2~\cite{agashe2025agents2}, Agent S3~\cite{gonzalez2025agents3}, GUI-assisted Codex / OpenClaw / ALE-Claw~\cite{sun2026ale}}, Others, text width=18.16cm, node options={align=left}]
    ]
]
\end{forest}
}}
\usebox{\scenarioinventorybox}%
\caption{Full inventory of surveyed application systems and comparison methods.}
\Description{A complete one-page application tree. Scientific research, healthcare, software engineering, social and economic simulation, and general-purpose agents connect through their subcategories to all listed works. Others lists additional application systems.}
\label{fig:scenarios}
\end{figure*}

%% file: fig/table_biochem.tex
\begin{table*}[!t]
\centering
\caption{Evidence access in biochemistry and material science. The middle column summarizes illustrative routes into reasoning; the final column indexes related work. Sources can overlap within a system. Table~\ref{tab:bio_mat} separately lists evaluation resources.}
\label{tab:biochem}
\small\setlength{\tabcolsep}{4pt}\renewcommand{\arraystretch}{1.12}
\begin{tabularx}{\textwidth}{@{}p{0.18\textwidth}>{\RaggedRight\arraybackslash}p{0.37\textwidth}>{\RaggedRight\arraybackslash}X@{}}
\toprule
\rowcolor{lightblue}
\textbf{Evidence source} & \textbf{How evidence enters reasoning} & \textbf{Works} \\
\midrule
Literature and reference knowledge & Retrieved passages supply reference knowledge to the reasoning context. & MOOSE-Chem~\cite{yangmoose}; Stewart et al.~\cite{stewart2025molecular}; BioAgents~\cite{mehandru2025bioagents}; CRISPR-GPT~\cite{qu2025crispr}; BioResearcher~\cite{luo2025intention}; DrugAgent~\cite{inoue2025drugagent}. \\
\midrule
Scientific records and knowledge graphs & Queried records and graph relations are combined as task-specific evidence. & ChatDrug~\cite{liu2024conversational}, CLADD~\cite{lee2025rag}, DrugAgent~\cite{inoue2025drugagent}; Chemist-X~\cite{chen2023chemist}; ChatMOF~\cite{kang2024chatmof}, LLaMP~\cite{chiang2025llamp}. \\
\midrule
Computational and simulation outputs & Tool outputs provide observations for revising intermediate reasoning and actions. & LIDDIA~\cite{averly2025liddia}, DrugAgent~\cite{inoue2025drugagent}; ChatChemTS~\cite{ishida2025large}; ChemCrow~\cite{bran2023chemcrow}, CACTUS~\cite{mcnaughton2024cactus}; El Agente Q~\cite{zou2025agente}; AtomAgents~\cite{ghafarollahi2025automating}; ProtAgents~\cite{ghafarollahi2024protagents}, TourSynbio-Search~\cite{liu2024toursynbio}. \\
\midrule
Execution and workflow feedback & Intermediate results and execution checks guide revision and stage handoffs. & AutoBA~\cite{zhou2024ai}; BioMaster~\cite{su2025biomaster}; ChemHTS~\cite{li2025chemhts}; PharmAgents~\cite{gao2025pharmagents}, Ruan et al.~\cite{ruan2024automatic}. \\
\midrule
Previous experimental observations & Retained observations inform reasoning in subsequent research rounds. & BioDiscoveryAgent~\cite{roohanibiodiscoveryagent}. \\
\bottomrule
\end{tabularx}
\end{table*}

%% file: fig/table_codegen.tex
\begin{table*}[!t]
\centering
\caption{Code-generation performance (Pass@1, \%). Direct baselines appear first, followed by methods grouped by task model. Source identifies the paper or model report providing the data and experimental design; ``--'' denotes an unreported result.}
\label{tab:codegen}
\begingroup
\setlength{\tabcolsep}{1pt}
\renewcommand{\arraystretch}{1.0}
\fontsize{6.4}{7.2}\selectfont
\begin{tabularx}{\linewidth}{@{}p{78pt}r*{9}{>{\centering\arraybackslash}X}@{}}
\toprule
\rowcolor{lightblue}
\textbf{Method / model} & \textbf{Source} & \shortstack{HE\\\cite{chen2021evaluating}} & \shortstack{HE-ET\\\cite{dong2025codescore}} & \shortstack{HE+\\\cite{liu2023your}} & \shortstack{MBPP\\\cite{austin2021program}} & \shortstack{MBPP-ET\\\cite{dong2025codescore}} & \shortstack{MBPP+\\\cite{liu2023your}} & \shortstack{DS-1000\\\cite{lai2023ds}} & \shortstack{LCB S112\\\cite{jain2025livecodebench}} & \shortstack{LCB V1\\\cite{jain2025livecodebench}} \\
\midrule
\rowcolor{black!6}
\multicolumn{11}{@{}l@{}}{\textbf{Direct baselines}} \\
GPT-3.5 & \cite{huang2023agentcoder} & 57.3 & 42.7 & -- & 52.2 & 36.8 & -- & -- & -- & -- \\
GPT-3.5 & \cite{islam2024mapcoder} & 48.1 & 37.2 & 66.5 & 49.8 & 37.7 & -- & -- & -- & -- \\
GPT-4 & \cite{huang2023agentcoder} & 67.6 & 50.6 & -- & 68.3 & 52.2 & -- & -- & -- & -- \\
GPT-4 & \cite{islam2024mapcoder} & 80.1 & 73.8 & 81.7 & 81.1 & 54.7 & -- & -- & -- & -- \\
GPT-4o & \cite{anthropic2024claude35modelcard} & 90.2 & -- & -- & -- & -- & -- & -- & -- & -- \\
Claude 3.5 Sonnet & \cite{anthropic2024claude35modelcard} & 92.0 & -- & -- & -- & -- & -- & -- & -- & -- \\
GPT-4.1-nano & \cite{mavencoder2026}\textsuperscript{$\ddagger$} & 89.0 & -- & 82.9 & 86.8 & -- & 72.8 & -- & -- & 48.3 \\
Qwen3-Coder+ & \cite{mavencoder2026}\textsuperscript{$\ddagger$} & 97.0 & -- & 93.3 & 94.4 & -- & 80.2 & -- & -- & 76.8 \\
\addlinespace[1.5pt]
\rowcolor{black!6}
\multicolumn{11}{@{}l@{}}{\textbf{GPT-3.5}} \\
CoT~\cite{wei2022chain} & \cite{huang2023agentcoder} & 44.6 & 37.2 & -- & 46.1 & 34.8 & -- & -- & -- & -- \\
CoT~\cite{wei2022chain} & \cite{islam2024mapcoder} & 68.9 & 55.5 & 65.2 & 54.5 & 39.6 & -- & -- & -- & -- \\
ReAct~\cite{yao2023react} & \cite{huang2023agentcoder} & 56.9 & 49.4 & -- & 67.0 & 45.9 & -- & -- & -- & -- \\
Reflexion~\cite{shinn2023reflexion} & \cite{huang2023agentcoder} & 68.1 & 50.6 & -- & 70.0 & 47.5 & -- & -- & -- & -- \\
Reflexion~\cite{shinn2023reflexion} & \cite{islam2024mapcoder} & 67.1 & 49.4 & 62.2 & 73.0 & 47.4 & -- & -- & -- & -- \\
Self-planning~\cite{jiang2024self} & \cite{huang2023agentcoder} & 65.2 & 48.8 & -- & 58.6 & 41.5 & -- & -- & -- & -- \\
Self-debugging~\cite{chenteaching} & \cite{huang2023agentcoder} & 61.6 & 45.8 & -- & 60.1 & 52.3 & -- & -- & -- & -- \\
Self-collaboration~\cite{dong2024self} & \cite{huang2023agentcoder} & 74.4 & 56.1 & -- & 68.2 & 49.5 & -- & -- & -- & -- \\
AgentCoder~\cite{huang2023agentcoder} & \cite{huang2023agentcoder} & 79.9 & 77.4 & -- & 89.9 & 89.1 & -- & -- & -- & -- \\
INTERVENOR~\cite{wang2024intervenor} & \cite{wang2024intervenor} & 75.6 & -- & -- & 69.8 & -- & -- & 39.7 & -- & -- \\
MapCoder~\cite{islam2024mapcoder} & \cite{islam2024mapcoder} & 80.5 & 70.1 & 71.3 & 78.3 & 54.4 & -- & -- & -- & -- \\
\addlinespace[1.5pt]
\rowcolor{black!6}
\multicolumn{11}{@{}l@{}}{\textbf{GPT-4}} \\
CoT~\cite{wei2022chain} & \cite{islam2024mapcoder} & 89.0 & 61.6 & -- & 82.4 & 56.2 & -- & -- & -- & -- \\
Reflexion~\cite{shinn2023reflexion} & \cite{islam2024mapcoder} & 91.0 & 78.7 & 81.7 & 78.3 & 51.9 & -- & -- & -- & -- \\
Self-debugging~\cite{chenteaching} & \cite{huang2023agentcoder} & -- & -- & -- & 80.6 & -- & -- & -- & -- & -- \\
Self-collaboration~\cite{dong2024self} & \cite{huang2023agentcoder} & 90.2 & 70.7 & -- & 78.9 & 62.1 & -- & -- & -- & -- \\
AgentCoder~\cite{huang2023agentcoder} & \cite{huang2023agentcoder} & 96.3 & 86.0 & -- & 91.8 & 91.8 & -- & -- & -- & -- \\
AgentVerse~\cite{chen2024agentverse} & \cite{huang2023agentcoder} & 89.0 & -- & -- & 73.5 & -- & -- & -- & -- & -- \\
ChatDev~\cite{qian2024chatdev} & \cite{huang2023agentcoder} & 84.1 & -- & -- & 79.8 & -- & -- & -- & -- & -- \\
MetaGPT~\cite{hongmetagpt} & \cite{huang2023agentcoder} & 85.9 & -- & -- & 87.7 & -- & -- & -- & -- & -- \\
CoCoST~\cite{he2024cocost} & \cite{he2024cocost} & -- & -- & -- & -- & -- & -- & 68.00 & -- & -- \\
MapCoder~\cite{islam2024mapcoder} & \cite{islam2024mapcoder} & 93.9 & 82.9 & 83.5 & 83.1 & 57.7 & -- & -- & -- & -- \\
\addlinespace[1.5pt]
\rowcolor{black!6}
\multicolumn{11}{@{}l@{}}{\textbf{GPT-4o-mini}} \\
MapCoder~\cite{islam2024mapcoder} & \cite{doan2026collabcoder} & 90.24 & 79.88 & -- & 84.13 & 56.93 & -- & -- & 34.82 & -- \\
CodeSIM~\cite{islam2025codesim} & \cite{doan2026collabcoder} & 94.51 & 81.70 & -- & 89.92 & 59.95 & -- & -- & 36.61 & -- \\
CollabCoder~\cite{doan2026collabcoder} & \cite{doan2026collabcoder} & 96.34 & 84.76 & -- & 91.69 & 60.20 & -- & -- & 41.96 & -- \\
\addlinespace[1.5pt]
\rowcolor{black!6}
\multicolumn{11}{@{}l@{}}{\textbf{Qwen2.5-Coder-32B}} \\
MapCoder~\cite{islam2024mapcoder} & \cite{doan2026collabcoder} & 90.24 & 79.00 & -- & 86.80 & 59.95 & -- & -- & -- & -- \\
CodeSIM~\cite{islam2025codesim} & \cite{doan2026collabcoder} & 93.29 & 81.70 & -- & 87.20 & 58.70 & -- & -- & -- & -- \\
CollabCoder~\cite{doan2026collabcoder} & \cite{doan2026collabcoder} & 95.73 & 84.15 & -- & 90.17 & 59.95 & -- & -- & -- & -- \\
\addlinespace[1.5pt]
\rowcolor{black!6}
\multicolumn{11}{@{}l@{}}{\textbf{GPT-4.1-nano}} \\
MapCoder~\cite{islam2024mapcoder} & \cite{mavencoder2026}\textsuperscript{$\ddagger$} & 87.8 & -- & 81.1 & 93.1 & -- & 75.4 & -- & -- & 35.5 \\
MavenCoder~\cite{mavencoder2026} & \cite{mavencoder2026}\textsuperscript{$\ddagger$} & 97.6 & -- & 90.2 & 97.1 & -- & 81.0 & -- & -- & 70.0 \\
\addlinespace[1.5pt]
\rowcolor{black!6}
\multicolumn{11}{@{}l@{}}{\textbf{Qwen3-Coder+}} \\
MapCoder~\cite{islam2024mapcoder} & \cite{mavencoder2026}\textsuperscript{$\ddagger$} & 97.6 & -- & 92.7 & 98.1 & -- & 81.0 & -- & -- & -- \\
MavenCoder~\cite{mavencoder2026} & \cite{mavencoder2026}\textsuperscript{$\ddagger$} & 98.8 & -- & 96.3 & 98.9 & -- & 81.7 & -- & -- & 87.5 \\
\addlinespace[1.5pt]
\rowcolor{black!6}
\multicolumn{11}{@{}l@{}}{\textbf{GPT-5.2}} \\
MapCoder~\cite{islam2024mapcoder} & \cite{doan2026collabcoder} & -- & -- & -- & 93.94 & -- & -- & -- & 63.39 & -- \\
CodeSIM~\cite{islam2025codesim} & \cite{doan2026collabcoder} & -- & -- & -- & 94.74 & -- & -- & -- & 64.29 & -- \\
CollabCoder~\cite{doan2026collabcoder} & \cite{doan2026collabcoder} & -- & -- & -- & 95.21 & -- & -- & -- & 65.18 & -- \\
\addlinespace[1.5pt]
\rowcolor{black!6}
\multicolumn{11}{@{}l@{}}{\textbf{Qwen3-Coder-Next}} \\
MapCoder~\cite{islam2024mapcoder} & \cite{doan2026collabcoder} & -- & -- & -- & 89.42 & -- & -- & -- & 50.89 & -- \\
CodeSIM~\cite{islam2025codesim} & \cite{doan2026collabcoder} & -- & -- & -- & 91.18 & -- & -- & -- & 52.68 & -- \\
CollabCoder~\cite{doan2026collabcoder} & \cite{doan2026collabcoder} & -- & -- & -- & 92.69 & -- & -- & -- & 55.38 & -- \\
\bottomrule
\end{tabularx}
\par\vspace{3pt}
\begin{minipage}{\linewidth}\scriptsize
HE = HumanEval; ET = extended tests; + = EvalPlus; LCB = LiveCodeBench. S112 and V1 retain their source-defined subset/release. \textsuperscript{$\ddagger$}Best of two runs; Direct uses greedy decoding in these rows.
\end{minipage}
\endgroup
\end{table*}

%% file: fig/table_repair.tex
\Needspace{355pt}
\begin{wraptable}{R}{0.5\surveycolumnwidth}
\centering
\captionsetup{skip=3pt}
\caption{Software-maintenance results, grouped by task model. Source identifies the reporting study. Defects4J reports successful/evaluated bugs; SWE-bench reports resolved issues (\%). Settings differ across reports.}
\label{tab:coderep}
\begingroup
\fontsize{7}{8}\selectfont
\setlength{\tabcolsep}{1.5pt}\renewcommand{\arraystretch}{1.02}
\begin{tabularx}{\linewidth}{@{}>{\RaggedRight\arraybackslash}Xcrr@{}}
\toprule
\rowcolor{lightblue}
\multicolumn{4}{c}{\textbf{Defects4J v1.2}~\cite{just2014defects4j}} \\
Method & Source & Loc. & Repair \\
\midrule
\multicolumn{4}{l}{\textit{GPT-3.5}} \\
AgentFL & \cite{qin2024agentfl} & 157/395 & -- \\
ChatGPT + Ochiai & \cite{qin2024agentfl} & 121/395 & -- \\
RepairAgent & \cite{bouzenia2025repairagent} & -- & 74/395 \\
\midrule
\rowcolor{lightblue}
\multicolumn{4}{c}{\textbf{SWE-bench}~\cite{jimenezswe}} \\
Method & Source & Full & Lite \\
\midrule
Direct (GPT-3.5) & \cite{tao2024magis} & 0.84\textsuperscript{*} & -- \\
Direct (GPT-4) & \cite{tao2024magis} & 1.74\textsuperscript{*} & -- \\
\addlinespace[2pt]
\multicolumn{4}{l}{\textit{GPT-4}} \\
AutoCodeRover & \cite{zhang2024autocoderover,xia2024agentless} & 12.42 & 19.00 \\
MAGIS & \cite{tao2024magis} & 13.94\textsuperscript{*} & 25.33 \\
\addlinespace[2pt]
\multicolumn{4}{l}{\textit{GPT-4 Turbo}} \\
SWE-agent & \cite{yang2024swe} & 12.47 & 18.00 \\
\addlinespace[2pt]
\multicolumn{4}{l}{\textit{GPT-4o}} \\
Agentless & \cite{xia2024agentless} & -- & 32.00 \\
AutoCodeRover-v2 & \cite{xia2024agentless} & -- & 30.67 \\
Moatless-Adapted & \cite{antoniades2025swesearch} & -- & 25.7 \\
SWE-Search & \cite{antoniades2025swesearch} & -- & 31.0 \\
SWE-agent & \cite{xia2024agentless} & -- & 18.33 \\
\addlinespace[2pt]
\multicolumn{4}{l}{\textit{Claude 3.5 Sonnet}} \\
OrcaLoca + Agentless-1.5 & \cite{yuorcaloca} & -- & 41.00 \\
\addlinespace[2pt]
\multicolumn{4}{l}{\textit{GPT-4o-mini}} \\
Moatless-Adapted & \cite{antoniades2025swesearch} & -- & 13.0 \\
SWE-Search & \cite{antoniades2025swesearch} & -- & 17.0 \\
\addlinespace[2pt]
\multicolumn{4}{l}{\textit{DeepSeek-V2.5}} \\
Moatless-Adapted & \cite{antoniades2025swesearch} & -- & 16.3 \\
SWE-Search & \cite{antoniades2025swesearch} & -- & 21.0 \\
\addlinespace[2pt]
\multicolumn{4}{l}{\textit{Qwen2.5-72B-Instruct}} \\
Moatless-Adapted & \cite{antoniades2025swesearch} & -- & 18.0 \\
SWE-Search & \cite{antoniades2025swesearch} & -- & 24.7 \\
\bottomrule
\end{tabularx}
\par\vspace{2pt}
\RaggedRight\fontsize{6.8}{7.8}\selectfont
Loc.: top-1 localization. --: unreported.\par
\textsuperscript{*}Source-defined subset with oracle file context~\cite{tao2024magis}. Moatless-Adapted uses Moatless Tools~\cite{moatless2026tools}.
\endgroup
\end{wraptable}

%% file: fig/table_general_agents.tex
\begin{table*}[!t]
\centering
\caption{General-purpose agent systems, grouped by model in release order. Composite configurations follow the primary model; undated or undisclosed models appear last. Scores are percentages; GAIA and HLE prioritize full-set over text-only results. GAIA reports level 1/level 2/level 3/avg; ALE reports Near-Term/Full-Spectrum/Last-Exam/overall full-pass rates. Overall scores are source-reported; ``-'' denotes an unreported component. Source identifies the paper or official resource reporting the data and experimental design.}
\label{tab:general_agents}
\begingroup
\setlength{\tabcolsep}{3pt}
\renewcommand{\arraystretch}{1.0}
\fontsize{9}{10.2}\selectfont
\resizebox{\textwidth}{!}{%
\begin{tabular}{@{}lllrrrrrrr@{}}
\toprule
\rowcolor{lightblue}
\textbf{Method} & \textbf{Model} & \textbf{Source} & \shortstack{TB 2.0\\\cite{merrill2026terminalbench}} & \shortstack{TB 4.0\\\cite{terminalbench2026v4}} & \multicolumn{1}{c}{\shortstack{GAIA\\\cite{mialon2024gaia}}} & \shortstack{BrowseComp\\\cite{wei2025browsecomp}} & \shortstack{HLE\\\cite{cais2026hle}} & \shortstack{xBench-DS\\\cite{chen2025xbench}} & \shortstack{ALE\\\cite{sun2026ale}} \\
\midrule
\rowcolor{black!3}
FRIDAY / OS-Copilot~\cite{wu2024oscopilot} & GPT-4-turbo-1106 & \cite{wu2024oscopilot} & -- & -- & 40.86/20.13/6.12/-\textsuperscript{$\dagger$} & -- & -- & -- & -- \\
\addlinespace[1pt]
AutoGen~\cite{wu2024autogen} & GPT-4o-2024-08-06 & \cite{lu2026octotools} & -- & -- & -/-/-/6.3\textsuperscript{\S} & -- & -- & -- & -- \\
OctoTools~\cite{lu2026octotools} & GPT-4o-2024-08-06 & \cite{lu2026octotools} & -- & -- & -/-/-/18.4\textsuperscript{\S} & -- & -- & -- & -- \\
\addlinespace[1pt]
\rowcolor{black!3}
AutoAgent~\cite{tang2026autoagent} & 3.5 Sonnet & \cite{tang2026autoagent} & -- & -- & 71.70/53.49/26.92/55.15 & -- & -- & -- & -- \\
\addlinespace[1pt]
DeepVerifier~\cite{wan2026deepverifier} & 3.7 Sonnet & \cite{wan2026deepverifier} & -- & -- & -/-/-/58.93 & 9.0 & -- & 44.0 & -- \\
Workforce / OWL~\cite{hu2025owl} & 3.7 Sonnet & \cite{hu2025owl} & -- & -- & 84.91/68.60/42.31/69.70 & -- & -- & -- & -- \\
\addlinespace[1pt]
\rowcolor{black!3}
Alita~\cite{qiu2025alita} & 3.7 Sonnet + GPT-4o & \cite{qiu2025alita} & -- & -- & 81.13/75.58/46.15/72.73 & -- & -- & -- & -- \\
\addlinespace[1pt]
OAgents~\cite{zhu2025oagents} & 3.7 Sonnet + helpers & \cite{zhu2025oagents} & -- & -- & 77.36/66.28/46.15/66.67 & 22.22\textsuperscript{*} & 14.86\textsuperscript{*} & -- & -- \\
\addlinespace[1pt]
\rowcolor{black!3}
Search-o1~\cite{li2025searcho1} & QwQ-32B & \cite{li2025webthinker} & -- & -- & 53.8/34.6/16.7/39.8\textsuperscript{\S} & -- & 10.8\textsuperscript{*} & -- & -- \\
\rowcolor{black!3}
WebThinker-Base~\cite{li2025webthinker} & QwQ-32B & \cite{li2025webthinker} & -- & -- & 53.8/44.2/16.7/44.7\textsuperscript{\S} & -- & 13.0\textsuperscript{*} & -- & -- \\
\addlinespace[1pt]
OAgents~\cite{zhu2025oagents} & GPT-4.1 & \cite{zhu2025oagents} & -- & -- & -- & 22.22\textsuperscript{*} & 15.43\textsuperscript{*} & -- & -- \\
\addlinespace[1pt]
\rowcolor{black!3}
FlowSearch~\cite{hu2026flowsearch} & o4-mini & \cite{hu2026flowsearch} & -- & -- & 92.45/82.56/61.54/82.42 & -- & 34.52 & -- & -- \\
\addlinespace[1pt]
FlowSearch~\cite{hu2026flowsearch} & Qwen3-235B & \cite{hu2026flowsearch} & -- & -- & 69.81/60.47/30.77/58.79 & -- & 14.84 & -- & -- \\
\addlinespace[1pt]
\rowcolor{black!3}
Alita~\cite{qiu2025alita} & Sonnet 4 + GPT-4o & \cite{qiu2025alita} & -- & -- & 77.36/76.74/65.38/75.15 & -- & -- & -- & -- \\
\addlinespace[1pt]
Alita~\cite{qiu2025alita} & Sonnet 4 + GPT-4.1 & \cite{qiu2025alitag} & -- & -- & 77.36/76.74/65.38/75.15 & -- & 24\textsuperscript{*} & -- & -- \\
Alita-G~\cite{qiu2025alitag} & Sonnet 4 + GPT-4.1 & \cite{qiu2025alitag} & -- & -- & 86.80/83.72/73.08/83.03 & -- & 33\textsuperscript{*} & -- & -- \\
\addlinespace[1pt]
\rowcolor{black!3}
JoyAgent-JDGenie~\cite{liu2025joyagent} & Sonnet 4 + o4-mini & \cite{liu2025joyagent} & -- & -- & 86.8/77.9/42.3/75.2 & -- & -- & -- & -- \\
\addlinespace[1pt]
MiroFlow~\cite{wang2026miroflow} & GPT-5 & \cite{wang2026miroflow} & -- & -- & -/-/-/71.9 & 63.4 & 35.5 & 73.5 & -- \\
MiroFlow (heavy)~\cite{wang2026miroflow} & GPT-5 & \cite{wang2026miroflow} & -- & -- & -/-/-/81.1\textsuperscript{$\dagger$} & -- & -- & -- & -- \\
\addlinespace[1pt]
\rowcolor{black!3}
Flash-Searcher~\cite{qin2026flashsearcher} & GPT-5 (m) & \cite{qin2026flashsearcher} & -- & -- & -/-/-/83.0 & 67.7 & 44.0\textsuperscript{*} & 83 & -- \\
\addlinespace[1pt]
Tongyi DeepResearch~\cite{tongyi2025deepresearch} & Tongyi-DR-30B-A3B & \cite{tongyi2025deepresearch} & -- & -- & -/-/-/70.9\textsuperscript{\S} & 43.4 & 32.9\textsuperscript{\S} & 75.0 & -- \\
\addlinespace[1pt]
\rowcolor{black!3}
Claude Code~\cite{anthropic2026claudecode} & Haiku 4.5 & \cite{lee2026metaharness} & 27.5 & -- & -- & -- & -- & -- & -- \\
\addlinespace[1pt]
Model with tools~\cite{zhang2026rethinker} & Gemini 3 Pro & \cite{zhang2026rethinker} & -- & -- & -/-/-/79.0\textsuperscript{\S} & -- & 38.3\textsuperscript{\S} & 87.0 & -- \\
ReThinker~\cite{zhang2026rethinker} & Gemini 3 Pro & \cite{zhang2026rethinker} & -- & -- & 82.05/80.77/83.33/81.6\textsuperscript{\S} & -- & 52.2\textsuperscript{\S} & 90.0 & -- \\
\addlinespace[1pt]
\rowcolor{black!3}
OpenHands~\cite{wangopenhands} & DeepSeek V3.2 & \cite{ruan2026aorchestra} & 21.43\textsuperscript{*} & -- & -/-/-/63.64 & -- & -- & -- & -- \\
\addlinespace[1pt]
AOrchestra~\cite{ruan2026aorchestra} & Gemini 3 Flash & \cite{ruan2026aorchestra} & 52.86\textsuperscript{*} & -- & -/-/-/80.00 & -- & -- & -- & -- \\
OpenHands~\cite{wangopenhands} & Gemini 3 Flash & \cite{ruan2026aorchestra} & 31.43\textsuperscript{*} & -- & -/-/-/66.06 & -- & -- & -- & -- \\
\addlinespace[1pt]
\rowcolor{black!3}
PoTRE~\cite{kankariya2026potre} & Gemini 3 Flash Pr. & \cite{kankariya2026potre} & -- & -- & -- & -- & 53.48 & -- & -- \\
\addlinespace[1pt]
Claude Code~\cite{anthropic2026claudecode} & Opus 4.6 & \cite{lee2026metaharness} & 58.0 & -- & -- & -- & -- & -- & -- \\
\addlinespace[1pt]
\rowcolor{black!3}
PoTRE~\cite{kankariya2026potre} & Gemini 3.1 Pro Pr. & \cite{kankariya2026potre} & -- & -- & -- & -- & 58.40 & -- & -- \\
\addlinespace[1pt]
Codex~\cite{openai2026codex} & GPT-5.4 (h) & \cite{lin2026ahe} & 71.9 & -- & -- & -- & -- & -- & -- \\
\addlinespace[1pt]
\rowcolor{black!3}
ALE-Claw~\cite{sun2026ale} & GPT-5.5 & \cite{sun2026ale} & -- & -- & -- & -- & -- & -- & 32.8/23.6/2.6/23.0 \\
\rowcolor{black!3}
Codex~\cite{openai2026codex} & GPT-5.5 & \cite{sun2026ale} & -- & -- & -- & -- & -- & -- & 38.1/22.7/0.0/24.0 \\
\rowcolor{black!3}
Cursor~\cite{cursor2026agent} & GPT-5.5 & \cite{sun2026ale} & -- & -- & -- & -- & -- & -- & 32.1/20.0/2.6/20.7 \\
\rowcolor{black!3}
OpenClaw~\cite{openclaw2026software} & GPT-5.5 & \cite{sun2026ale} & -- & -- & -- & -- & -- & -- & 35.8/18.2/0.0/21.1 \\
\addlinespace[1pt]
Codex~\cite{openai2026codex} & GPT-5.6 Sol (max) & \cite{terminalbench2026v4} & -- & 37.27 & -- & -- & -- & -- & -- \\
\addlinespace[1pt]
\rowcolor{black!3}
Claude Code~\cite{anthropic2026claudecode} & Opus 5 (max) & \cite{terminalbench2026v4} & -- & 51.82 & -- & -- & -- & -- & -- \\
\addlinespace[1pt]
Claude Code~\cite{anthropic2026claudecode} & Fable 5.1 (max) & \cite{terminalbench2026v4} & -- & 57.88 & -- & -- & -- & -- & -- \\
\addlinespace[1pt]
\rowcolor{black!3}
Codex~\cite{openai2026codex} & GPT-6 Astra (max) & \cite{terminalbench2026v4} & -- & 58.18 & -- & -- & -- & -- & -- \\
\addlinespace[1pt]
ReThinker~\cite{zhang2026rethinker} & OpenPangu-72B & \cite{zhang2026rethinker} & -- & -- & 79.49/75.00/41.67/72.8\textsuperscript{\S} & -- & 33.1\textsuperscript{\S} & 78.0 & -- \\
\addlinespace[1pt]
\rowcolor{black!3}
Manus~\cite{qiu2025alita} & -- & \cite{qiu2025alita} & -- & -- & 86.5/70.1/57.7/73.3 & -- & -- & -- & -- \\
\bottomrule
\end{tabular}%
}
\par\vspace{3pt}
\begin{minipage}{\textwidth}\footnotesize
\textit{Notation.} ``--'': unreported; h/m/max: model effort. \textsuperscript{\S}Text-only; \textsuperscript{$\dagger$}GAIA test; \textsuperscript{*}other source-defined benchmark subsets. Each marked score follows its cited Source. Terminal-Bench leaderboard scores retain the September 2026 snapshots in Source.
\end{minipage}
\endgroup
\end{table*}

%% file: fig/table_self_evolving.tex
\begin{table*}[!htb]
\centering
\caption{Self-evolving systems and harness optimizers (scores in \%). Source identifies the reporting study. $x$--$y$: baseline/control and method scores; single values: final scores; ``--'': unreported; $\dagger$: independent reference comparison. *: TB = Terminal-Bench 2.0; $\ddagger$: TB version unspecified by the source. SWE-V/Pro: SWE-bench Verified/Pro. \textit{Evaluation split:} O: optimization/search tasks; G: held-out promotion-gate tasks, hidden from the proposer; T: test tasks excluded from search and selection; M: task-weighted total over search and held-out tasks; X: frozen cross-model transfer; U: split unspecified; ``cell'': per-score O/T labels.}
\label{tab:self_evolving}
\textbf{(a) Execution, interactive tasks and abstract reasoning}\par\vspace{2pt}
\begingroup\fontsize{6.5}{8}\selectfont\setlength{\tabcolsep}{2pt}\renewcommand{\arraystretch}{1.0}
\begin{tabularx}{\textwidth}{@{}llrc*{6}{>{\centering\arraybackslash}X}@{}}
\toprule\rowcolor{lightblue}
\textbf{Method} & \textbf{Task model} & \textbf{Source} & \textbf{Split} & \shortstack{TB\textsuperscript{*}\\\cite{merrill2026terminalbench}} & \shortstack{SWE-V\\\cite{openai2024sweverified}} & \shortstack{SWE-Pro\\\cite{deng2025swebenchpro}} & \shortstack{Gaia2-mini\\\cite{froger2026gaia2}} & \shortstack{AppWorld\\\cite{trivedi2024appworld}} & \shortstack{ARC-AGI-2\\\cite{arcprize2025arcagi2}} \\
\midrule
GEPA~\cite{agrawal2026gepa} & GPT-5 mini & \cite{chen2026harnessfix} & T & 17.6--19.6 & 45.3--46.7 & -- & -- & 36.7--37.4 & -- \\
HarnessFix~\cite{chen2026harnessfix} & GPT-5 mini & \cite{chen2026harnessfix} & T & 17.6--26.5 & 45.3--57.3 & -- & -- & 36.7--43.0 & -- \\
Meta-Harness~\cite{lee2026metaharness} & GPT-5 mini & \cite{chen2026harnessfix} & T & 17.6--23.5 & 45.3--54.7 & -- & -- & 36.7--40.4 & -- \\
ReCreate~\cite{hao2026recreate} & GPT-5 mini & \cite{chen2026harnessfix} & T & 17.6--21.6 & 45.3--51.7 & -- & -- & 36.7--39.3 & -- \\
SCOPE~\cite{pei2025scope} & GPT-5 mini & \cite{chen2026harnessfix} & T & 17.6--20.6 & 45.3--48.3 & -- & -- & 36.7--38.9 & -- \\
\addlinespace[1pt]
HarnessFix~\cite{chen2026harnessfix} & Sonnet 4.5 & \cite{chen2026harnessfix} & T & 34.3--47.1 & 61.3--71.7 & -- & -- & 61.1--70.4 & -- \\
\addlinespace[1pt]
Meta-Harness~\cite{lee2026metaharness} & Haiku 4.5 & \cite{lee2026metaharness} & O & 37.6 & -- & -- & -- & -- & -- \\
\addlinespace[1pt]
HarnessFix~\cite{chen2026harnessfix} & Gemini 3 Pro & \cite{chen2026harnessfix} & T & 46.1--55.9 & 60.3--68.7 & -- & -- & 50.7--59.6 & -- \\
\addlinespace[1pt]
HarnessFix~\cite{chen2026harnessfix} & DeepSeek V3.2 & \cite{chen2026harnessfix} & T & 31.4--39.2 & 58.7--69.3 & -- & -- & 56.3--62.2 & -- \\
\addlinespace[1pt]
ADAS~\cite{hu2025adas} & Gemini 3 Flash & \cite{zhang2026aevo} & O & 28.6--38.6\textsuperscript{$\ddagger\dagger$} & -- & -- & -- & -- & 21.8--36.0\textsuperscript{$\dagger$} \\
AEvo (procedure)~\cite{zhang2026aevo} & Gemini 3 Flash & \cite{zhang2026aevo} & O & 28.6--53.8\textsuperscript{$\ddagger\dagger$} & -- & -- & -- & -- & 21.8--47.0\textsuperscript{$\dagger$} \\
AFlow~\cite{zhang2025aflow} & Gemini 3 Flash & \cite{zhang2026aevo} & O & 28.6--44.3\textsuperscript{$\ddagger\dagger$} & -- & -- & -- & -- & 21.8--31.8\textsuperscript{$\dagger$} \\
DGM~\cite{zhang2026dgm} & Gemini 3 Flash & \cite{zhang2026aevo} & O & 28.6--44.3\textsuperscript{$\ddagger\dagger$} & -- & -- & -- & -- & 21.8--29.8\textsuperscript{$\dagger$} \\
GEPA~\cite{agrawal2026gepa} & Gemini 3 Flash & \cite{zhang2026aevo} & O & 28.6--41.4\textsuperscript{$\ddagger\dagger$} & -- & -- & -- & -- & 21.8--22.5\textsuperscript{$\dagger$} \\
SPO~\cite{xiang2025spo} & Gemini 3 Flash & \cite{zhang2026aevo} & O & 28.6--42.9\textsuperscript{$\ddagger\dagger$} & -- & -- & -- & -- & 21.8--25.0\textsuperscript{$\dagger$} \\
\addlinespace[1pt]
Meta-Harness~\cite{lee2026metaharness} & Opus 4.6 & \cite{lee2026metaharness} & O & 76.4 & -- & -- & -- & -- & -- \\
\addlinespace[1pt]
Self-Harness~\cite{zhang2026selfharness} & GLM-5 & \cite{zhang2026selfharness} & G & 42.9--57.1 & 48.5--50.0 & -- & -- & 41.1--77.8 & -- \\
\addlinespace[1pt]
Self-Harness~\cite{zhang2026selfharness} & MiniMax M2.5 & \cite{zhang2026selfharness} & G & 40.5--61.9 & 34.8--40.9 & -- & -- & 45.6--55.0 & -- \\
\addlinespace[1pt]
HarnessFix~\cite{chen2026harnessfix} & Qwen3.5 Plus & \cite{chen2026harnessfix} & T & 42.2--52.9 & 62.7--72.3 & -- & -- & 53.3--61.9 & -- \\
\addlinespace[1pt]
HarnessCompass (transfer)~\cite{zhang2026harnesscompass} & Sonnet 4.6 & \cite{zhang2026harnesscompass} & M,X & -- & 70.0--73.8 & -- & -- & -- & -- \\
HarnessX~\cite{chen2026harnessx} & Sonnet 4.6 & \cite{chen2026harnessx} & O & -- & 76.4--87.3 & -- & -- & -- & -- \\
\addlinespace[1pt]
Self-Harness~\cite{zhang2026selfharness} & Qwen3.5-35B-A3B & \cite{zhang2026selfharness} & G & 23.8--38.1 & 18.2--39.4 & -- & -- & 20.0--44.4 & -- \\
\addlinespace[1pt]
HarnessX~\cite{chen2026harnessx} & Qwen3.5-9B & \cite{chen2026harnessx} & O & -- & 23.6--41.8 & -- & -- & -- & -- \\
\addlinespace[1pt]
ACE~\cite{zhang2026ace} & GPT-5.4 (h) & \cite{lin2026ahe} & O & 69.7--68.9 & 75.2--74.6 & -- & -- & -- & -- \\
\addlinespace[1pt]
AHE~\cite{lin2026ahe} & GPT-5.4 & \cite{zhang2026harnesscompass} & M & -- & 51.8--55.5 & -- & -- & -- & -- \\
\addlinespace[1pt]
AHE~\cite{lin2026ahe} & GPT-5.4 (h) & \cite{lin2026ahe} & O & 69.7--77.0 & 75.2--75.6 & -- & -- & -- & -- \\
\addlinespace[1pt]
HarnessCompass~\cite{zhang2026harnesscompass} & GPT-5.4 & \cite{zhang2026harnesscompass} & M & -- & 51.8--61.0 & -- & -- & -- & -- \\
HarnessX~\cite{chen2026harnessx} & GPT-5.4 & \cite{chen2026harnessx} & O & -- & 45.5--63.6 & -- & -- & -- & -- \\
\addlinespace[1pt]
TF-GRPO~\cite{cai2025tfgrpo} & GPT-5.4 (h) & \cite{lin2026ahe} & O & 69.7--72.3 & 75.2--74.2 & -- & -- & -- & -- \\
\addlinespace[1pt]
HarnessBank~\cite{luo2026harnessbank} & Qwen3.6-27B & \cite{luo2026harnessbank} & U & 36.1--45.4 & 47.4--52.6 & -- & -- & 41.3--56.7 & -- \\
\addlinespace[1pt]
Dynamic Cheatsheet~\cite{suzgun2025dynamic} & GPT-5.5 (h) & \cite{pan2026rho} & T & 71--73 & -- & 59--62 & 29--30 & -- & -- \\
ReasoningBank~\cite{ouyang2026reasoningbank} & GPT-5.5 (h) & \cite{pan2026rho} & T & 71--73 & -- & 59--61 & 29--28 & -- & -- \\
RHO~\cite{pan2026rho} & GPT-5.5 (h) & \cite{pan2026rho} & T & 71--76 & -- & 59--78 & 29--37 & -- & -- \\
Sleep-time Compute~\cite{lin2025sleeptime} & GPT-5.5 (h) & \cite{pan2026rho} & T & 71--73 & -- & 59--64 & 29--32 & -- & -- \\
\bottomrule
\end{tabularx}
\endgroup
\end{table*}
\begin{table*}[!htb]
\ContinuedFloat
\centering
\caption{Self-evolving systems and harness optimizers (continued).}
\label{tab:self_evolving_cont}
\textbf{(b) Information, reasoning and environment tasks}\par\smallskip
\begingroup\fontsize{6.5}{8}\selectfont\setlength{\tabcolsep}{2pt}\renewcommand{\arraystretch}{1.12}
\begin{tabularx}{\textwidth}{@{}llrc*{4}{>{\centering\arraybackslash}X}@{}}
\toprule\rowcolor{lightblue}
\textbf{Method} & \textbf{Task model} & \textbf{Source} & \textbf{Split} & \shortstack{GAIA\\\cite{mialon2024gaia}} & \shortstack{HLE\\\cite{cais2026hle}} & \shortstack{ALFWorld\\\cite{shridharalfworld}} & \shortstack{WebShop\\\cite{yao2022webshop}} \\
\midrule
AFlow~\cite{zhang2025aflow} & GPT-5-Chat & \cite{cheng2026mem2evolve} & U & 18.47--19.75\textsuperscript{$\dagger$} & -- & 86.87--93.40\textsuperscript{$\dagger$} & 25.10--37.90\textsuperscript{$\dagger$} \\
Alita~\cite{qiu2025alita} & GPT-5-Chat & \cite{cheng2026mem2evolve} & U & 18.47--72.73\textsuperscript{$\dagger$} & -- & 86.87--86.13\textsuperscript{$\dagger$} & 25.10--30.21\textsuperscript{$\dagger$} \\
Mem$^2$Evolve~\cite{cheng2026mem2evolve} & GPT-5-Chat & \cite{cheng2026mem2evolve} & U & 18.47--76.31\textsuperscript{$\dagger$} & -- & 86.87--94.31\textsuperscript{$\dagger$} & 25.10--39.20\textsuperscript{$\dagger$} \\
\addlinespace[2pt]
GEPA~\cite{agrawal2026gepa} & GPT-5 mini & \cite{chen2026harnessfix} & T & 43.3--46.7 & -- & -- & -- \\
HarnessFix~\cite{chen2026harnessfix} & GPT-5 mini & \cite{chen2026harnessfix} & T & 43.3--61.7 & -- & -- & -- \\
Meta-Harness~\cite{lee2026metaharness} & GPT-5 mini & \cite{chen2026harnessfix} & T & 43.3--56.7 & -- & -- & -- \\
ReCreate~\cite{hao2026recreate} & GPT-5 mini & \cite{chen2026harnessfix} & T & 43.3--52.8 & -- & -- & -- \\
SCOPE~\cite{pei2025scope} & GPT-5 mini & \cite{chen2026harnessfix} & T & 43.3--45.0 & -- & -- & -- \\
\addlinespace[2pt]
HarnessFix~\cite{chen2026harnessfix} & Sonnet 4.5 & \cite{chen2026harnessfix} & T & 66.7--84.4 & -- & -- & -- \\
\addlinespace[2pt]
HarnessFix~\cite{chen2026harnessfix} & Gemini 3 Pro & \cite{chen2026harnessfix} & T & 69.4--83.9 & -- & -- & -- \\
\addlinespace[2pt]
HarnessFix~\cite{chen2026harnessfix} & DeepSeek V3.2 & \cite{chen2026harnessfix} & T & 58.9--75.6 & -- & -- & -- \\
\addlinespace[2pt]
HarnessFix~\cite{chen2026harnessfix} & Qwen3.5 Plus & \cite{chen2026harnessfix} & T & 63.3--78.9 & -- & -- & -- \\
\addlinespace[2pt]
HarnessX~\cite{chen2026harnessx} & Sonnet 4.6 & \cite{chen2026harnessx} & O & 73.8--83.5 & -- & 83.6--94.8 & 60.0--76.0 \\
\addlinespace[2pt]
HarnessX~\cite{chen2026harnessx} & Qwen3.5-9B & \cite{chen2026harnessx} & O & 20.3--37.4 & -- & 53.0--97.0 & 36.0--49.0 \\
\addlinespace[2pt]
HarnessX~\cite{chen2026harnessx} & GPT-5.4 & \cite{chen2026harnessx} & O & 73.8--73.8 & -- & 76.9--97.8 & 55.0--73.0 \\
\addlinespace[2pt]
ACE~\cite{zhang2026ace} & DeepSeek V4 Flash & \cite{zheng2026seagym} & T & -- & 22.5--25.0 & -- & -- \\
AHE~\cite{lin2026ahe} & DeepSeek V4 Flash & \cite{zheng2026seagym} & T & -- & 22.5--28.8 & -- & -- \\
TF-GRPO~\cite{cai2025tfgrpo} & DeepSeek V4 Flash & \cite{zheng2026seagym} & T & -- & 26.3--23.8 & -- & -- \\
\addlinespace[2pt]
Memento-Skills~\cite{zhou2026mementoskills} & Gemini 3.1 Flash & \cite{zhou2026mementoskills} & cell & 52.3--66.0\textsuperscript{O} & 17.9--38.7\textsuperscript{T} & -- & -- \\
\bottomrule
\end{tabularx}
\endgroup
\par\bigskip
\textbf{(c) Additional evaluation settings}\par\smallskip
\begingroup\fontsize{6.5}{8}\selectfont\setlength{\tabcolsep}{2pt}\renewcommand{\arraystretch}{1.12}
\begin{tabularx}{\textwidth}{@{}>{\RaggedRight\arraybackslash}p{61pt}>{\RaggedRight\arraybackslash}p{47pt}rc*{5}{>{\centering\arraybackslash}X}@{}}
\toprule\rowcolor{lightblue}
\textbf{Method} & \textbf{Task model} & \textbf{Source} & \textbf{Split} & \shortstack{BC+\\\cite{chen2025browsecompplus}} & \shortstack{GDPval\\\cite{patwardhan2025gdpval}} & \shortstack{LiveCode\\\cite{jain2025livecodebench}} & \shortstack{Omni-MATH\\\cite{gao2024omnimath}} & \shortstack{SWE\\\cite{jimenezswe}} \\
\midrule
EverOS~\cite{hu2026evermemos} & GPT-5.2 & \cite{tang2026msce} & U & 12.31--21.54 & 44.83--48.28 & 51.28--48.72 & 30.00--43.00 & 38.46--26.92 \\
EvoSkill~\cite{alzubi2026evoskill} & GPT-5.2 & \cite{tang2026msce} & U & 12.31--12.31 & 44.83--48.28 & 51.28--61.54 & 30.00--29.00 & 38.46--26.92 \\
Memento~\cite{zhou2025memento} & GPT-5.2 & \cite{tang2026msce} & U & 12.31--13.85 & 44.83--46.55 & 51.28--58.97 & 30.00--32.00 & 38.46--38.46 \\
MemSkill~\cite{zhang2026memskill} & GPT-5.2 & \cite{tang2026msce} & U & 12.31--15.38 & 44.83--46.55 & 51.28--56.41 & 30.00--29.00 & 38.46--38.46 \\
MSCE~\cite{tang2026msce} & GPT-5.2 & \cite{tang2026msce} & U & 12.31--26.15 & 44.83--53.45 & 51.28--61.54 & 30.00--47.00 & 38.46--53.85 \\
OpenSpace~\cite{hkuds2026openspace} & GPT-5.2 & \cite{tang2026msce} & U & 12.31--15.38 & 44.83--31.03 & 51.28--56.41 & 30.00--28.00 & 38.46--30.77 \\
SkillFlow-Evolve~\cite{zhang2026skillflow} & GPT-5.2 & \cite{tang2026msce} & U & 12.31--16.92 & 44.83--46.55 & 51.28--56.41 & 30.00--33.00 & 38.46--34.62 \\
\addlinespace[2pt]
HarnessBank~\cite{luo2026harnessbank} & Qwen3.6-27B & \cite{luo2026harnessbank} & U & 16.9--30.8 & 43.7--52.9 & 58.1--71.8 & 54.3--66.0 & -- \\
\bottomrule
\end{tabularx}
\endgroup
\par\smallskip\begin{minipage}{\textwidth}\footnotesize BC+: BrowseComp-Plus; LiveCode: LiveCodeBench; SWE: SWE-bench version unspecified by the source.\end{minipage}
\end{table*}

%% file: fig/table_cross_scenario.tex
\begin{table*}[!t]
\centering
\caption{Task constraints and framework choices across scenarios. Entries summarize conditional patterns; supporting examples and within-scenario differences are discussed in Sections~\ref{sec:sci}--\ref{sec:general_agents}.}
\label{tab:cross_scenario}
\fontsize{8}{9.2}\selectfont\setlength{\tabcolsep}{2pt}\renewcommand{\arraystretch}{1.12}
\begin{tabularx}{\textwidth}{@{}>{\RaggedRight\arraybackslash}p{0.16\textwidth}>{\RaggedRight\arraybackslash}p{0.21\textwidth}>{\RaggedRight\arraybackslash}p{0.225\textwidth}>{\RaggedRight\arraybackslash}p{0.155\textwidth}>{\RaggedRight\arraybackslash}X@{}}
\toprule
\rowcolor{lightblue}
\textbf{Scenario} & \textbf{Task constraints} & \textbf{Organization} & \textbf{Tool use} & \textbf{Oversight} \\
\midrule
\shortstack[l]{Scientific\\research} & \shortstack[l]{Computational checks;\\experimental feedback} & \shortstack[l]{Tool loops;\\specialist review} & \shortstack[l]{Retrieval and\\computation} & \shortstack[l]{Goal setting;\\proposal review} \\
\midrule
Healthcare & \shortstack[l]{Incomplete evidence;\\high cost of error} & \shortstack[l]{Specialist teams;\\deliberation} & \shortstack[l]{Records and\\diagnostic tools} & \shortstack[l]{Tiered and\\clinician review} \\
\midrule
\shortstack[l]{Software\\engineering} & \shortstack[l]{Artifact dependencies;\\executable feedback} & \shortstack[l]{Controllers or teams;\\artifact handoffs} & \shortstack[l]{Execution and\\repair search} & \shortstack[l]{Sandboxes;\\independent tests} \\
\midrule
\shortstack[l]{Social/economic\\simulation} & \shortstack[l]{Private objectives;\\interacting behavior} & \shortstack[l]{Individuals or groups;\\cooperation/competition} & \shortstack[l]{Memory and\\environment tools} & \shortstack[l]{Behavioral validation;\\risk control} \\
\midrule
\shortstack[l]{General-purpose\\agents} & \shortstack[l]{Varied tasks;\\uneven feedback} & \shortstack[l]{Planner--worker teams;\\verification} & \shortstack[l]{Search, code and\\reusable skills} & \shortstack[l]{Permissions;\\update gates} \\
\bottomrule
\end{tabularx}
\end{table*}

%% file: chapters/05_future_prospects.tex
\section{Challenges and Future Prospects}
\label{sec:future}
As discussed in previous chapters, agentic reasoning frameworks have made significant progress in both theory and application. However, the path to achieving a truly general, trustworthy, and efficient agent system is still full of challenges. In this chapter, we propose six potential directions for future development.

\subsection{Scalability and Efficiency of Reasoning}
\label{sec:efficiency}
As task complexity increases, the scalability and efficiency of agent frameworks have become major bottlenecks for large-scale applications. For instance, in multi-agent systems, specification, alignment and verification failures can undermine task completion \cite{pan2025multiagent}. Simply increasing the number of agents or extending reasoning time is often unsustainable, increasing inference and coordination costs. Evaluation must therefore account for resource budgets when attributing performance gains~\cite{kapoor2025hal}. Future works could focus on innovations at the framework level. On one hand, it will be crucial to design more efficient context management mechanisms for large-scale expansion. On the other hand, frameworks should also be equipped with dynamic task allocation and adaptive resource scheduling capabilities to handle complex tasks flexibly and efficiently.

\subsection{Recursive Self-Reasoning}
\label{sec:recursive}
As agent systems accumulate experience, a central challenge is to turn local corrections into lasting improvements in how they reason. Recursive self-reasoning extends this goal to the procedures that produce those improvements. Recent work has begun to explore this direction, but much of the adaptation remains in the code surrounding a fixed language model. For example, the Darwin G\"odel Machine and Meta-Harness revise the code that coordinates model calls and tools, without updating the underlying model weights~\cite{zhang2026dgm,lee2026metaharness}. Hyperagents also revises the procedure that proposes changes, although its main experiments retain fixed rules for selecting and evaluating new versions~\cite{zhang2026hyperagents}. Model adaptation offers a complementary approach: SEAL generates data and instructions for finetuning, but repeated updates can cause forgetting and require costly training and evaluation~\cite{zweiger2025seal}. Future work should connect improvements to the model and its surrounding framework, identifying when a failure calls for a change in reasoning organization or in learned capabilities. Evaluation should then examine whether successive updates preserve earlier skills and transfer to new tasks, using independent tests and comparable resource budgets.

\subsection{Execution Security and Privilege Control}
\label{sec:safety}
\label{sec:safety_practice}
The safety challenges for agent frameworks have evolved from securing a single language model to protecting the interfaces through which reasoning produces actions~\cite{wang2025comprehensive}. A central challenge is to enforce execution boundaries when tools, retrieved content and memory can influence subsequent decisions. IsolateGPT isolates third-party LLM applications and mediates communication, while OpenHands sandboxes code execution~\cite{wu2025isolategpt,wangopenhands}. Progent restricts tool calls through explicit privilege policies~\cite{shi2025progent}. Information-flow controls address a complementary boundary: CaMeL separates privileged planning from untrusted content processing and enforces capability policies~\cite{debenedetti2025camel}, while G-Safeguard filters propagation through multi-agent communication~\cite{wang2025gsafeguard}. Guardrail agents monitor another agent's actions rather than its text: GuardAgent maps a safety request to executable checking code that admits or denies the target agent's actions, and ShieldAgent verifies action trajectories against rules extracted from policy documents~\cite{xiang2025guardagent,chen2025shieldagent}. These mechanisms constrain available actions and information flows; they cannot by themselves establish the correctness of an authorized decision. Future work should examine how such boundaries remain effective across tool chains, shared memory and delegated agents, and how permissions are revalidated when a harness changes. This requires preserving the intended restrictions throughout the reasoning loop while allowing the interactions needed to complete a task.

\subsection{Human Oversight and Accountability}
\label{sec:oversight}
Human oversight concerns who can intervene in an agent's decisions and how responsibility is preserved across automated steps. Progent supports termination or user inspection when a tool call is disallowed~\cite{shi2025progent}; this connects policy enforcement to a person who can decide what happens next. In healthcare, Tiered Agentic Oversight combines automated review levels with a separate clinician-in-the-loop study in which physicians provide corrective feedback on triage cases~\cite{kim2025tiered}. Such studies motivate clearer handoff protocols: the reviewer needs the relevant evidence, proposed action and unresolved uncertainty, together with authority to pause, reject or resume the workflow. Action records can support this review by linking evidence, tools and decision authority~\cite{chan2024visibility,mu2025responsibility}. Accountability also includes examining whether role assignment, information access and aggregation propagate bias across affected groups~\cite{wang2024ali}. Future work should establish when escalation is necessary, how responsibility passes between agents and people, and how to preserve effective intervention without making review so burdensome that it becomes nominal.

\subsection{Safety Evaluation and Robustness}
\label{sec:safety_eval}
Safety evaluation asks whether execution controls and oversight remain effective under the failures and attacks a workflow may encounter. AgentDojo tests prompt injection, AgentHarm and Agent-SafetyBench examine harmful or unsafe behavior, and RedCode evaluates risky code~\cite{debenedetti2024agentdojo,andriushchenko2025agentharm,zhang2025agentsafetybench,guo2024redcode}. ToolEmu supplies emulated tool-use environments for risk testing~\cite{ruan2024toolemu}. These resources assess different properties, so their scores need to be interpreted against explicit threat models and allowed actions. Adaptive attacks can expose weaknesses missed by fixed attack suites~\cite{nasr2026attacker}; longer workflows also require testing how failures propagate across tools, memory and delegation. Future evaluations should measure attack success and unsafe actions alongside legitimate task completion, false blocking and recovery costs. For human oversight, missed escalations, reviewer workload and errors after resumption are equally relevant. Repeating these assessments after model or harness updates would help establish whether an improvement in task performance preserves the system's safety properties.

\subsection{Confidence Estimation and Explainable Agentic Reasoning}
\label{sec:confidence}  
When an agent system becomes an automated decision-maker, it needs a way to evaluate and communicate the trustworthiness of its reasoning process. This requires distinguishing uncertainty about a model's answer from uncertainty in the information and tools on which it depends. For instance, introspective reasoning can examine uncertainty within the reasoning process~\cite{han2024towards,liang2024introspective}, while active information seeking can help clarify ambiguity~\cite{hu2024uncertainty}. Confidence during tool invocation also matters, since external interactions introduce further sources of uncertainty~\cite{liu2024uncertainty}. Future evaluation should examine calibration on the intended tasks and whether uncertainty estimates improve decisions to gather evidence, abstain or request review. Explanations should connect conclusions to recorded observations and actions, with their faithfulness checked against that record. Such evidence can support oversight without treating a confident or plausible explanation as a safety guarantee.

%% file: chapters/06_limitations.tex
\section{Limitations}
\label{sec:limitations}
\textbf{Coverage and source maturity.} The core coverage and targeted updates in \S\ref{sec:related} provide a representative snapshot of a rapidly changing field. Recent preprints and industrial releases broaden coverage, but their reported results remain provisional where independent evaluation is unavailable. Later developments may require additional mechanisms or extensions to the taxonomy.

\textbf{Selection bias.} Selection by venue, mechanism, citation and reuse can favor established, visible and English-language work, while overlooking less visible systems or unsuccessful designs. The five main scenario categories also receive more detailed treatment than the additional areas in \S\ref{sec:scenarios}. The resulting design comparisons reflect the selected literature; they do not measure how frequently each architecture is used across the field.

\textbf{Evidence and abstraction.} Reported benchmark results differ in model, prompts, tools, budgets and evaluation protocols, as the table notes explain. They provide a dated reference for system capabilities and evaluation practice; isolating framework effects requires matched configurations. Likewise, the domain comparisons synthesize plausible relationships between constraints and design choices rather than identify causal performance effects. The framework-level abstraction separates reasoning organization from model training and abstracts the internal policies of memory and context management (\S\ref{sec:emerging}). These choices enable comparison but leave implementation-specific differences, and interactions between model and framework improvements, to more specialized analyses.

%% file: chapters/07_conclusion.tex
\section{Conclusion}
\label{sec:conclusion}
With the explosive growth of large language model (LLM) based agentic reasoning methods and applications, a systematic understanding of these approaches and their scenarios has become crucial. We propose a unified taxonomy that breaks down agentic systems into three compositional levels, from single-agent methods, tool-based methods, to multi-agent systems. This framework offers a clear views through which to analyze the field. Building on this, we systematically reviewed how these frameworks are put into practice across major application domains and general-purpose tasks, covering their core methodologies, key focuses, and evaluation approaches. Finally, we present our insights on the future directions of agentic reasoning, aiming to promote the development of agentic frameworks for the future generation.